\documentclass{article} % For LaTeX2e
\usepackage[final]{colm2026_conference}

\usepackage{microtype}
\usepackage{hyperref}
\usepackage{url}
\usepackage{graphicx}
\usepackage{amsmath,amssymb}
\usepackage{booktabs}
\usepackage{enumitem}
\usepackage{xspace}
\usepackage{multirow}
\usepackage{colortbl}
\usepackage{pifont}
\usepackage{cleveref}
\usepackage{wrapfig}
\usepackage{float}
\usepackage{placeins}
\usepackage{tcolorbox}
\usepackage{algorithm}
\usepackage{algpseudocode}
\usepackage{nicefrac}

\usepackage{lineno}

\definecolor{darkblue}{rgb}{0, 0, 0.5}
\hypersetup{colorlinks=true, citecolor=darkblue, linkcolor=darkblue, urlcolor=darkblue}

\title{Where Did It Go Wrong? Process-Level Evaluation\\of Web Agents with Semantic State Tracking}

\author{
Jiwan Chung\textsuperscript{1}
\quad
JiHyuk Byun\textsuperscript{1}
\quad
Vibhav Vineet\textsuperscript{2}
\quad
Seon Joo Kim\textsuperscript{1}
\\
\textsuperscript{1}Yonsei University
\quad
\textsuperscript{2}Microsoft Research
\\
\texttt{jiwan.chung.research@gmail.com}
}
\newcommand{\DATANAME}{\textsc{WebStep}\xspace}

\begin{document}

\ifcolmsubmission
\linenumbers
\fi

\maketitle

\begin{abstract}
Web agents act through long interaction sequences, yet existing benchmarks evaluate only terminal success, discarding all process information and offering little guidance on improvement. 
In this work, we conduct a process-level analysis of web agents.
We introduce~\DATANAME, a benchmark of 1,800 task instances with controlled difficulty and automatic semantic state tracking. Each website exposes a deterministic semantic MDP alongside the GUI: the agent operates on the interface, while the environment records high-level states and transitions in the background, enabling fine-grained analysis without manual annotation. 
Based on the semantic trajectory, we first show that process metrics reveal differences invisible to outcome evaluation: three agents whose success rates cluster within 34--37\% diverge in exploration reach versus execution accuracy. Then, decomposing by skill characterizes the nature of these differences, exposing opposite per-skill rankings hidden within the same website: e.g., on Q\&A, Claude CUA outperforms OpenAI CUA by 30\% on navigation actions yet underperforms it by 6.7\% on inspection, pinpointing a concrete skill to improve even within a domain. Bifurcation analysis further localizes the decisive error that loses the task and shows that this error is agent-specific rather than shared. Finally, these differences widen as tasks grow harder: success rate is similar on easy tasks but separates sharply as exploration becomes more demanding. Our process-level analysis opens a new avenue in web agent evaluation, providing fine-grained and actionable insight into where and how each agent should be improved.
Project page: \url{https://jiwanchung.github.io/webstep/}
\end{abstract}

\section{Introduction}
\label{sec:intro}

Web agents automate interactive digital tasks such as shopping, booking, and document management~\citep{fara,uitars,molmoweb}. Unlike single-turn instruction following, these tasks require multi-step navigation, information gathering across pages, and dependent action execution~\citep{webarena,visualwebarena}. Thus, the final outcome alone does not fully characterize agent behavior: the trajectory itself carries meaningful information about the interaction process.

\begin{figure*}[t]
\centering
\includegraphics[width=\linewidth]{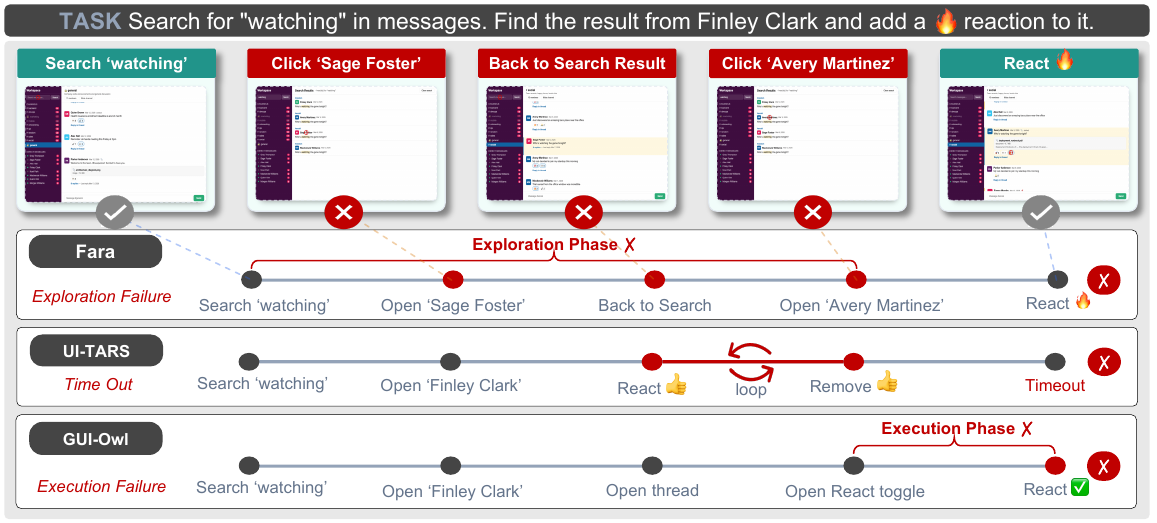}
\caption{\textbf{Terminal outcome alone cannot distinguish qualitatively different failures.} In the task of finding Finley Clark's message and reacting with a :fire: emoji, Fara fails to locate the correct target, while GUI-Owl reaches the correct thread but applies the wrong reaction.}
\label{fig:teaser}
\end{figure*}

Most existing web agent benchmarks, however, evaluate agents primarily by terminal success~\citep{webvoyager,gaia,mind2web,onlinemind2web}. This outcome-only view compresses an entire interaction trajectory into a single binary result, obscuring qualitatively different failure modes. As illustrated in~\Cref{fig:teaser}, an agent that reaches the correct colleague’s message thread but sends the wrong emoji receives the same score as one that never finds the relevant page at all. Terminal success therefore provides little actionable insight into where an agent failed or which capability limited performance, as it does not support accurate credit assignment~\citep{pignatellisurvey,luagentrewardbench,gritta2026process}.

In this work, we introduce \DATANAME, a benchmark for process-level evaluation of web agents. The core idea is to construct each website as a \emph{semantic Markov Decision Process (MDP) dual}: agents interact only with the rendered GUI, while the benchmark records the corresponding semantic state transitions induced underneath. Because the GUI is rendered directly from the semantic model, the full semantic trajectory is recovered exactly, enabling automatic state tracking and process-level evaluation without manual annotation.

Using \DATANAME, we show that process-level evaluation extends web agent evaluation from ranking to diagnosis. It separates agents that appear similar under terminal success by disentangling exploration from execution (\Cref{subsec:exp_agg}), attributing their differences to specific skills (\Cref{subsec:exp_skill}), localizing the step at which unsuccessful trajectories first diverge (\Cref{subsec:exp_bifurcation}), and identifying the task-complexity regimes in which these failures become most pronounced (\Cref{subsec:exp_complexity}). In turn, this yields concrete targets for improvement: which skills to strengthen, which temporal failure patterns to correct, and which complexity regimes to emphasize during training and evaluation.

Our main contributions are as follows:
\begin{enumerate}[leftmargin=*, labelsep=0.5em, itemsep=0.1em]
    \item We introduce \textbf{\DATANAME}, a benchmark of 10 self-hosted websites and 1,800 tasks, each paired with a semantic MDP dual that tracks structured states and transitions.
    \item We develop a \textbf{process-level evaluation framework} that decomposes agent behavior into exploration success, execution outcomes, and skill invocation patterns, and supports trajectory comparison at shared semantic states to localize failure modes.
    \item We present a \textbf{process-level analysis} of existing web agents, showing that models with similar terminal success rates exhibit qualitatively distinct failure signatures.
\end{enumerate}

\begin{figure*}[t]
\centering
\includegraphics[width=\linewidth]{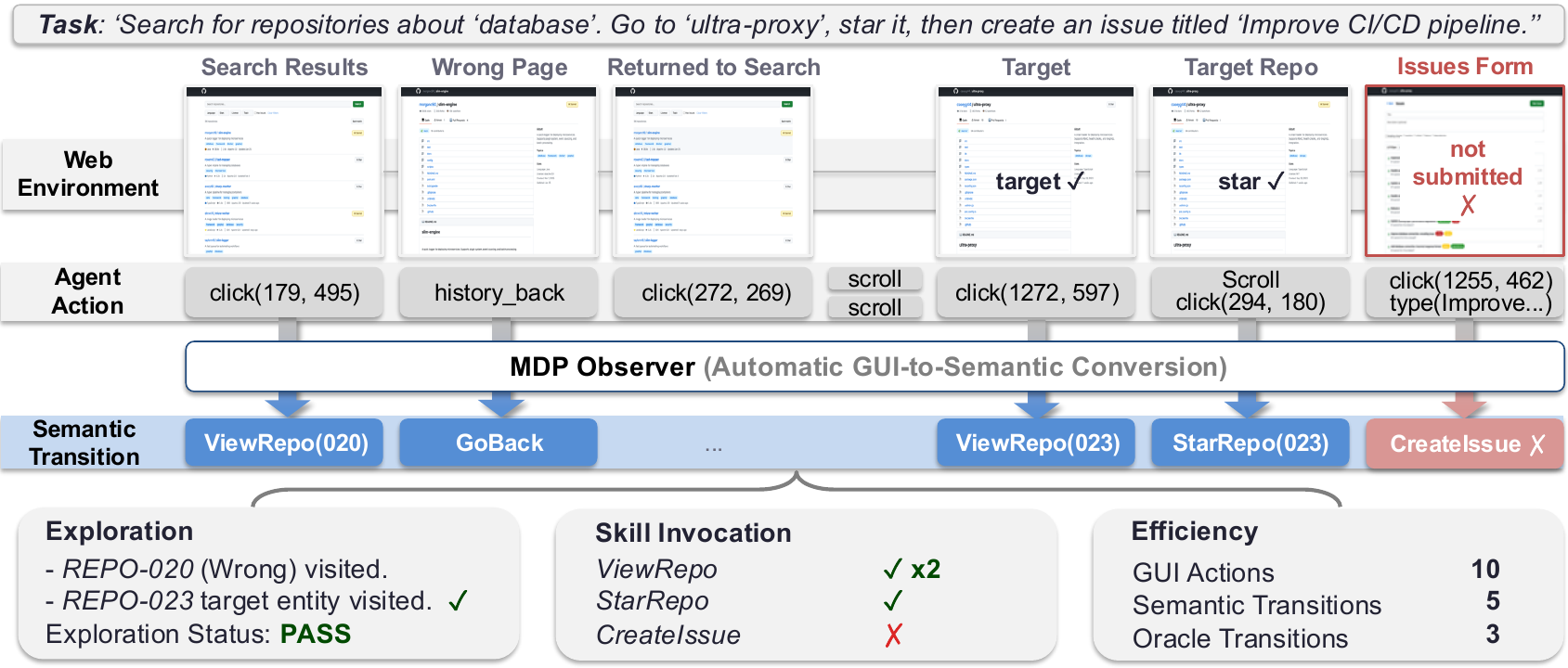}
\caption{\textbf{Overview of \DATANAME.} Each visual website is paired with a semantic MDP dual that converts raw GUI actions into interpretable semantic transitions. For example, clicking and typing merge to transitions such as \textsc{ViewRepo}, \textsc{StarRepo}, and \textsc{CreateIssue}, making explicit which item was visited and which semantic action was taken. These traces enable automatic process-level analyses such as exploration, skill invocation, and step efficiency.}
\label{fig:method}
\end{figure*}

\section{Process-level evaluation via semantic MDP duals}
\label{sec:concept}

Process-level evaluation requires access to the semantic effects of agent actions, yet web agents interact only through low-level operations such as clicking, typing, and scrolling. As shown in \Cref{fig:method}, we address this mismatch by mapping each raw GUI trajectory to an aligned semantic trace that records which items were visited, what information became available, and which high-level action was taken. This is enabled by constructing each website as a \emph{semantic MDP dual}: the agent acts only on the graphical user interface, while the benchmark tracks the corresponding semantic state transitions underneath.

\subsection{Semantic MDP}
\label{sec:semantic_mdp}

For each website, we define a deterministic semantic MDP
$\mathcal{M} = (\mathcal{S}, \mathcal{A}, T, \rho_0)$, where:

\begin{itemize}[leftmargin=*, labelsep=0.5em, itemsep=0.1em]
    \item $\mathcal{S}$: a structured semantic \textit{state} space. Each state is decomposed as $s=(s^p,s^i)$, where:
    \begin{itemize}[labelsep=0.5em, itemsep=0.1em]
        \item $s^p$: the current interface position (e.g., URL, search and filter settings, pagination and open modals);
        \item $s^i$: the item attributes revealed to the agent (e.g., a product's price, score, or seller).
    \end{itemize}
    \item $\mathcal{A}$: a typed semantic \textit{action} space abstracting away raw coordinate-level interactions (e.g., issuing a search, opening a detail page, or executing the final task action);
    \item $T$: a deterministic \textit{transition} function that maps each state-action pair $(s_t,a_t)$ to the next semantic state $s_{t+1}=T(s_t,a_t)$;
    \item $\rho_0$: the task and environment configurations, elaborated in~\Cref{sec:worldgen}.
\end{itemize}

For each agent run on a task, we record the resulting MDP trace $\tau$
\begin{equation}
\tau = (s_0, a_0, s_1, a_1, \dots, s_T),
\qquad
s_{t+1}=T(s_t,a_t),
\end{equation}
which serves as the basis for process-level evaluation.

\paragraph{GUI–MDP coupling.}
Each website is implemented so that the semantic MDP is the source of truth for all behavior. When the web agent performs a GUI action sequence (e.g., searching, filtering, or opening a detail page) the frontend forwards it to the semantic model instead of updating the page on its own. The semantic model applies the corresponding state transition, determines what information is visible in the new state, and the page is then re-rendered from that state. As a result, the task-relevant GUI trajectory is a direct execution of the semantic MDP, which allows the full semantic trajectory to be recovered exactly.
    
\subsection{Task and world generation}
\label{sec:worldgen}
 
Given a website's semantic MDP, each task consists of two jointly generated components: a natural-language \textit{instruction} and a corresponding \textit{world}, defined as a set of items with concrete attribute values (e.g., brand, price, seller, or delivery option). Each task is constructed by combining a task template (e.g., "Add a bag with [Constraints] to the cart") with a randomly sampled set of item-attribute constraints (e.g., price $\leq 300$ and rating $\geq 3.5$). The constraints are instantiated into text instruction, and the world is then populated to satisfy the resulting task specification. This coupled generation process enables:

\paragraph{Benchmark validity.} The world is guaranteed to contain exactly one target that satisfies all constraints in the instruction, so every task is \emph{solvable} and has a \emph{unique} answer.
 
\paragraph{Task complexity control.} Alongside the target, the generator places \emph{hard negative} items: distractors that share the target's visible attributes on list and search-result pages (e.g., same title, price, and category) but differ on an attribute visible only after opening the item's detail page (e.g., seller or delivery policy). Identifying the target on such tasks therefore requires navigating to individual detail pages. How many hard negatives are planted is itself a design variable: templates whose instruction names the target directly plant none, so tasks range from those solvable at list level to those demanding detail-page comparison.
This allows explicit problem complexity control, enabling complexity measure on three axes:
hard negative count, informational access level, and oracle trajectory length.
The corresponding definitions and analysis are in \Cref{subsec:exp_complexity}.

\paragraph{Oracle trajectory.} 
We can statically derive an \emph{oracle trajectory} for each task since both the MDP and the item set are fully specified. Refer to Appendix~\ref{sec:ax_oracle_gen} for details.
%, and the full generation procedure is described in Appendix~\ref{sec:ax_world_gen}.

\section{\DATANAME}
\label{sec:dataset}

\begin{table}[t]
\centering
\small
\begin{tabular}{@{}lcccc@{}}
\toprule
 & \textbf{Deterministic} & \textbf{Tasks} & \textbf{Process-eval} & \textbf{Hard negative} \\
\midrule
% Verified: 2504.01382, §2.2: 300 tasks, live web, WebJudge binary outcome
Online-Mind2Web \citep{onlinemind2web} & \ding{55} & 300 & \ding{55} & \ding{55} \\
% Verified: 2401.13919, §4.2: 643 tasks; live web; §5.1: binary GPT-4V judge
WebVoyager \citep{webvoyager} & \ding{55} & 643 & \ding{55} & \ding{55} \\
% Verified: 2311.12983, abstract: 466 questions; hybrid live+files; exact string match
%GAIA \citep{gaia} & \ding{55} & 466 & \ding{55} & \ding{55} \\
% Verified: 2501.07572, HuggingFace: 680 tasks; live web; QA accuracy
WebWalkerQA \citep{webwalkerqa} & \ding{55} & 680 & \ding{55} & \ding{55} \\
% Verified: 2407.15711, §3.3: 214 tasks; live web; §3.4: F1-based
AssistantBench \citep{assistantbench} & \ding{55} & 214 & \ding{55} & \ding{55} \\
% Verified: 2307.13854, §3.1: 812 tasks; self-hosted 4 clones; §3.2: functional verification
WebArena \citep{webarena} & {\color{green!60!black}\ding{51}} & 812 & \ding{55} & \ding{55} \\
% Verified: 2401.13649, intro+Table 3: 910 tasks; self-hosted 3 clones; §3.3: functional+visual
VisualWebArena \citep{visualwebarena} & {\color{green!60!black}\ding{51}} & 910 & \ding{55} & \ding{55} \\
% Verified: 2407.05291, abstract: 682 tasks; self-hosted ServiceNow; functional verification
WorkArena++ \citep{workarenaplus} & {\color{green!60!black}\ding{51}} & 682 & \ding{55} & \ding{55} \\
% Verified: 2406.12373, ICML 2024; abstract: 542 tasks, 2,439 key nodes
Mind2Web-Live \citep{mind2weblive} & \ding{55} & 542 & Key node & \ding{55} \\
\midrule
\rowcolor{blue!6}
\textbf{\DATANAME} & {\color{green!60!black}\ding{51}} & \textbf{1,800} & \textbf{MDP} & {\color{green!60!black}\ding{51}} \\
\bottomrule
\end{tabular}
\caption{\textbf{Comparison with existing web agent benchmarks.} \textit{Deterministic}: self-hosted deterministic websites. \textit{Process eval}: Key node uses manual milestones; MDP enables automatic stage decomposition and skill attribution. \textit{Hard negative}: controlled distractors.}
\label{tab:comparison}
\end{table}

\DATANAME\ is a benchmark for process-level evaluation of web agents in visually realistic environments with deterministic and semantic underlying dynamics.
It consists of 10 self-hosted websites spanning diverse domains (\Cref{fig:ax_sunburst}), including productivity tools, commerce platforms, professional and collaborative platforms, and technical knowledge sites. Each website is modeled after a real site from the same domain (Appendix~\ref{sec:ax_site_screenshots}), while replacing the backend with the deterministic semantic model defined in \Cref{sec:semantic_mdp}. Additional dataset statistics are provided in \Cref{tab:ax_data_numbers}.

\Cref{tab:comparison} positions \DATANAME\ among existing web agent benchmarks. Relative to live-website settings, it provides deterministic self-hosted environments. 
Compared to terminal-level benchmarks, it adds process-level evaluation grounded in an explicit semantic MDP and complexity control with task-conditioned world generation. Further discussion on related work is in Appendix~\ref{sec:ax_related}.

We design a multi-stage pipeline to construct \DATANAME. Starting from live website workflow traces, a coding agent~\citep{claude} with iterative human review produces a formal MDP specification and converts it into a self-hosted website driven by the semantic model. Each site is then validated through automated testing, including MDP unit tests, GUI interaction tests, and oracle trajectory replay, together with manual verification. Full pipeline details and example MDPs are provided in Appendices~\ref{sec:ax_site_construction} and~\ref{sec:ax_mdp_graphs}, respectively.

\section{Experiments and Results}
\label{sec:experiments}

We analyze web agent behavior using the process-level metrics automatically produced with \DATANAME. We begin by showing that they reveal differences hidden by outcome metrics (\Cref{subsec:exp_agg}), then characterize the nature of those differences (\Cref{subsec:exp_skill}), localize where they arise within trajectories (\Cref{subsec:exp_bifurcation}), and finally show that they widen as exploration difficulty increases (\Cref{subsec:exp_complexity}). Full results and qualitative examples appear in Appendices~\ref{sec:ax_more_results} and~\ref{sec:ax_qualitative}, respectively.

\paragraph{Models.} We evaluate six web agents spanning both small specialist models, UI-TARS-1.5~\citep{uitars}, Fara~\citep{fara}, and GUI-Owl-1.5~\citep{guiowl}, and large generalist models, Qwen3.5~\citep{qwen3}, OpenAI CUA (GPT-5.4)~\citep{openai_cua}, and Claude CUA (Claude Sonnet 4.6). To ensure a consistent evaluation protocol, we disable external actions such as cross-website navigation and Google search, and evaluate all tasks at non-critical terminal points.
More details are provided in Appendices~\ref{sec:ax_compute} and~\ref{sec:ax_eval_protocol}.

\begin{table}[t]
\centering
\resizebox{\linewidth}{!}
{
\small
\begin{tabular}{l >{\columncolor{gray!15}}c cc c ccc}
\toprule
\multirow{2}{*}{\textbf{Agent}}
& \multicolumn{3}{c}{\textbf{Success Rate} (\%) $\uparrow$}
& \textbf{Information} $\uparrow$
& \multicolumn{3}{c}{\textbf{Steps}} \\
\cmidrule(lr){2-4} \cmidrule(lr){5-5} \cmidrule(lr){6-8}
& \textbf{Terminal} & \textbf{Exploration} & \textbf{Execution}
& \textbf{Coverage (\%)}
& \textbf{GUI} & \textbf{Semantic} & $\nicefrac{\mathrm{GUI}}{\mathrm{Semantic}}$ $\downarrow$ \\
\midrule
Fara-7B & 35.7 & 47.3 & 75.0 & 76.4 & 18.9 & 8.3 & 2.3 \\
GUI-Owl-1.5-8B & 34.4 & 43.3 & 78.6 & 79.5 & 28.3 & 10.4 & 2.7 \\
UI-TARS-1.5-7B & 37.1 & 49.2 & 73.9 & 80.3 & 35.0 & 14.0 & 2.5 \\
\midrule
Qwen3.5-122B & 58.2 & 66.3 & 86.7 & 89.1 & 22.1 & 9.8 & 2.3 \\
OpenAI CUA & 82.7 & 89.4 & 92.4 & 96.8 & 19.7 & 10.0 & 2.0 \\
Claude CUA & 85.3 & 91.0 & 93.7 & 96.3 & 14.3 & 9.3 & 1.5 \\
\bottomrule
\end{tabular}
}
\caption{
\textbf{Aggregate evaluation results.}
We report terminal success together with process-level metrics from semantic MDP traces: exploration, execution, and information coverage. These metrics reveal behavioral differences not visible from terminal success alone.}
\label{tab:main}
\end{table}

\subsection{Aggregate Results}
\label{subsec:exp_agg}

We begin with coarse trajectory-level summaries derived from the semantic MDP traces. While these metrics do not yet expose fine-grained process-level breakdowns, they already reveal behavioral differences that terminal success alone obscures.

\paragraph{Metrics.}
All are computed automatically from the MDP trace $\tau = (s_0, a_0, s_1, \ldots, s_T)$.
\begin{itemize}[leftmargin=*, labelsep=0.5em, itemsep=0.1em]
    \item \textit{Terminal Success Rate (SR):} The standard binary outcome measuring whether the agent found the correct target and successfully executes the final commit action on it;
    \item \textit{Exploration SR:} Evaluates whether the agent isolates the correct target item $e^\star$ before committing. Formally, if $V_c = (v_1, \ldots, v_k)$ denotes the sequence of items visited on detail surfaces prior to the first commit, exploration succeeds if and only if $v_k = e^\star$;
    \item \textit{Execution SR:} The terminal success rate conditioned strictly on successful exploration. This isolates the agent's ability to finalize a task once the correct target has been found;
    \item \textit{Informational Coverage:} The fraction of task-relevant attributes the agent witnessed, serving as a measure of information-gathering thoroughness (Refer to~\Cref{subsec:ax_info_coverage});
    \item \textit{Step Efficiency:} We record both raw \textit{GUI steps} and \textit{semantic steps} (state-changing MDP transitions). Their ratio quantifies the agent's interaction  efficiency.
\end{itemize}

\paragraph{Results.}
Table~\ref{tab:main} reports aggregate performance. Under terminal success, the small agents appear largely similar, clustering at 34--37\%. The trajectory-derived metrics, however, reveal substantial behavioral differences. UI-TARS identifies the correct target more often than GUI-Owl ($\Delta$ +5.9\%) but performs worse during execution ($\Delta$ -4.7\%). Its higher information coverage and step count indicate a broader, more exploratory strategy. In contrast, Fara takes the fewest steps among the small agents and has the lowest information coverage of any agent evaluated, suggesting more limited exploration.

The larger models, Qwen3.5, OpenAI CUA, and Claude CUA, perform better in both exploration and execution, and also achieve higher information coverage. But this higher coverage is not just a result of exploring more: OpenAI CUA covers more information than Qwen3.5 while using steps more efficiently, and its lower GUI-to-semantic step ratio shows that more of its actions produce meaningful semantic progress.

\begin{tcolorbox}[colback=gray!10, colframe=black!100, boxrule=0.5pt, left=6pt, right=6pt, top=4pt, bottom=4pt]
Finding 1. \textbf{Terminal success hides behavioral differences.} e.g., while the small-scale agents cluster at similar terminal success, trajectory-level summaries separate UI-TARS as exploration-strong but execution-weak, and Fara as under-exploratory (\Cref{tab:main}).
\end{tcolorbox}

\subsection{Skill-level diagnosis}
\label{subsec:exp_skill}

\begin{figure*}[t]
\centering
\includegraphics[width=\linewidth]{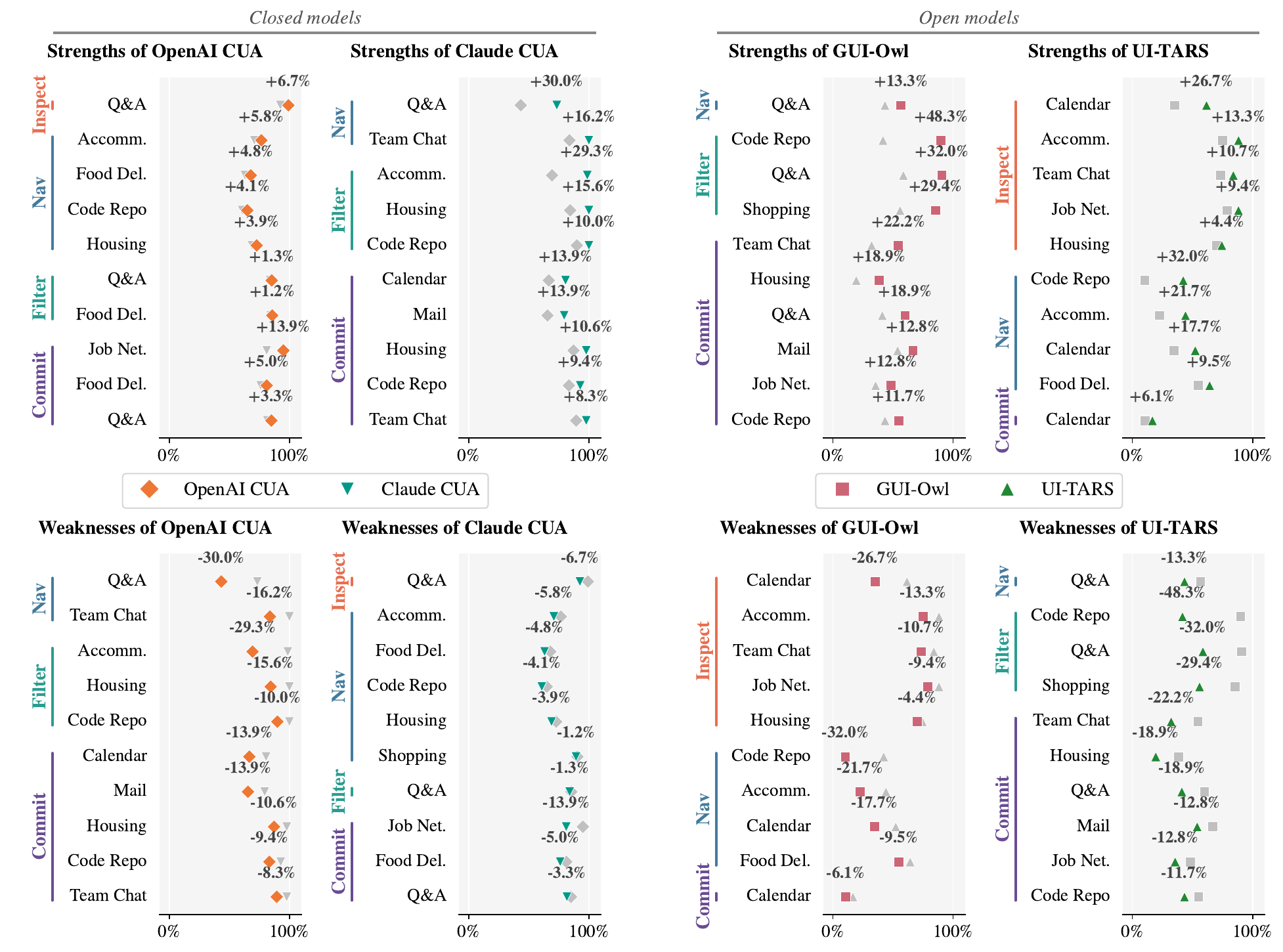}
\caption{\textbf{Skill-level strengths and weaknesses.} Agents exhibit distinct within-website skill profiles. e.g., on Housing, OpenAI CUA is stronger on \textit{navigation} but weaker on \textit{filter}, showing that an agent's weakness can be concentrated in a specific interaction skill. Point labels (e.g., $+16.2\%$) denote the gap relative to the compared model.}
\label{fig:skills}
\end{figure*}

Aggregate summaries show overall patterns, but they do not isolate the specific interaction capability responsible for a model's weakness. A model may appear strong or weak on a website for different underlying reasons, and broad summaries can hide these differences by mixing opposing skill-level signals. To make the diagnosis actionable, the evaluation must identify which interaction skills are present in the model's behavior and which are not.

We operationalize this with skill \textit{invocation}. For each task, we use an oracle trajectory as a reference. Although oracle trajectories are not unique solutions, they instantiate shortest valid strategies for completing the task. Let $S^\star$ be the set of skill types that appear in the oracle trajectory, and let $S$ be the set of skill types that appear in the model trajectory. For each skill $k \in S^\star$, we define \textit{invocation} as whether $k \in S$.\footnote{We do not measure skill \emph{success} because a semantic skill is usually identifiable only upon completion. Unsuccessful partial attempts therefore often lack a well-defined skill label. For example, a \textit{search} is identifiable only after query entry and submission.}

\begin{wrapfigure}{rt}{0.5\textwidth}
  \centering
  \includegraphics[width=0.48\textwidth]{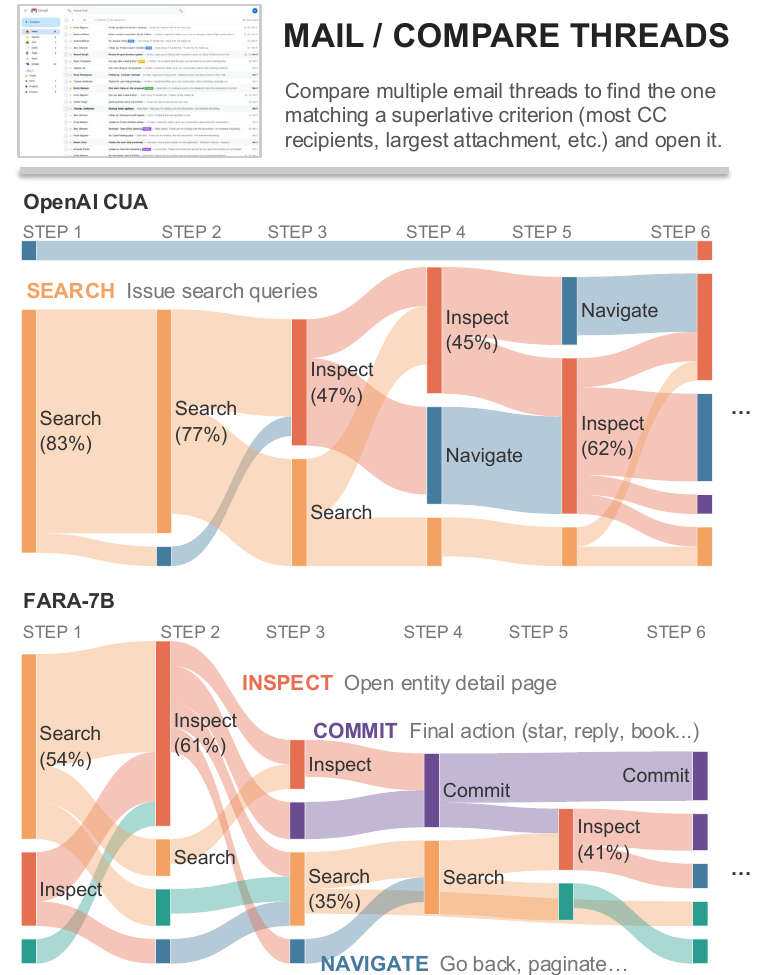}
  \caption{\textbf{Temporal skill invocation patterns.}}
  \label{fig:trajectory_sankey}
\end{wrapfigure}

\paragraph{Skills.}
We map MDP actions into five categories (full definitions in~\Cref{tab:ax_skill_categories,tab:ax_skills}):
\begin{itemize}[leftmargin=*, labelsep=0.5em, itemsep=0.1em]
\item \textit{inspect}: opening an item's detail view;
\item \textit{navigate}: transitioning between interfaces without directly narrowing candidates;
\item \textit{search}: text query to retrieve candidates;
\item \textit{filter}: using the filtering interface to reduce the candidate set;
\item \textit{commit}: performing the final task-completing action on the selected target.
\end{itemize}

\paragraph{Strengths and weaknesses.}
\Cref{fig:skills} shows that agents exhibit distinct skill profiles. Between Claude CUA and OpenAI CUA, Claude CUA is generally stronger at executing the final \textit{Commit} action and performing \textit{Filter}, although their relative proficiencies vary across websites. Among the smaller agents, GUI-Owl is stronger at \textit{Filter}, whereas UI-TARS is stronger at \textit{Navigate}. These contrasts show that performance differences between agents are often concentrated in specific interaction skills.

These differences become even sharper within a single website. On the Q\&A website, Claude CUA substantially outperforms OpenAI CUA on \textit{Navigate} ($\Delta +30.0\%$) but underperforms on \textit{Inspect} ($\Delta -6.7\%$). A website-level score would average these opposing signals and obscure the underlying diagnosis. Skill-level analysis instead pinpoints the specific weakness: in this case, item inspection. This diagnosis directly suggests adding training examples that strengthen inspection behavior.

\paragraph{Temporal structure of skill invocation.}
Semantic MDP traces also reveal \emph{when} skills are invoked. While~\Cref{fig:skills} shows which required skills agents tend to miss, \Cref{fig:trajectory_sankey} shows when they are used along the trajectory. OpenAI CUA follows a concentrated early-stage pattern, typically beginning with \textit{Search}, whereas Fara's early actions are more dispersed. Fara also invokes \textit{Commit} earlier, while OpenAI CUA spends more early steps on information-gathering skills such as \textit{Search}, \textit{Navigate}, and \textit{Inspect}. This suggests targeted interventions: reinforcing a more consistent initial skill sequence for Fara, and delaying commitment until more information has been gathered.

\begin{tcolorbox}[colback=gray!10, colframe=black!100, boxrule=0.5pt, left=6pt, right=6pt, top=4pt, bottom=4pt]
Finding 2: \textbf{Agent weaknesses can be skill-specific.} In Q\&A, Claude CUA shows strong \textit{Navigate} skills, but underperforms on \textit{Inspect} usage, indicating that its weakness in this website is not uniform, but concentrated in a specific interaction skill (\Cref{fig:skills}).
\end{tcolorbox}

\begin{figure*}[t]
\centering
\includegraphics[width=\linewidth]{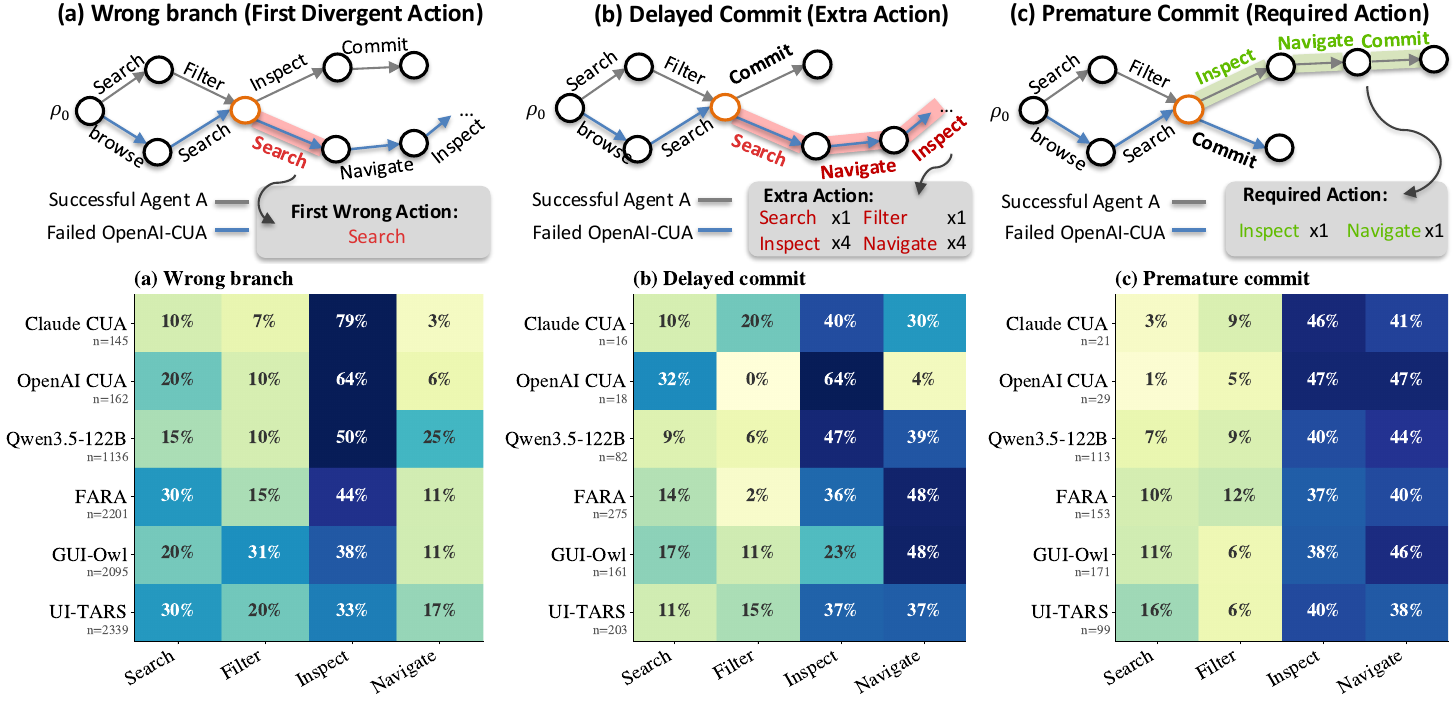}
\caption{\textbf{Trajectory bifurcations at shared semantic states.} Bifurcation analysis localizes where failure is introduced. (a) \textit{Wrong branch}: the decision that first sends the failing trajectory off course. (b) \textit{Delayed commit}: the extra behavior after missing the right time to finish. (c) \textit{Premature commit}: the missing behavior skipped before finishing too early.}
\label{fig:failure_attribution}
\end{figure*}

\subsection{Trajectory bifurcations}
\label{subsec:exp_bifurcation}

The preceding analyses identify which skills are weak, but they do not yet show \emph{where} failure is introduced within a trajectory. We address this by comparing successful and failing trajectories from the same task at the level of semantic MDP states. Since all agents act in the same Markovian environment, trajectories can be aligned whenever they visit the same state. We define the \emph{bifurcation point} as the last shared state before divergence.

We categorize bifurcation points into three types, as shown in the top row of~\Cref{fig:failure_attribution}:
\begin{itemize}[leftmargin=*, labelsep=0.5em, itemsep=0.1em]
    \item \textit{Wrong branch:} the two next actions differ, and neither is \textsc{Commit}. We show the mismatched next action.
    \item \textit{Delayed commit:} the successful next action is \textsc{Commit}, but the failing next action is not. We show the remaining failing suffix.
    \item \textit{Premature commit:} the failing next action is \textsc{Commit}, but the successful next action is not. We show the remaining successful suffix.
\end{itemize}

\paragraph{Wrong branches.}
\Cref{fig:failure_attribution}~(a) identifies the decision that first sends a trajectory onto a losing branch, revealing the branching error each agent is most prone to. For most agents, this first wrong step is \textit{Inspect}, suggesting that they often branch off by opening the wrong item page (33--79\%). Fara and UI-TARS also frequently fall back to \textit{Search} when they go off track (30\%), whereas GUI-Owl more often diverges through \textit{Filter} (31\%). These patterns show that agents tend to enter failure through different kinds of decisions.

\paragraph{Delayed commit.}
\Cref{fig:failure_attribution}~(b) characterizes how agents behave after missing the point at which a successful trajectory would already have finished, revealing the kinds of unnecessary actions they continue to take.
 GUI-Owl and FARA spend 48\% of these extra actions on \textit{Navigate}, whereas OpenAI CUA instead continues mainly with \textit{Inspect} (64\%) and \textit{Search} (32\%). Their delayed failures therefore take different forms: continued evidence gathering and query refinement for OpenAI CUA, and continued navigation for GUI-Owl and FARA.

\paragraph{Premature commit.}
\Cref{fig:failure_attribution}~(c) shows which types of evidence agents most often fail to gather before committing prematurely.
Across all models, the most frequently omitted actions are \textit{Inspect} and \textit{Navigate}, indicating that premature commits primarily arise from insufficient examination of the current page and inadequate exploration of alternative states.
The two stronger models, OpenAI CUA and Claude CUA, rarely omit \textit{Search} actions (1--3\%), suggesting that they generally identify when external information retrieval is necessary but may still commit before fully inspecting or navigating through the retrieved evidence.
Smaller models, including FARA, GUI-Owl, and UI-TARS, exhibit a different pattern: in 10--16\% of premature-commit cases, the omitted action is \textit{Search}.
This indicates a more fundamental planning failure, in which the agent commits without first recognizing that the task requires information beyond the currently available interface state.

\begin{tcolorbox}[colback=gray!10, colframe=black!100, boxrule=0.5pt, left=6pt, right=6pt, top=4pt, bottom=4pt]
Finding 3. \textbf{The decisive error point that loses the task differs across agents.} e.g., although \textit{Inspect} is the most frequent first divergence for every agent, GUI-Owl branches off through \textit{Filter} far more often than the others (31\%), while Fara and UI-TARS instead fall back to \textit{Search} (30\%) (\Cref{fig:failure_attribution}).
\end{tcolorbox}

\begin{figure*}[t]
\centering
\includegraphics[width=\linewidth]{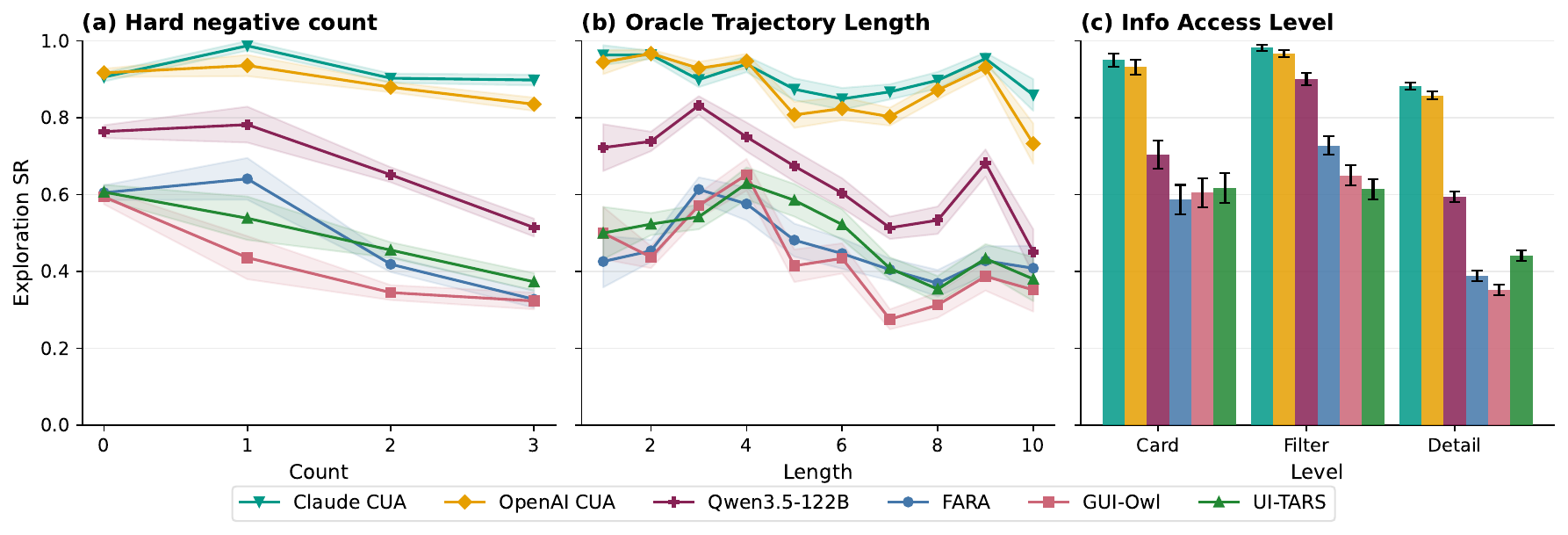}

\caption{\textbf{Exploration success by task complexity.} \textbf{(a)}~Performance by \emph{hard negative counts} separates OpenAI CUA from others. \textbf{(b)}~\emph{Oracle trajectory length} shows downward trend. \textbf{(c)}~Tasks requiring page \textit{detail} evidence show different model gaps from list-level tasks.}%Shaded regions: $\pm$1~SE.}
\label{fig:complexity}
\end{figure*}

\subsection{Exploration success by task complexity}
\label{subsec:exp_complexity}

Having localized where individual trajectories first go wrong, we next ask whether failure follows systematic patterns as exploration difficulty increases.

\paragraph{Task-complexity measures.}
We consider three complementary measures:
\begin{itemize}[leftmargin=*, labelsep=0.5em, itemsep=0.1em]
    \item \textit{Hard negative count:} the number of plausible distractor items in the world;
    \item \textit{Oracle trajectory length:} measures the minimum number of steps in the oracle trajectory;
    \item \textit{Information access level:} groups tasks by the exploration type required to identify the target.
    \begin{itemize}[labelsep=0.5em, itemsep=0.1em]
        \item \textit{Card} for list-level information only;
        \item \textit{Filter} for tasks requiring filtering or search;
        \item \textit{Detail} for tasks requiring explicit visits to item detail pages.
    \end{itemize}
\end{itemize}
%Exploration becomes harder from \textit{Card} to \textit{Filter} to \textit{Detail}, as identifying the target requires increasingly active information gathering.

\paragraph{Hard negative count.}
\Cref{fig:complexity}~(a) shows that task structure strongly shapes exploration difficulty. We read the trend over HN$\geq$1, since HN$=$0 tasks name the target directly and therefore plant no distractors by construction (\Cref{sec:ax_world_gen}). Over this range, exploration success declines monotonically for all six agents, confirming that the procedurally generated distractors induce genuine exploration difficulty. The strong models Claude CUA and OpenAI CUA are the least sensitive, retaining roughly 90\% of their HN$=$1 exploration success at HN$=$3, whereas the remaining agents lose 26--49\% of theirs, suggesting substantially stronger exploration capacity.

\paragraph{Oracle trajectory length.} \Cref{fig:complexity}~(b) shows a similar pattern. As the minimum successful trajectory becomes longer, exploration performance generally degrades. Qwen3.5 exhibits a relatively monotonic drop as trajectory length increases, whereas Claude CUA and OpenAI CUA maintain their performance for longer, suggesting better generalization to more complex tasks.

\paragraph{Information access level.} \Cref{fig:complexity}~(c) further separates the agents by exploration types required. Performance is relatively similar on card tasks, where only list-level inspection is needed. Filter tasks begin to separate the larger models from the smaller ones, and detail tasks further separate the strongest models, Claude CUA and OpenAI CUA, from the third-best model, Qwen3.5.
Thus, complex exploration requirements reveal differences in the agents.

\begin{tcolorbox}[colback=gray!10, colframe=black!100, boxrule=0.5pt, left=6pt, right=6pt, top=4pt, bottom=4pt]
Finding 4. \textbf{Agent differences in exploration widen as tasks become harder.} Exploration performance is relatively similar on easier tasks, but separates much more clearly as exploration requirements become more demanding (\Cref{fig:complexity}).
\end{tcolorbox}

%Finding 4. \textbf{Trajectory bifurcations localize failures to characteristic action types.} GUI-Owl's first mistakes are predominantly incorrect filters (45\%), while OpenAI CUA's originate from search formulation (55\%); hesitation and premature commitment emerge as separable failure modes with distinct per-agent skill compositions

%\subsection{Qualitative analysis}
%\label{sec:exp_qual}
%\input{figs/figure9}

\section{Conclusion}

%We introduced \DATANAME, a benchmark for automatic process-level evaluation of web agents. By pairing each self-hosted website with structured state tracking, \DATANAME enables fine-grained analysis of agent behavior without manual trajectory annotation. Our results show that terminal success alone obscures substantial differences in how agents explore, act, and fail, while process-level evaluation makes these differences interpretable and actionable. We hope \DATANAME will support future work on more diagnostic and informative evaluation of web agents. Limitations are discussed in Appendix~\ref{sec:ax_limitations}.

We introduced \DATANAME, a benchmark for automatic process-level evaluation of web agents. By pairing each self-hosted website with semantic state tracking, \DATANAME enables fine-grained analysis of agent behavior without manual trajectory annotation. Our results show that terminal success alone hides major differences in how agents explore, act, and fail, whereas process-level evaluation makes these differences directly diagnosable. We hope \DATANAME supports deeper diagnosis of web agents. Limitations are in Appendix~\ref{sec:ax_limitations}.

\section*{Ethics Statement}

\DATANAME is a benchmark for evaluating web agent behavior in controlled, self-hosted environments. It does not involve human-subject experiments, real user accounts, or interaction with live online services during evaluation. All websites are independently implemented reproductions of publicly observable front-end interaction patterns (screenshots and keyboard+mouse movements), and the benchmark uses synthetic world data rather than real user data or production backends. We recognize that improved web agent evaluation can have dual-use implications, including downstream misuse on real websites. Our work is intended to support diagnostic analysis in sandboxed environments, and all experiments are confined to offline, self-hosted replicas with no external side effects. Details of LLM usage are provided in Appendix~\ref{subsec:ax_llm_usage}.

\section*{Reproducibility Statement}

We will publicly release the full benchmark, including all 10 website implementations, all 1,800 task definitions with world seeds and oracle trajectories, and the complete evaluation suite. Each environment is formalized as a deterministic MDP with a fixed world seed, so the environment dynamics, verification outcomes, and semantic traces are reproducible given the same agent actions. This guarantees reproducibility of the benchmark conditions, although full end-to-end run reproducibility may still be affected by model-side sources of non-determinism such as stochastic decoding or API variation. Agent hyperparameters, model identifiers, and decoding settings are reported in \Cref{tab:ax_agent_config}. Additional details on environment determinism and residual sources of non-determinism are provided in Appendix~\ref{sec:ax_reproducibility}.

\bibliography{colm2026_conference}

@inproceedings{webarena,
  title={WebArena: A Realistic Web Environment for Building Autonomous Agents},
  author={Zhou, Shuyan and Xu, Frank F and Zhu, Hao and Zhou, Xuhui and Lo, Robert and Sridhar, Abishek and Cheng, Xianyi and Bisk, Yonatan and Fried, Daniel and Alon, Uri and others},
  booktitle={ICLR},
  year={2024}
}

@inproceedings{mind2web,
  title={Mind2Web: Towards a Generalist Agent for the Web},
  author={Deng, Xiang and Gu, Yu and Zheng, Boyuan and Chen, Shijie and Stevens, Samuel and Wang, Boshi and Sun, Huan and Su, Yu},
  booktitle={NeurIPS},
  year={2023}
}

@article{webvoyager,
  title={WebVoyager: Building an End-to-End Web Agent with Large Multimodal Models},
  author={He, Hongliang and Yao, Wenlin and Ma, Kaixin and Yu, Wenhao and Dai, Yong and Zhang, Hongming and Lan, Zhenzhong and Yu, Dong},
  journal={arXiv preprint arXiv:2401.13919},
  year={2024}
}

@article{visualwebarena,
  title={VisualWebArena: Evaluating Multimodal Agents on Realistic Visual Web Tasks},
  author={Koh, Jing Yu and Lo, Robert and Jang, Lawrence and Duvvur, Vikram and Lim, Ming Chong and Huang, Po-Yu and Neubig, Graham and Zhou, Shuyan and Salakhutdinov, Ruslan and Fried, Daniel},
  journal={arXiv preprint arXiv:2401.13649},
  year={2024}
}

@inproceedings{agentbench,
  title={AgentBench: Evaluating LLMs as Agents},
  author={Liu, Xiao and Yu, Hao and Zhang, Hanchen and Xu, Yifan and Lei, Xuanyu and Lai, Hanyu and Gu, Yu and Ding, Hangliang and Men, Kaiwen and Yang, Kejuan and others},
  booktitle={ICLR},
  year={2024}
}

@article{miniwob,
  title={Reinforcement Learning on Web Interfaces Using Workflow-Guided Exploration},
  author={Liu, Evan Zheran and Guo, Kelvin and Pasupat, Panupong and Shi, Tianlin and Liang, Percy},
  journal={arXiv preprint arXiv:1802.08802},
  year={2018}
}

@article{osworld,
  title={OSWorld: Benchmarking Multimodal Agents for Open-Ended Tasks in Real Computer Environments},
  author={Xie, Tianbao and Zhang, Danyang and Chen, Jixuan and Li, Xiaochuan and Zhao, Siheng and Cao, Ruisheng and Hua, Toh Jing and Cheng, Zhoujun and Shin, Dongchan and Lei, Fangyu and others},
  journal={arXiv preprint arXiv:2404.07972},
  year={2024}
}

@article{workarena,
  title={WorkArena: How Capable Are Web Agents at Solving Common Knowledge Work Tasks?},
  author={Drouin, Alexandre and Gasse, Maxime and Caccia, Massimo and Laradji, Issam H and Del Verme, Manuel and Marber, Tom and Vazquez, David and Chapados, Nicolas and Lacoste, Alexandre},
  journal={arXiv preprint arXiv:2403.07718},
  year={2024}
}

@article{webshop,
  title={WebShop: Towards Scalable Real-World Web Interaction with Grounded Language Agents},
  author={Yao, Shunyu and Chen, Howard and Yang, John and Narasimhan, Karthik},
  journal={NeurIPS},
  year={2022}
}

@article{prm,
  title={Let's Verify Step by Step},
  author={Lightman, Hunter and Kosaraju, Vineet and Burda, Yura and Edwards, Harri and Baker, Bowen and Lee, Teddy and Leike, Jan and Schulman, John and Sutskever, Ilya and Cobbe, Karl},
  journal={ICLR},
  year={2024}
}

@article{gaia,
  title={GAIA: A Benchmark for General AI Assistants},
  author={Mialon, Gr{\'e}goire and Fourrier, Cl{\'e}mentine and Swift, Craig and Wolf, Thomas and LeCun, Yann and Scialom, Thomas},
  journal={arXiv preprint arXiv:2311.12983},
  year={2023}
}

@inproceedings{onlinemind2web,
  title={An Illusion of Progress? Assessing the Current State of Web Agents},
  author={Xue, Tianci and Qi, Weijian and Shi, Tianneng and Song, Chan Hee and Gou, Boyu and Song, Dawn and Sun, Huan and Su, Yu},
  booktitle={COLM},
  year={2025}
}

@article{workarenaplus,
  title={{WorkArena++}: Towards Compositional Planning and Reasoning-based Common Knowledge Work Tasks},
  author={Boisvert, L{\'e}o and Thakkar, Megh and Gasse, Maxime and Caccia, Massimo and {Le Sellier De Chezelles}, Thibault and Cappart, Quentin and Chapados, Nicolas and Lacoste, Alexandre and Drouin, Alexandre},
  journal={arXiv preprint arXiv:2407.05291},
  year={2024}
}

@article{webwalkerqa,
  title={{WebWalker}: Benchmarking {LLMs} in Web Traversal},
  author={Wu, Jialong and Yin, Wenbiao and Jiang, Yong and Wang, Zhenglong and Xi, Zekun and Fang, Runnan and Zhang, Linhai and He, Yulan and Zhou, Deyu and Xie, Pengjun and Huang, Fei},
  journal={arXiv preprint arXiv:2501.07572},
  year={2025}
}

@article{assistantbench,
  title={{AssistantBench}: Can Web Agents Solve Realistic and Time-Consuming Tasks?},
  author={Yoran, Ori and Amouyal, Samuel Joseph and Malaviya, Chaitanya and Bogin, Ben and Press, Ofir and Berant, Jonathan},
  journal={arXiv preprint arXiv:2407.15711},
  year={2024}
}

@inproceedings{agentboard,
  title={{AgentBoard}: An Analytical Evaluation Board of Multi-turn {LLM} Agents},
  author={Ma, Chang and Zhang, Junlei and Liu, Zhihao and Chen, Jiawei and Wang, Yitao and others},
  booktitle={NeurIPS},
  year={2024}
}

@inproceedings{mind2weblive,
  title={{WebCanvas}: Benchmarking Web Agents in Online Environments},
  author={Pan, Yichen and Kong, Dehan and Zhou, Sida and Cui, Cheng and Leng, Yifei and Jiang, Bing and Liu, Hangyu and Shang, Yanyi and Zhou, Shuyan and Wu, Tongshuang and Wu, Zhengyang},
  booktitle={ICML},
  year={2024}
}

@article{fara,
  title={Fara-7B: An Efficient Agentic Model for Computer Use},
  author={Awadallah, Ahmed and Lara, Yash and Magazine, Raghav and Mozannar, Hussein and Nambi, Akshay and Pandya, Yash and Rajeswaran, Aravind and Rosset, Corby and Taymanov, Alexey and Vineet, Vibhav and Whitehead, Spencer and Zhao, Andrew},
  journal={arXiv preprint arXiv:2511.19663},
  year={2025}
}

@article{uitars,
  title={{UI-TARS}: Pioneering Automated {GUI} Interaction with Native Agents},
  author={Qin, Yujia and Ye, Yining and Fang, Junjie and Wang, Haoming and Liang, Shihao and Tian, Shizuo and Zhang, Junda and Li, Jiahao and Li, Yunxin and Huang, Shijue and Zhong, Wanjun and Li, Kuanye and Yang, Jiale and Miao, Yu and Lin, Woyu and Liu, Longxiang and Jiang, Xu and Ma, Qianli and Li, Jingyu and Xiao, Xiaojun and Cai, Kai and Li, Chuang and Zheng, Yaowei and Jin, Chaolin and Li, Chen and Zhou, Xiao and Wang, Minchao and Chen, Haoli and Li, Zhaojian and Yang, Haihua and Liu, Haifeng and Lin, Feng and Peng, Tao and Liu, Xin and Shi, Guang},
  journal={arXiv preprint arXiv:2501.12326},
  year={2025}
}

@article{guiowl,
  title={Mobile-Agent-v3: Fundamental Agents for {GUI} Automation},
  author={Ye, Jiabo and Zhang, Xi and Xu, Haiyang and Liu, Haowei and Wang, Junyang and Zhu, Zhaoqing and Zheng, Ziwei and Gao, Feiyu and Cao, Junjie and Lu, Zhengxi and Liao, Jitong and Zheng, Qi and Huang, Fei and Zhou, Jingren and Yan, Ming},
  journal={arXiv preprint arXiv:2508.15144},
  year={2025}
}

@techreport{molmoweb,
  title={{MolmoWeb}: Open Visual Web Agent and Open Data for the Open Web},
  author={Gupta, Tanmay and others},
  institution={Allen Institute for AI},
  year={2026},
  url={https://allenai.org/papers/molmoweb}
}

@article{qwen3,
  title={Qwen3 Technical Report},
  author={Yang, An and Li, Anfeng and Yang, Baosong and Zhang, Beichen and Hui, Binyuan and Zheng, Bo and Yu, Bowen and Gao, Chang and Huang, Chengen and Lv, Chenxu and Zheng, Chujie and Liu, Dayiheng and Zhou, Fan and Huang, Fei and Hu, Feng and Ge, Hao and Wei, Haoran and Lin, Huan and Tang, Jialong and Yang, Jian and Tu, Jianhong and Zhang, Jianwei and Yang, Jianxin and Yang, Jiaxi and Zhou, Jing and Zhou, Jingren and Lin, Junyang and Dang, Kai and Bao, Keqin and Yang, Kexin and Yu, Le and Deng, Lianghao and Li, Mei and Xue, Mingfeng and Li, Mingze and Zhang, Pei and Wang, Peng and Zhu, Qin and Men, Rui and Gao, Ruize and Liu, Shixuan and Luo, Shuang and Li, Tianhao and Tang, Tianyi and Yin, Wenbiao and Ren, Xingzhang and Wang, Xinyu and Zhang, Xinyu and Ren, Xuancheng and Fan, Yang and Su, Yang and Zhang, Yichang and Zhang, Yinger and Wan, Yu and Liu, Yuqiong and Wang, Zekun and Cui, Zeyu and Zhang, Zhenru and Zhou, Zhipeng and Qiu, Zihan},
  journal={arXiv preprint arXiv:2505.09388},
  year={2025}
}

@misc{openai_cua,
  title={Operator System Card},
  author={{OpenAI}},
  year={2025},
  url={https://cdn.openai.com/operator_system_card.pdf}
}

@misc{claude,
  title={Claude Opus 4.6 System Card},
  author={{Anthropic}},
  year={2026},
  url={https://www-cdn.anthropic.com/14e4fb01875d2a69f646fa5e574dea2b1c0ff7b5.pdf}
}

@article{pignatellisurvey,
  title={A Survey of Temporal Credit Assignment in Deep Reinforcement Learning},
  author={Pignatelli, Eduardo and Ferret, Johan and Geist, Matthieu and Mesnard, Thomas and van Hasselt, Hado and Toni, Laura},
  journal={Transactions on Machine Learning Research},
  year={2024}
}

@inproceedings{gritta2026process,
  title={Process evaluation for agentic systems},
  author={Gritta, Milan and Paul, Debjit and Li, Xiaoguang and Shang, Lifeng and Wang, Jun and Lampouras, Gerasimos},
  booktitle={Findings of the Association for Computational Linguistics: EACL 2026},
  pages={2678--2692},
  year={2026}
}

@inproceedings{luagentrewardbench,
  title={AgentRewardBench: Evaluating Automatic Evaluations of Web Agent Trajectories},
  author={L{\`u}, Xing Han and Kazemnejad, Amirhossein and Meade, Nicholas and Patel, Arkil and Shin, Dongchan and Zambrano, Alejandra and Stanczak, Karolina and Shaw, Peter and Pal, Christopher and Reddy, Siva},
  booktitle={Second Conference on Language Modeling},
  year={2025}
}

@article{androidenv,
  title={Androidenv: A reinforcement learning platform for android},
  author={Toyama, Daniel and Hamel, Philippe and Gergely, Anita and Comanici, Gheorghe and Glaese, Amelia and Ahmed, Zafarali and Jackson, Tyler and Mourad, Shibl and Precup, Doina},
  journal={arXiv preprint arXiv:2105.13231},
  year={2021}
}

@inproceedings{androidworld,
  title={AndroidWorld: A Dynamic Benchmarking Environment for Autonomous Agents},
  author={Rawles, Christopher and Clinckemaillie, Sarah and Chang, Yifan and Waltz, Jonathan and Lau, Gabrielle and Fair, Marybeth and Li, Alice and Bishop, William E and Li, Wei and Campbell-Ajala, Folawiyo and others},
  booktitle={The Thirteenth International Conference on Learning Representations},
  year={2025}
}

@article{realevals,
  title={Real: Benchmarking autonomous agents on deterministic simulations of real websites},
  author={Garg, Divyansh and VanWeelden, Shaun and Caples, Diego and Draguns, Andis and Ravi, Nikil and Putta, Pranav and Garg, Naman and Abraham, Tomas and Lara, Michael and Lopez, Federico and others},
  journal={arXiv preprint arXiv:2504.11543},
  year={2025}
}

@inproceedings{aitw,
  title={AndroidInTheWild: A Large-Scale Dataset For Android Device Control},
  author={Rawles, Christopher and Li, Alice and Rodriguez, Daniel and Riva, Oriana and Lillicrap, Timothy P},
  booktitle={Thirty-seventh Conference on Neural Information Processing Systems Datasets and Benchmarks Track},
  year={2023}
}

@inproceedings{alfworld,
  title={ALFWorld: Aligning Text and Embodied Environments for Interactive Learning},
  author={Shridhar, Mohit and Yuan, Xingdi and Cote, Marc-Alexandre and Bisk, Yonatan and Trischler, Adam and Hausknecht, Matthew},
  booktitle={International Conference on Learning Representations},
  year={2021}
}

@inproceedings{scienceworld,
  title={Scienceworld: Is your agent smarter than a 5th grader?},
  author={Wang, Ruoyao and Jansen, Peter and C{\^o}t{\'e}, Marc-Alexandre and Ammanabrolu, Prithviraj},
  booktitle={Proceedings of the 2022 Conference on Empirical Methods in Natural Language Processing},
  pages={11279--11298},
  year={2022}
}

@article{chen2021evaluating,
  title={Evaluating large language models trained on code},
  author={Chen, Mark and Tworek, Jerry and Jun, Heewoo and Yuan, Qiming and Pinto, Henrique Ponde De Oliveira and Kaplan, Jared and Edwards, Harri and Burda, Yuri and Joseph, Nicholas and Brockman, Greg and others},
  journal={arXiv preprint arXiv:2107.03374},
  year={2021}
}

@inproceedings{gu2024cruxeval,
  title={CRUXEval: a benchmark for code reasoning, understanding and execution},
  author={Gu, Alex and Rozi{\`e}re, Baptiste and Leather, Hugh and Solar-Lezama, Armando and Synnaeve, Gabriel and Wang, Sida I},
  booktitle={Proceedings of the 41st International Conference on Machine Learning},
  pages={16568--16621},
  year={2024}
}

@book{sutton1998reinforcement,
  title={Reinforcement learning: An introduction},
  author={Sutton, Richard S and Barto, Andrew G},
  volume={1},
  year={1998},
  publisher={MIT press Cambridge}
}

@inproceedings{andrychowicz2017hindsight,
  title={Hindsight experience replay},
  author={Andrychowicz, Marcin and Wolski, Filip and Ray, Alex and Schneider, Jonas and Fong, Rachel and Welinder, Peter and McGrew, Bob and Tobin, Josh and Pieter Abbeel, OpenAI and Zaremba, Wojciech},
  booktitle={Advances in neural information processing systems},
  year={2017}
}

@inproceedings{icarte2018using,
  title={Using reward machines for high-level task specification and decomposition in reinforcement learning},
  author={Icarte, Rodrigo Toro and Klassen, Toryn and Valenzano, Richard and McIlraith, Sheila},
  booktitle={International Conference on Machine Learning},
  pages={2107--2116},
  year={2018},
  organization={PMLR}
}

@article{chezelles2024browsergym,
  title={The browsergym ecosystem for web agent research},
  author={Chezelles, De and Le Sellier, Thibault and Shayegan, Sahar Omidi and Jang, Lawrence Keunho and L{\`u}, Xing Han and Yoran, Ori and Kong, Dehan and Xu, Frank F and Reddy, Siva and Cappart, Quentin and others},
  journal={arXiv preprint arXiv:2412.05467},
  year={2024}
}

@article{chung2025mllms,
  title={What MLLMs Learn about When they Learn about Multimodal Reasoning: Perception, Reasoning, or their Integration?},
  author={Chung, Jiwan and Joshi, Neel and Sharma, Pratyusha and Yu, Youngjae and Vineet, Vibhav},
  journal={arXiv preprint arXiv:2510.01719},
  year={2025}
}
\bibliographystyle{colm2026_conference}

\clearpage 

\appendix

% =============================================================================
%  Appendix — Reorganized v2
% =============================================================================

\clearpage
\mbox{}% ensure page starts

% ── Appendix title page with Table of Contents ──

\begin{center}
%\vspace*{2cm}
{\LARGE\bfseries Appendix}
\vspace{1.5cm}

\hrule

\vspace{0.8cm}
\end{center}

{
\hypersetup{linkcolor=black}
% Fixed: Changed \begin{large} to a localized switch
{\large\noindent\textbf{Table of Contents}}
\vspace{0.5em}

% Fixed: Changed \begin{normalsize} to a standard command
\normalsize
\noindent
\hyperref[sec:ax_limitations]{\textbf{A}\hspace{0.5em} Limitations} \dotfill \pageref{sec:ax_limitations} \\[4pt]
\hyperref[sec:ax_related]{\textbf{B}\hspace{0.5em} Related Work} \dotfill \pageref{sec:ax_related} \\[4pt]
\hyperref[sec:ax_impl]{\textbf{C}\hspace{0.5em} Implementation Details} \dotfill \pageref{sec:ax_impl} \\[1pt]
\hspace{1.2em} \hyperref[sec:ax_agent_config]{C.1\hspace{0.4em} Agent Configurations} \dotfill \pageref{sec:ax_agent_config} \\[1pt]
\hspace{1.2em} \hyperref[sec:ax_compute]{C.2\hspace{0.4em} Compute and Runtime Statistics} \dotfill \pageref{sec:ax_compute} \\[1pt]
\hspace{1.2em} \hyperref[sec:ax_eval_protocol]{C.3\hspace{0.4em} Evaluation Protocol} \dotfill \pageref{sec:ax_eval_protocol} \\[1pt]
\hspace{1.2em} \hyperref[subsec:ax_info_coverage]{C.4\hspace{0.4em} Informational Coverage} \dotfill \pageref{subsec:ax_info_coverage} \\[1pt]
\hspace{1.2em} \hyperref[sec:ax_reproducibility]{C.5\hspace{0.4em} Reproducibility} \dotfill \pageref{sec:ax_reproducibility} \\[1pt]
\hspace{1.2em} \hyperref[subsec:ax_llm_usage]{C.6\hspace{0.4em} LLM Usage} \dotfill \pageref{subsec:ax_llm_usage} \\[4pt]
\hyperref[sec:ax_benchmark]{\textbf{D}\hspace{0.5em} Benchmark Specification} \dotfill \pageref{sec:ax_benchmark} \\[1pt]
\hspace{1.2em} \hyperref[sec:ax_task_dist]{D.1\hspace{0.4em} Task Distribution} \dotfill \pageref{sec:ax_task_dist} \\[1pt]
\hspace{1.2em} \hyperref[sec:ax_site_screenshots]{D.2\hspace{0.4em} Site Screenshots} \dotfill \pageref{sec:ax_site_screenshots} \\[1pt]
\hspace{1.2em} \hyperref[sec:ax_site_specs]{D.3\hspace{0.4em} Per-Site MDP Specifications} \dotfill \pageref{sec:ax_site_specs} \\[4pt]
\hyperref[sec:ax_datagen]{\textbf{E}\hspace{0.5em} Data Generation Pipeline} \dotfill \pageref{sec:ax_datagen} \\[1pt]
\hspace{1.2em} \hyperref[sec:ax_site_construction]{E.1\hspace{0.4em} Site Construction Pipeline} \dotfill \pageref{sec:ax_site_construction} \\[1pt]
\hspace{1.2em} \hyperref[sec:ax_world_gen]{E.2\hspace{0.4em} World Generation} \dotfill \pageref{sec:ax_world_gen} \\[1pt]
\hspace{1.2em} \hyperref[sec:ax_oracle_gen]{E.3\hspace{0.4em} Oracle Trajectory Generation} \dotfill \pageref{sec:ax_oracle_gen} \\[4pt]
\hyperref[sec:ax_more_results]{\textbf{F}\hspace{0.5em} Extended Quantitative Results} \dotfill \pageref{sec:ax_more_results} \\[1pt]
\hyperref[subsec:ax_persite]{F.1\hspace{0.4em} Per-site performance decomposition} \dotfill \pageref{subsec:ax_persite} \\[1pt]
\hyperref[subsec:ax_skill_persite]{F.2\hspace{0.4em} Per-site skill invocation} \dotfill \pageref{subsec:ax_skill_persite} \\[1pt]
\hyperref[subsec:ax_skill_agg]{F.3\hspace{0.4em} Aggregate skill invocation} \dotfill \pageref{subsec:ax_skill_agg} \\[1pt]
\hyperref[subsec:ax_expl_complex]{F.4\hspace{0.4em} Exploration SR by problem complexities} \dotfill \pageref{subsec:ax_expl_complex} \\[1pt]
\hyperref[sec:ax_viewer]{\textbf{G}\hspace{0.5em} Artifact Viewer} \dotfill \pageref{sec:ax_viewer} \\[4pt]
\hyperref[sec:ax_examples]{\textbf{H}\hspace{0.5em} Data Examples} \dotfill \pageref{sec:ax_examples} \\[1pt]
\hspace{1.2em} \hyperref[sec:ax_template_catalog]{H.1\hspace{0.4em} Task Template Catalog} \dotfill \pageref{sec:ax_template_catalog} \\[1pt]
\hspace{1.2em} \hyperref[sec:ax_traj_examples]{H.2\hspace{0.4em} Example Trajectories: Agent vs.\ Oracle Trajectory} \dotfill \pageref{sec:ax_traj_examples} \\[1pt]
\hspace{1.2em} \hyperref[sec:ax_mdp_graphs]{H.3\hspace{0.4em} MDP Surface Transition Graphs} \dotfill \pageref{sec:ax_mdp_graphs} \\[4pt]
\hyperref[sec:ax_qualitative]{\textbf{I}\hspace{0.5em} Case Study} \dotfill \pageref{sec:ax_qualitative} \\[1pt]
\hspace{1.2em} \hyperref[sec:case_scale]{I.1\hspace{0.4em} Model Scale and Browsing Quality} \dotfill \pageref{sec:case_scale} \\[1pt]
\hspace{1.2em} \hyperref[sec:case_hardneg]{I.2\hspace{0.4em} Premature Commitment to Hard Negative} \dotfill \pageref{sec:case_hardneg} \\[1pt]
\hspace{1.2em} \hyperref[sec:case_thorough]{I.3\hspace{0.4em} Thorough Exploration Before Commitment} \dotfill \pageref{sec:case_thorough} \\[1pt]
\hspace{1.2em} \hyperref[sec:case_expexec]{I.4\hspace{0.4em} Exploration Success, Execution Failure} \dotfill \pageref{sec:case_expexec} \\[1pt]
\hspace{1.2em} \hyperref[sec:case_expfail_execsucc]{I.5\hspace{0.4em} Exploration Failure, Execution Success} \dotfill \pageref{sec:case_expfail_execsucc} \\[1pt]
\hspace{1.2em} \hyperref[sec:case_redundant]{I.6\hspace{0.4em} Redundant Process Despite Success} \dotfill \pageref{sec:case_redundant} \\[1pt]
\hspace{1.2em} \hyperref[sec:case_safepass]{I.7\hspace{0.4em} Failure by Safeguard, Not by Competence} \dotfill \pageref{sec:case_safepass}

\vspace{0.8cm}
\hrule
}

\clearpage

% =============================================================================
%  A. Limitations
% =============================================================================
\section{Limitations}
\label{sec:ax_limitations}

Our benchmark prioritizes controllability, reproducibility, and trajectory-level diagnosis. These design choices improve measurement precision, but they also introduce limitations in realism, coverage, and action scope.

\paragraph{Semantic abstraction.}
Each site is implemented as a deterministic MDP that captures the task-relevant interaction structure of its real-world counterpart. As a result, the benchmark does not capture some properties of live websites, including dynamic content updates, session-dependent state, asynchronous loading, personalized recommendations, and third-party integrations. Agent behavior in these self-hosted environments may therefore differ from behavior on live websites with noisier layouts and less predictable dynamics.

\paragraph{Domain coverage.}
The benchmark spans ten web domains, but it does not cover every application type. In particular, domains involving sensitive data handling, complex authorization flows, or rich media interaction are not included. Still, the covered domains are diverse, and the benchmark is broader in scope and substantially larger in task count than existing alternatives~\citep{webvoyager,webarena,mind2web}. We therefore view the remaining gap as a matter of incomplete coverage rather than narrow domain design.

\paragraph{Synthetic world data.}
World data is generated synthetically through seeded procedures. This improves reproducibility and allows controlled construction of hard negatives, but it may miss properties of organic real-world data such as long-tail popularity patterns, organically correlated attributes, or temporal drift.

\paragraph{Template-based tasks.}
Tasks are instantiated from structured templates with deterministic verification conditions. This does not capture the full ambiguity, underspecification, or open-endedness of real user requests. However, the loss is constrained by that the templates still span diverse task types and difficulty factors, while the resulting structure enables explicit control over complexity, scalable generation, and reliable aggregation across task families. This design is therefore necessary for systematic process-level evaluation at scale.

\paragraph{Skill invocation rather than skill success.}
Our skill-level analysis measures whether agents invoke the required action types, not whether each action succeeds in the agent's intended sense. A finer notion of per-action success would require reliable access to latent intent, which is not available from GUI traces alone and is difficult to infer faithfully from model-generated rationales. We therefore evaluate behavior in terms of the realized causal trajectory, which is the more robust and less assumption-dependent object of analysis.

\paragraph{Interaction modality.}
The current evaluation focuses on coordinate-based visual interaction, specifically clicking, typing, and scrolling. Other interaction modes, such as drag-and-drop, keyboard shortcuts, or direct URL navigation, are not included. This narrows the covered action space and may omit behaviors that matter in some real web settings.

\section{Related Work}
\label{sec:ax_related}

\paragraph{Web agent benchmarks.}
Web agent benchmarks now span a broad range of environment regimes, from tightly controlled synthetic interfaces such as MiniWoB++~\citep{miniwob} and WebShop~\citep{webshop}, to self-hosted realistic websites such as WebArena~\citep{webarena} and VisualWebArena~\citep{visualwebarena}, to online and live-web settings such as Mind2Web~\citep{mind2web}, WebVoyager~\citep{webvoyager}, Mind2Web-Live~\citep{mind2weblive}, Online-Mind2Web~\citep{onlinemind2web}, and AssistantBench~\citep{assistantbench}. Related benchmarks further broaden this landscape to enterprise workflows and general computer-use settings, including WorkArena~\citep{workarena}, BrowserGym~\citep{chezelles2024browsergym}, and OSWorld~\citep{osworld}. Despite this rapid progress in environment realism and scope, evaluation remains centered primarily on terminal outcomes such as task success, sometimes with partial-credit variants or key-node style intermediate checks. \DATANAME differs in focusing on built-in process instrumentation: each environment is co-developed with an explicit semantic MDP dual, enabling automatic process-level evaluation grounded in the environment's own state and transition structure.

\paragraph{Environment modeling and GUI-semantic bridging.}
Constructing faithful simulated environments for agent evaluation is a longstanding challenge. 
AndroidEnv~\citep{androidenv} and AndroidWorld~\citep{androidworld} wrap real mobile applications in RL-compatible interfaces, while OSWorld~\citep{osworld} provides full desktop VMs with scripted evaluators. These environments expose low-level interaction traces, but they do not provide an explicit semantic account of what each action reveals or accomplishes. A related line of work maps GUI events to higher-level action abstractions for training or evaluation~\citep{aitw,androidworld}, but usually does so post hoc through heuristic alignment rather than through the environment construction itself. In the web domain, WebArena~\citep{webarena}, VisualWebArena~\citep{visualwebarena}, and REAL Evals~\citep{realevals} move toward sandboxed and more reproducible evaluation, but their evaluation remains decoupled from an explicit website semantics and is largely limited to terminal success. Our semantic MDP dual instead co-develops the GUI and the semantic model so that all task-relevant rendering is generated from semantic state, making the correspondence explicit by construction.

\paragraph{Process-level and intermediate evaluation.}
The limitations of outcome-only evaluation are well recognized across adjacent areas. In mathematical reasoning, process supervision and process reward models explicitly score intermediate reasoning steps rather than only final answers~\citep{prm}. In code evaluation, benchmarks such as HumanEval assess functional correctness by executing generated programs against tests~\citep{chen2021evaluating}, while CRUXEval further targets code understanding and execution behavior rather than final output alone~\citep{gu2024cruxeval}. In sequential decision making, the broader credit assignment problem has long been recognized as a central challenge in reinforcement learning~\citep{pignatellisurvey,sutton1998reinforcement}, precisely because terminal rewards alone are often too sparse or too delayed to localize which decisions mattered. This has motivated methods that expose or reshape intermediate structure, including reward decomposition~\citep{icarte2018using} and hindsight-based relabeling~\citep{andrychowicz2017hindsight}.
\citet{chung2025mllms} also considered decomposition of multimodal reasoning capacities.
Our setting is analogous: terminal task success does not reveal whether a web agent failed to find the right target, failed to gather the decisive evidence, or failed only at the final action, which is why process-level evaluation is necessary.

 In web agent evaluation, however, most existing analyses remain at the task level: AgentBench~\citep{agentbench} and GAIA~\citep{gaia} report performance by task category, separating tasks from one another but not analyzing behavior within a trajectory. Recent work has begun to introduce within-task intermediate evaluation. AgentBoard~\citep{agentboard} measures progress rate via manually annotated subgoals, and Mind2Web-Live~\citep{mind2weblive} defines key nodes along reference trajectories. These approaches demonstrate that binary success discards substantial signal, but their reliance on static human-defined checkpoints limits both scalability and robustness, particularly on live websites where interface changes can invalidate annotations.

\paragraph{Automated trajectory evaluation.}
Trajectory-level evaluation that avoids manual annotation has been explored through several paradigms. LLM-as-judge approaches~\citep{luagentrewardbench, gritta2026process} use language models to score intermediate agent steps, offering flexibility but introducing stochastic evaluation noise and dependence on judge model capability. Rule-based evaluation in structured environments~\citep{alfworld, scienceworld} enables deterministic assessment but typically operates at the level of task completion predicates rather than fine-grained process analysis. Our framework differs from both: because the evaluation environment is built on an explicit semantic MDP, process metrics (including exploration success, skill invocation, and trajectory bifurcation analysis) are derived deterministically from the MDP trace without requiring either human annotation or model-based judgment.

\section{Implementation Details}
\label{sec:ax_impl}

\subsection{Agent Configurations}
\label{sec:ax_agent_config}

\begin{table}[H]
\centering
\resizebox{\textwidth}{!}{%
\begin{tabular}{llllcccc}
\toprule
\textbf{Agent} & \textbf{Model} & \textbf{Backend} & \textbf{Input} & \textbf{Action Space} & \textbf{Max Imgs} & \textbf{Temp.} & \textbf{Max Tokens} \\
\midrule
Fara & microsoft/Fara-7B & vLLM & Screenshot & Function call (coordinate) & 3 & 0.0 & 800 \\
UI-TARS & ByteDance-Seed/UI-TARS-1.5-7B & vLLM & Screenshot & Thought + Action (coordinate) & 5 & 0.0 & 2048 \\
GUI-Owl & mPLUG/GUI-Owl-1.5-8B & vLLM & Screenshot & Tool call (coordinate) & 4 & 0.0 & 2048 \\
Qwen3.5 & Qwen/Qwen3.5-122B-FP8 & vLLM & Screenshot & Thought + Action (coordinate) & 5 & 0.7 & 2048 \\
OpenAI CUA & GPT-5.4 & OpenAI API & Screenshot & Built-in computer tool (coordinate) & -- & -- & -- \\
Claude CUA & claude-sonnet-4-6 & Anthropic API & Screenshot + DOM & Tool call (ref + coordinate) & 3 & -- & -- \\
\bottomrule
\end{tabular}
}
\caption{Agent configurations. \textbf{Input}: what the agent receives each turn. \textbf{Action space}: how the agent specifies actions. \textbf{Max imgs}: number of recent screenshots retained in context. Claude CUA follows Anthropic's DOM-augmented browser harness (ref-based targeting with coordinate fallback), whereas all other agents operate purely from pixels.}
\label{tab:ax_agent_config}
\end{table}

All agents interact with the browser through a shared viewport of $1440 \times 900$ pixels. Each episode is capped at a maximum of 50 turns; if the agent has not completed the task within this budget, the episode is terminated and scored as a failure. Between consecutive agent actions, a brief blocking period is enforced to allow page transitions and rendering to complete before the next screenshot is captured.

We evaluate the following agents in \DATANAME:

\begin{itemize}[leftmargin=1.5em,itemsep=2pt]

    \item \textbf{UI-TARS}~\cite{uitars}: A vision--language model for GUI interaction developed by ByteDance.\footnote{\url{https://github.com/bytedance/ui-tars}}
    \item \textbf{Fara}~\cite{fara}: A 7B web agent from Microsoft Research, fine-tuned on Qwen2.5-VL.\footnote{\url{https://github.com/microsoft/fara}}
    \item \textbf{GUI-Owl}~\cite{guiowl}: A multimodal agent from Alibaba mPLUG. We use the 8B-model. \footnote{\url{https://github.com/x-plug/mobileagent}}  %  and 32B-model.
    \item \textbf{Qwen3.5}~\cite{qwen3}: Alibaba's Qwen3.5-122B model, quantized to FP8.\footnote{\url{https://github.com/QwenLM/Qwen-Agent}}
    \item \textbf{OpenAI CUA}~\cite{openai_cua}: OpenAI's Computer-Use Agent, accessed via the OpenAI API. We implemented the CUA-agent following the official github repository. \footnote{\url{https://github.com/openai/openai-agents-python}}
    \item \textbf{Claude CUA}: Anthropic's computer-use agent (claude-sonnet-4-6), accessed via the Anthropic API. Following Anthropic's browser-use reference implementation, it receives each screenshot together with a list of interactive DOM elements and targets elements by reference id, falling back to coordinates when needed.\footnote{\url{https://github.com/anthropics/anthropic-quickstarts}}
\end{itemize}

\subsection{Compute and Runtime Statistics}
\label{sec:ax_compute}

All GPU-based agents (Fara, UI-TARS, GUI-Owl, and Qwen3.5-122B) are served via vLLM on a single node with 4$\times$ NVIDIA H200 GPUs (140\,GB each). The full benchmark evaluation across all GPU models completes in approximately 3--4 days of wall-clock time. The closed-source agents (OpenAI CUA and Claude CUA) use their respective vendor APIs and each complete within a day with 8 parallel workers.

\subsection{Evaluation Protocol}
\label{sec:ax_eval_protocol}

Each task episode proceeds as follows. First, a headless Chromium browser instance is launched and initialized with the task's world data and starting state. The agent then enters a turn-based loop: at each turn, the runner captures a screenshot of the current browser viewport, constructs an observation (including the screenshot, the current page URL, and history of previous conversation), and passes it to the agent. The agent returns an action (e.g., a click at a coordinate, a text input, or a scroll) which the runner executes in the browser. This loop continues until the agent signals task completion, the maximum turn limit of 50 is reached, or the agent reports the task as infeasible.

Upon episode termination, the verifier evaluates the stored semantic MDP trajectory against the task's verifier conditions. These conditions check the application's terminal semantic state for example, whether a specific item has been added to a cart, a message has been starred, or a booking has been confirmed with the correct parameters. An episode is scored as successful only if all verifier conditions are satisfied. 

Additionally, we observed that some models (e.g., Fara) consistently fail to execute the final commit action even after correctly identifying the target entity. To ensure fair comparison, we adopt a \emph{safe pass} policy: if the agent reaches the critical decision point, i.e., it has navigated to the correct entity and gathered sufficient information to act, but does not issue the final commit, the episode is still counted as a success for the terminal success rate. This prevents penalizing models whose failure is limited to the mechanical execution of a single concluding action rather than a deficiency in exploration or reasoning. All terminal success rates reported in this paper are measured under the safe pass criterion.

\subsection{Information Coverage}
\label{subsec:ax_info_coverage}

Information coverage measures how much task-relevant evidence an agent has gathered by the time it commits. It is a graded metric of evidence acquisition, not a binary test of whether the agent observed everything that could have been useful. For each task, the benchmark derives a set of information constraints
\(
\mathcal{C} = \{c_1, \dots, c_K\}
\)
from the task's information gap, template type, and reasoning type using per-site declarative rules. This set is best understood as a conservative upper bound on useful information: it includes fields, entities, and comparisons that may support the decision, and agents can succeed while observing less (e.g., by narrowing candidates through search). At the same time, the rules are calibrated against the ground-truth trajectories -- the oracle solution observes every constraint before committing -- so the set never demands information that a correct solution would not gather. A small set of tasks (226 of 1{,}800) satisfies all constraints from the initial observation alone; coverage is non-discriminative on this subset by construction.

Each site defines a visibility specification that maps every interface surface \(\sigma\) to the set of item attributes rendered on that surface. Some attributes are visible on list-level surfaces, while others become visible only after opening a detail page. A detail-only constraint can also be satisfied through a search operation: when the agent issues a query containing the task's discriminating value and the target remains in the results, the fields that the site's search function scans count as verified (the match is disjunctive over the scanned fields, so this credit is conservative and value-guarded). As the agent follows a semantic trajectory
\(
\tau = (s_0, a_0, \dots, s_T),
\)
the benchmark tracks cumulatively which task-relevant constraints have become satisfied. Coverage at step \(t\) is then defined as
\begin{equation}
\text{COV}_t
=
\frac{|\{c \in \mathcal{C} : \mathrm{sat}(c, t)\}|}{|\mathcal{C}|},
\end{equation}
where \(\mathrm{sat}(c,t)\) indicates whether constraint \(c\) has been satisfied by the information observed up to step \(t\).

We report coverage at the commit step,
\(
t_{\mathrm{commit}},
\)
defined as the first step at which the agent executes a terminal action such as \texttt{PlaceOrder} or \texttt{done}. The reported metric,
\(
\text{COV}_{t_{\mathrm{commit}}},
\)
therefore measures how much of this benchmark-defined task-relevant evidence the agent had gathered when it chose to act.

\subsection{Reproducibility}
\label{sec:ax_reproducibility}

We take the following steps to ensure reproducibility:

\paragraph{Code and data release.}
We will publicly release the complete benchmark codebase, including all 10 site implementations, the task generation pipeline, the evaluation runner, and the metric computation scripts. All 1,800 task definitions (with world seeds, instructions, oracle trajectories, and verifier conditions) will be included in the release.

\paragraph{Environment determinism.}
\begin{itemize}[leftmargin=1.5em,itemsep=2pt]
    \item \textbf{Fixed world seeds.} Each task is paired with a deterministic random seed that fully determines the world data (user profiles, item catalogs, message contents, etc.), ensuring identical evaluation conditions across runs.
    \item \textbf{Deterministic MDPs.} All site implementations are pure deterministic MDPs: the same state--action pair always produces the same next state and observation, eliminating environmental stochasticity.
\end{itemize}

\paragraph{Agent hyperparameters.}
All agent hyperparameters are reported in \Cref{tab:ax_agent_config} (Section~B.1). Key settings:
\begin{itemize}[leftmargin=1.5em,itemsep=2pt]
    \item \textbf{Greedy decoding.} All open-weight agents (Fara, UI-TARS, GUI-Owl) are served with temperature 0, ensuring deterministic outputs given identical inputs. Qwen3.5 uses temperature 0.7, top-$p$ 0.8, top-$k$ 20 to follow official implementation.
    \item \textbf{Episode budget.} All agents are given a maximum of 50 GUI turns per task.
    \item \textbf{Viewport.} All agents share a $1440 \times 900$ browser viewport.

\end{itemize}

\paragraph{Non-determinism.}
For the closed-source OpenAI CUA and Claude CUA agents, exact reproducibility cannot be guaranteed due to potential server-side non-determinism in the API; however, we observe low variance across repeated runs.

\subsection{LLM Usage}
\label{subsec:ax_llm_usage}

We use LLMs in three aspects of this work. First, in the benchmark construction pipeline, we use Claude Opus 4.6 as a coding subroutine to help convert recorded website interaction traces into semantic MDP specifications and to build the executable website implementations from those specifications, as described in \Cref{sec:ax_datagen}. This process involved substantial manual validation, editing, and re-iteration between steps; the LLM served as an accelerator within a human-driven development loop rather than as an autonomous construction agent. Second, the web agents evaluated in our experiments (\Cref{sec:experiments}) are themselves LLM-based systems; their use constitutes the object of study rather than a methodological choice. Third, LLMs assisted with result visualization by providing initial seeds for plotting code and performing instruction-based edits; all figures were subsequently hand-edited in both code and output. In all three cases, the authors reviewed and verified the outputs. No LLM was used to originate research ideas or generate experimental data. LLMs provided assistance with prose editing during drafting.

%\paragraph{Software dependencies.}
%All Python packages, the Playwright browser version, and the vLLM serving version are pinned to exact versions in the released codebase. The benchmark runs on Python 3.11 with Chromium managed by Playwright.

% =============================================================================
%  C. Benchmark Specification
% =============================================================================

\section{Benchmark Specification}
\label{sec:ax_benchmark}

\subsection{Task Distribution}
\label{sec:ax_task_dist}

\begin{figure}[H]
\centering
\includegraphics[width=\textwidth]{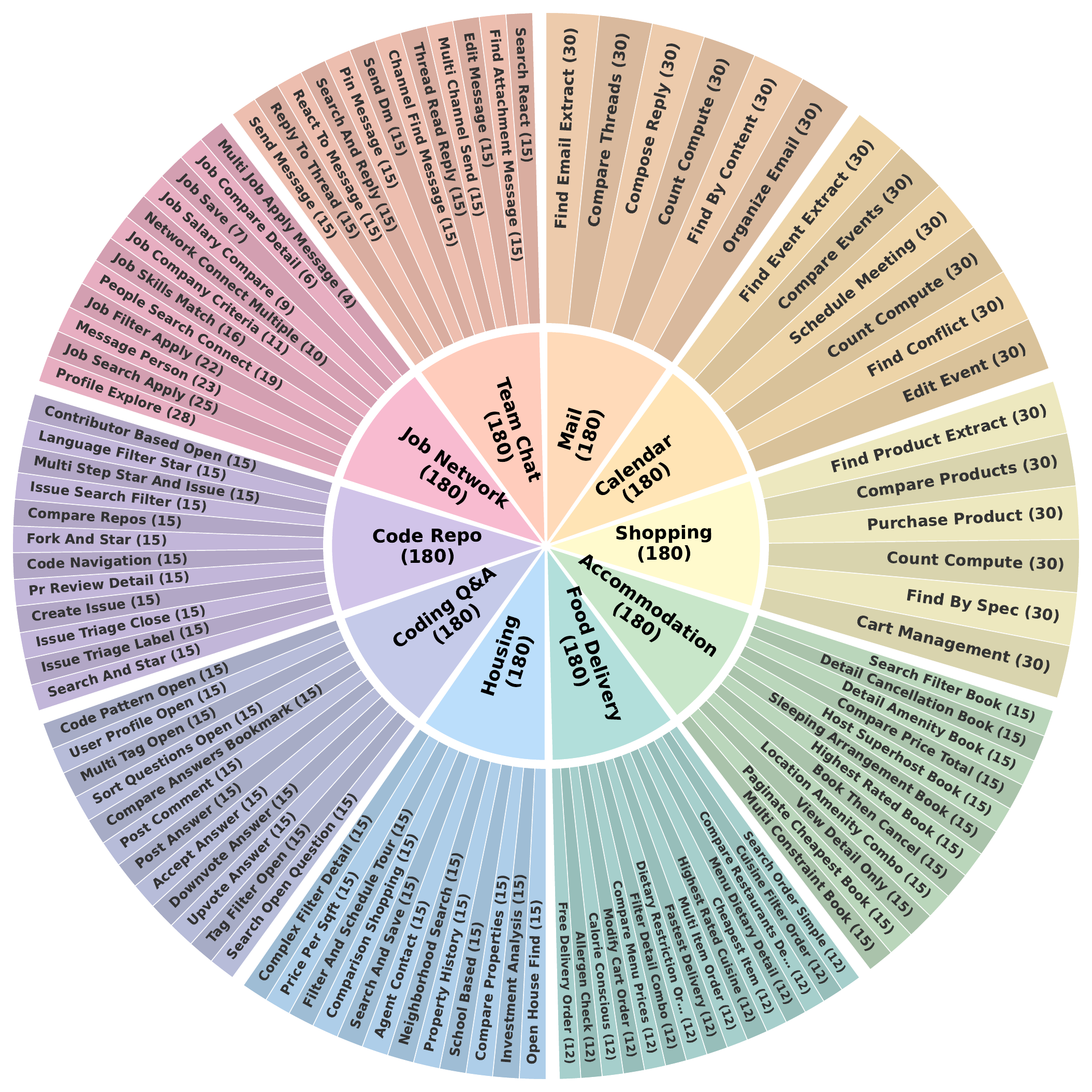}
\caption{\textbf{\DATANAME task distribution.} The inner ring shows 10 websites (180 tasks each). The outer ring shows task templates with instance counts. Templates range from 4 to 30 instances per template, reflecting the diversity of interaction patterns within each site.}
\label{fig:ax_sunburst}
\end{figure}

\Cref{fig:ax_sunburst} visualizes the hierarchical task distribution across the ten sites in \DATANAME. Each site contributes 180 tasks instantiated from multiple templates, yielding a total of 1{,}800 tasks spanning diverse interaction patterns and difficulty levels.

\subsection{Site Screenshots}
\label{sec:ax_site_screenshots}

\begin{figure}[H]
\centering
\includegraphics[width=\textwidth]{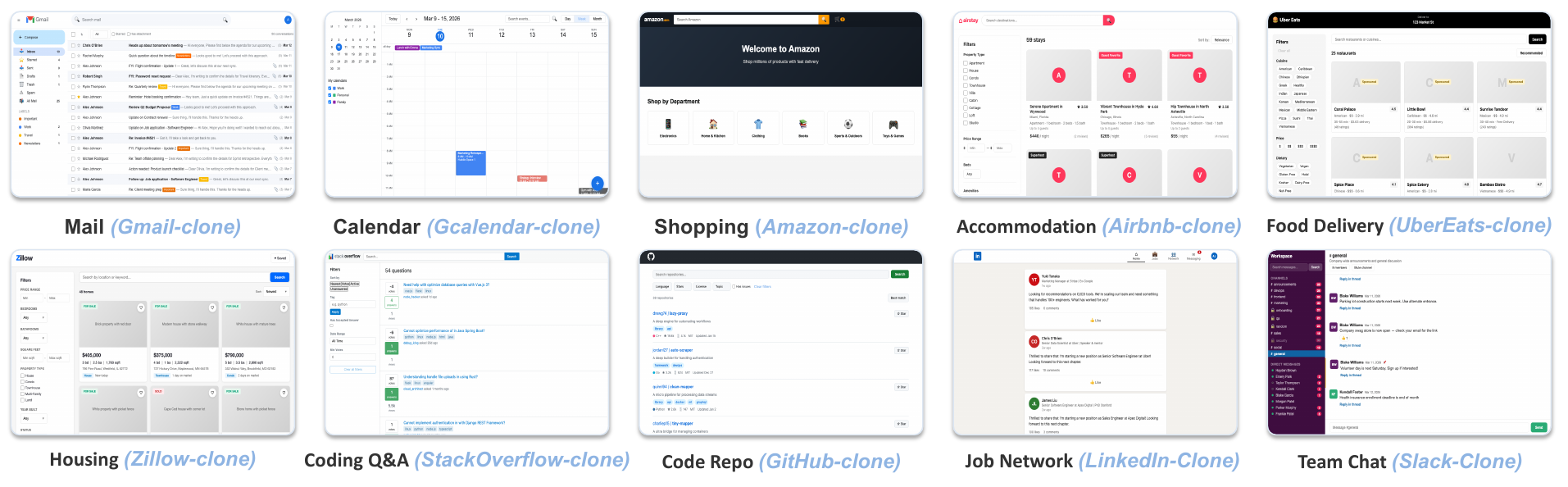}
\caption{Screenshots of all 10 \DATANAME websites. \textbf{Top row}: Mail, Calendar, Shopping, Accommodation, Food Delivery. \textbf{Bottom row}: Housing, Coding Q\&A, Code Repository, Job Network, and Team Chat. Each site is a standalone HTML/CSS/JS application with full MDP instrumentation.}
\label{fig:ax_sites}
\end{figure}

\Cref{fig:ax_sites} shows representative screenshots from each of the ten sites in \DATANAME. The sites span a broad range of web application categories--from e-mail and calendaring to e-commerce, social networking, and developer tools--each with distinct visual layouts, navigation patterns, and interaction paradigms. Each site's visual layout and interaction flow is modeled after a representative real-world web application in the same domain.

\subsection{Per-Site MDP Specifications}
\label{sec:ax_site_specs}

\paragraph{State decomposition.}
Each site in \DATANAME is modeled as a deterministic MDP whose state is decomposed into a set of typed variables. These variables capture the current navigation surface (e.g., inbox, search results, item detail), UI component states (e.g., which dropdown is open, current scroll position, active filters), and application-level data (e.g., selected items, form field contents, user preferences). \Cref{tab:ax_data_numbers} summarizes the number of surfaces, state variables, actions, and tasks for each site.

\paragraph{Visibility partitioning.}
Not all state variables are visible on every surface. \Cref{tab:ax_visibility} specifies the visibility partition: for each site, it lists which state variables are observable on each surface. This partition determines what information is available to the agent at any given point in the episode and is a key factor in task difficulty: tasks that require reasoning about state variables only visible on other surfaces necessitate multi-step navigation.

\paragraph{Skill taxonomy.}
\Cref{tab:ax_skill_categories} defines the five skill categories used throughout the benchmark. \Cref{tab:ax_skills} then lists the full set of skills for each site, organized by category. Each task is labeled with the set of skills it exercises, enabling fine-grained analysis of agent capabilities across navigation, search, inspection, and execution dimensions. The four exploration categories are global action-type sets, whereas \emph{commit} is task-relative: each task's commit is the final task-completing step of its oracle trajectory, performed on the task's target entity.

\begin{table}[H]
\resizebox{\textwidth}{!}{%
\begin{tabular}{lccccccl}
\toprule
\textbf{Site} & \textbf{Domain} & \textbf{Surfaces} & \textbf{Actions} & \textbf{Templates} & \textbf{Tasks} & \textbf{Card / Detail Fields} & \textbf{Decision Object} \\
\midrule
Mail           & Productivity  & 3 & 36 & 6  & 180 & 10 / 9  & Thread \\
Calendar       & Productivity  & 6 & 14 & 6  & 180 & 8 / 6   & Event \\
Team Chat      & Productivity  & 3 & 24 & 12 & 180 & 6 / 5   & Message \\
Shopping       & E-Commerce    & 6 & 23 & 6  & 180 & 12 / 12 & Product \\
Food Delivery  & E-Commerce    & 6 & 24 & 15 & 180 & 8 / 7   & Restaurant / Item \\
Accommodation  & Discovery     & 4 & 16 & 12 & 180 & 9 / 8   & Listing \\
Housing        & Discovery     & 4 & 16 & 12 & 180 & 8 / 10  & Property \\
Coding Q\&A   & Information   & 4 & 18 & 12 & 180 & 7 / 6   & Question \\
Code Repo      & Collaboration & 6 & 32 & 12 & 180 & 9 / 8   & Repository / Issue \\
Job Network    & Social        & 6 & 26 & 12 & 180 & 7 / 9   & Job / Profile \\
\midrule
\textbf{Total} & -- & 48 & 229 & 105 & \textbf{1,800} & 84 / 80 & -- \\
\bottomrule
\end{tabular}
}
\centering
\caption{Per-site MDP statistics. \textbf{Surfaces}: distinct UI views (pages, modals). \textbf{Actions}: typed semantic action count. \textbf{Templates}: task template count. \textbf{Tasks}: total task instances. \textbf{Card fields}: attributes visible on list views. \textbf{Detail fields}: attributes requiring detail-page navigation.}
\label{tab:ax_data_numbers}
\end{table}

\begin{table}[H]
\centering

\resizebox{\textwidth}{!}{%
\begin{tabular}{llllcc}
\toprule
\textbf{Site} & \textbf{Card Surfaces} & \textbf{Detail Surfaces} & \textbf{Commit Surfaces} & $|\mathcal{V}_{\text{card}}|$ & $|\mathcal{V}_{\text{detail}}|$ \\
\midrule
Mail & ThreadList & ThreadView & ComposeModal, ThreadList, ThreadView & 10 & 9 \\
Calendar & WeekView, DayView, MonthView & EventDetail & EventEditor & 8 & 6 \\
Team Chat & ChannelView, SearchResults & ChannelView (thread) & ChannelView, DirectMessageView & 6 & 5 \\
Shopping & SearchResults & ProductDetail & Checkout, OrderConfirmation & 12 & 12 \\
Food Delivery & RestaurantList & RestaurantDetail, MenuItemDetail & Checkout, OrderStatus & 8 & 7 \\
Accommodation & SearchResults & ListingDetail & Checkout, Reservations & 9 & 8 \\
Housing & SearchResults & PropertyDetail & PropertyDetail, SavedHomes & 8 & 10 \\
Coding Q\&A & QuestionList & QuestionDetail & QuestionDetail & 7 & 6 \\
Code Repo & RepoList, IssueList, PullRequestList & RepoDetail, IssueDetail, PRDetail & IssueDetail, RepoDetail & 9 & 8 \\
Job Network & JobList, ConnectionList & JobDetail, ProfileView & JobDetail, Messaging & 7 & 9 \\
\bottomrule
\end{tabular}
}
\caption{Visibility partition for each site. $|\mathcal{V}_{\text{card}}|$: attributes visible on list surfaces. $|\mathcal{V}_{\text{detail}}|$: attributes visible only on detail surfaces. Surfaces are categorized as \textbf{Card} (list views), \textbf{Detail} (entity detail pages), and \textbf{Commit} (surfaces where task-completing actions are available).}
\label{tab:ax_visibility}
\end{table}

\begin{table}[H]
\centering
\small
\begin{tabular}{l p{12cm}}
\toprule
\textbf{Category} & \textbf{Description} \\
\midrule
Navigate & Actions that move between surfaces or manage the current view without modifying application state. Includes pagination, folder/tab switching, back navigation, closing detail views, and temporal navigation (e.g., next/previous week). \\[4pt]
Search & Actions that issue or clear search queries to change the visible entity set. Includes keyword search, people search, repository search, and clearing search context. \\[4pt]
Filter & Actions that narrow or reorder the visible entities within the current result set. Includes applying categorical or range filters, sorting, and clearing filters. \\[4pt]
Inspect & Actions that reveal additional information about an entity without modifying application state. Includes opening detail views, expanding hidden content, viewing profiles, and selecting entities for closer examination. \\[4pt]
Commit & The task-completing action: the final step of the task's oracle trajectory, performed on the target entity---e.g., placing the order, sending the reply, starring the target thread, or, for read-only tasks, opening the target item. State-modifying actions that are not the task's commit (e.g., adding an item to the cart before placing the order, or filling a form before sending) form the task's \emph{pre-commit sequence}; in trajectory analyses these steps are merged into the step that follows them rather than counted as a separate category. \\
\bottomrule
\end{tabular}
\caption{Skill category definitions. The four exploration categories (Navigate, Search, Filter, Inspect) are global action-type sets; \emph{Commit} is task-relative, defined per task as the final step of its oracle trajectory.}
\label{tab:ax_skill_categories}
\end{table}

\begin{table}[H]
\centering
\resizebox{\textwidth}{!}{%
\begin{tabular}{l l p{10.5cm}}
\toprule
\textbf{Site} & \textbf{Category} & \textbf{Skills} \\
\midrule
\multirow{5}{*}{Mail}
  & Navigate & folder\_navigation, pagination, backtrack\_navigation \\
  & Search & search\_refinement \\
  & Filter & filter\_application \\
  & Inspect & thread\_inspection, message\_expansion \\
  & Commit & compose\_setup, compose\_field\_entry, email\_management, bulk\_management, send\_commit \\
\midrule
\multirow{5}{*}{Calendar}
  & Navigate & temporal\_navigation, view\_switching, backtrack\_navigation \\
  & Search & search\_query \\
  & Filter & calendar\_filtering \\
  & Inspect & event\_inspection, information\_extraction, comparison, conflict\_detection \\
  & Commit & event\_creation, event\_editing, field\_entry, commit\_timing \\
\midrule
\multirow{5}{*}{Shopping}
  & Navigate & pagination \\
  & Search & query\_formulation \\
  & Filter & department\_filtering, category\_filtering, brand\_filtering, price\_filtering, attribute\_filtering, sort\_usage \\
  & Inspect & product\_inspection, review\_inspection, comparison \\
  & Commit & option\_selection, cart\_management, checkout\_flow, commit\_timing \\
\midrule
\multirow{5}{*}{Accommodation}
  & Navigate & pagination, search\_return \\
  & Search & query\_formulation \\
  & Filter & property\_type\_filtering, price\_filtering, beds\_filtering, amenity\_filtering, sort\_usage \\
  & Inspect & listing\_inspection, amenity\_expansion, comparison \\
  & Commit & booking\_initiation, booking\_confirmation, booking\_cancellation, checkout\_cancellation \\
\midrule
\multirow{5}{*}{Food Delivery}
  & Navigate & pagination \\
  & Search & query\_formulation \\
  & Filter & cuisine\_filtering, price\_filtering, dietary\_filtering, attribute\_filtering, sort\_usage \\
  & Inspect & restaurant\_inspection, menu\_item\_inspection, comparison \\
  & Commit & customization\_selection, cart\_management, checkout\_flow, commit\_timing \\
\bottomrule
\end{tabular}
}
\caption{Skill taxonomy per site (Part 1 of 2). Skills are organized into five categories: \textbf{Navigate} (surface traversal), \textbf{Search} (query formulation), \textbf{Filter} (attribute filtering and sorting), \textbf{Inspect} (entity viewing), and \textbf{Commit} (task-completing actions).}
\label{tab:ax_skills}
\end{table}

\clearpage

\begin{table}[H]
\centering
\resizebox{\textwidth}{!}{%
\begin{tabular}{l l p{10.5cm}}
\toprule
\textbf{Site} & \textbf{Category} & \textbf{Skills} \\
\midrule
\multirow{5}{*}{Housing}
  & Navigate & pagination \\
  & Search & query\_formulation \\
  & Filter & price\_filtering, beds\_filtering, baths\_filtering, sqft\_filtering, type\_filtering, attribute\_filtering, sort\_usage \\
  & Inspect & property\_inspection, agent\_inspection, comparison \\
  & Commit & save\_management, tour\_scheduling, agent\_contact, saved\_homes\_review \\
\midrule
\multirow{5}{*}{Coding Q\&A}
  & Navigate & back\_navigation, pagination \\
  & Search & query\_formulation \\
  & Filter & tag\_filtering, acceptance\_filtering, date\_filtering, vote\_filtering, sort\_usage, answer\_sort\_usage, filter\_reset \\
  & Inspect & question\_inspection, user\_inspection, tag\_inspection \\
  & Commit & voting, answer\_posting, commenting, answer\_acceptance, bookmarking \\
\midrule
\multirow{5}{*}{Code Repo}
  & Navigate & navigate\_to\_issues, navigate\_to\_prs, back\_navigation, pagination \\
  & Search & repo\_search \\
  & Filter & language\_filtering, stars\_filtering, license\_filtering, topic\_filtering, repo\_attribute\_filtering, repo\_sort\_usage, issue\_state\_filtering, issue\_label\_filtering, issue\_author\_filtering, issue\_sort\_usage, pr\_filtering, pr\_sort\_usage \\
  & Inspect & repo\_inspection, issue\_inspection, pr\_inspection \\
  & Commit & issue\_creation, issue\_commenting, issue\_assignment, issue\_labeling, issue\_state\_change, star\_management, commit\_timing \\
\midrule
\multirow{5}{*}{Job Network}
  & Navigate & tab\_navigation, pagination \\
  & Search & job\_query\_formulation, people\_query\_formulation \\
  & Filter & location\_filtering, job\_type\_filtering, remote\_filtering, salary\_filtering, experience\_filtering, sort\_usage, filter\_clearing \\
  & Inspect & job\_inspection, profile\_inspection, job\_comparison, profile\_comparison \\
  & Commit & job\_application, connection\_request, messaging, job\_save\_action, conversation\_management, feed\_interaction \\
\midrule
\multirow{5}{*}{Team Chat}
  & Navigate & channel\_navigation, dm\_navigation, scroll\_navigation, search\_result\_navigation \\
  & Search & search\_query, search\_clear \\
  & Filter & --- \\
  & Inspect & thread\_inspection, member\_inspection, thread\_close, member\_close \\
  & Commit & send\_channel\_message, send\_thread\_reply, send\_dm, message\_reaction, message\_pin, message\_edit, message\_delete, channel\_mute, commit\_timing \\
\bottomrule
\end{tabular}
}
\caption{Skill taxonomy per site (Part 2 of 2). See \Cref{tab:ax_skill_categories} for category definitions.}
\label{tab:ax_skills_2}
\end{table}

% MDP Surface Transition Graphs moved to Section F.3 (Examples)

% =============================================================================
%  D. Data Generation Pipeline
% =============================================================================
\section{Data Generation Pipeline}
\label{sec:ax_datagen}

\subsection{Site Construction Pipeline}
\label{sec:ax_site_construction}
\begin{figure}[H]
\centering
\includegraphics[width=\linewidth]{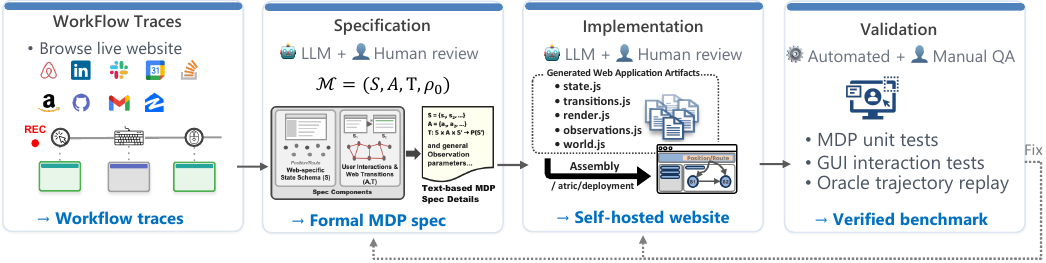}
\caption{\textbf{Overview of the website pipeline.} }
\label{fig:pipeline}
\end{figure}

Each site in \DATANAME is developed through an iterative human-in-the-loop pipeline. An LLM coding agent (Claude Opus 4.6~\cite{claude}) carries out much of the implementation at each stage, while all intermediate outputs are manually reviewed and refined before the next stage. The process is iterative rather than strictly linear: when downstream validation reveals issues, earlier stages are revisited and updated accordingly.

\paragraph{Phase 1: Grounded exploration.}
We begin by interacting with representative user workflows on the live target website, including searching, filtering, inspecting items, and completing final actions, while recording user actions and screenshots at each step. These traces provide grounded evidence for the subsequent specification stage. All observations are derived solely from user-visible interaction, namely the screenshot and user action at each step, without relying on backend code, API traffic, or other non-GUI signals.

\paragraph{Phase 2: Specification.}
The coding agent generalizes the recorded traces into a formal semantic MDP specification through the following staged procedure. Each stage must satisfy its validation criteria before the pipeline proceeds.

The state schema refines the main text's positional/informational decomposition: interface, query, and session state together form the positional state, while revealed state corresponds to the informational state. 

\Cref{alg:spec_construction} details the steps for generating MDP specification from the Workflow-traces of Phase 1.

\paragraph{Phase 3: Website implementation.}
The specification is then implemented as a self-contained single-page web application whose visible GUI is a rendered view of the underlying semantic MDP state, with minor exceptions on non-semantic actions:
\begin{itemize}[nosep,leftmargin=*]
\item \texttt{state.js}: State schema and \texttt{createInitialState(world)} initializer.
\item \texttt{transitions.js}: Pure transition function \texttt{transition(state, action, world) \textrightarrow \{state, error?\}}, with one case per typed action and precondition checking.
\item \texttt{observations.js}: Pure observation function \texttt{observe(state, world) \textrightarrow observation}, resolving visible entities, applying surface-specific visibility rules, and computing derived values.
\item \texttt{world.js}: Seeded pseudo-random generator \texttt{generateWorld(seed)} producing a complete entity catalog with cross-entity consistency enforced by construction.
\item \texttt{render.js}: DOM rendering as a pure projection of the observation. Every interactive element is bound to \texttt{dispatch(action)} calls via standardized \texttt{data-test-id} attributes.
\end{itemize}
Thus, GUI interactions are executed at the coordinate level, but every task-relevant interaction is mapped to a typed semantic action through the dispatch layer.
The entry page wires these modules into a dispatch loop: each user interaction calls \texttt{dispatch}, which runs the transition, recomputes the observation, and re-renders the page. Every dispatch also persists the current state and the full action trajectory to \texttt{localStorage}.

\begin{algorithm}[H]
\caption{\textbf{Specification construction from grounded traces}}
\label{alg:spec_construction}
\begin{algorithmic}[1]
\Require Workflow traces from Phase 1
\Ensure Structured MDP specification

\State Identify domain entities with typed attributes and unique identifiers
\State Define relations between entities with explicit cardinalities
\State Map surfaces, including visible entities, displayed fields, and available actions
\State Construct the navigation graph and verify surface reachability
\State Define the state schema:
\Statex \hspace{\algorithmicindent} interface state, query state, session state, revealed state
\State Specify the observation function from state to visible fields and available actions
\State Define each action with parameters, preconditions, state updates, reveals, and failure cases
\State Specify filtering, sorting, pagination, and the deterministic query function
\State Define flow guards for sequential dependencies and stage-locked fields
\State Specify all derived display values as deterministic functions of state and world data
\State Define the rendering contract, including layout templates, rendering rules, DOM bindings, and style parameters
\State Define the world generator, including field distributions, consistency constraints, variation axes, and invariants

\State Run specification verification:
\Statex \hspace{\algorithmicindent} coverage matrix
\Statex \hspace{\algorithmicindent} reachability
\Statex \hspace{\algorithmicindent} state liveness
\Statex \hspace{\algorithmicindent} constraint binding
\Statex \hspace{\algorithmicindent} observation sufficiency
\Statex \hspace{\algorithmicindent} world generator consistency

\If{any check fails}
    \State Return to the relevant earlier stage and revise
\EndIf

\State Compile the final structured specification for the implementation stage
\end{algorithmic}
\end{algorithm}

\paragraph{Multi-level validation.}
Each site is validated at four complementary levels:

\begin{enumerate}[nosep,leftmargin=*]
\item \textbf{MDP unit tests} (\texttt{validate.js}). These tests validate the semantic model in isolation, including:
\begin{itemize}[nosep,leftmargin=1.5em]
    \item correct execution of happy-path trajectories,
    \item rejection of invalid actions,
    \item surface-specific observation visibility,
    \item correctness of derived values,
    \item navigation and back-stack behavior,
    \item filter, sort, and search behavior, and
    \item deterministic but varied world generation across seeds.
\end{itemize}

\item \textbf{GUI interaction validation} (\texttt{validate\_interactions.py}). Playwright-based browser tests verify:
\begin{itemize}[nosep,leftmargin=1.5em]
    \item discovery of all instrumented \texttt{data-test-id} elements,
    \item coordinate-level interactability without occlusion,
    \item execution of interactions through raw mouse and keyboard actions,
    \item agreement between GUI interactions and expected MDP state transitions, and
    \item absence of unsupported controls such as native \texttt{<select>} elements.
\end{itemize}

\item \textbf{GUI trajectory replay} (\texttt{verify\_gui.py}). End-to-end replay of every oracle trajectory in headless Chromium checks:
\begin{itemize}[nosep,leftmargin=1.5em]
    \item that every semantic action produces a real state change,
    \item that the full oracle trajectory executes without error, and
    \item that the final state satisfies the task verifier.
\end{itemize}

\item \textbf{Manual QA iteration}. Early agent runs are manually inspected to identify residual issues in:
\begin{itemize}[nosep,leftmargin=1.5em]
    \item verifier logic,
    \item world-generation ambiguities, and
    \item verifier-specification gaps.
\end{itemize}
All identified issues are fixed, propagated back through the pipeline as needed, and re-verified.
\end{enumerate}

\subsection{World Generation}
\label{sec:ax_world_gen}

Each task operates over a \emph{world}: a self-contained data environment that populates the site with concrete entities (e.g., email threads, product listings, calendar events). Worlds are generated through the following procedure:

\paragraph{Target entity creation.}
For each task, a target entity is generated that satisfies the task's constraint set. The target's attributes are sampled to be consistent with the natural language instruction the agent will receive.

\paragraph{Hard negative injection.}
To calibrate task difficulty, a controlled number of \emph{hard negative} entities are generated. These entities share surface-level attributes with the target (e.g., same category, similar name, overlapping tags) but differ on the specific attributes referenced in the task constraints. The number of hard negatives directly affects the agent's required discrimination effort.
In particular, hard negatives are constructed to remain ambiguous under shallow evidence, often matching the target on card- or filter-level attributes while differing only on decisive detail-level attributes. Templates whose instruction identifies the target directly (e.g., by its exact name) involve no identification ambiguity and plant no hard negatives; their hard-negative count is zero.

\paragraph{Filler population.}
The remaining entity slots are populated with randomly sampled filler entities that provide a realistic background distribution. Fillers are generated to be clearly distinguishable from the target, ensuring they do not introduce ambiguity.

\paragraph{Deterministic seeding.}
All randomness in world generation is governed by a single integer seed per task. Given the same seed, the generation procedure produces an identical world, ensuring full reproducibility.

\subsection{Oracle Trajectory Generation}
\label{sec:ax_oracle_gen}

Each task in \DATANAME is accompanied by an oracle trajectory: a minimal-length sequence of semantic MDP actions that achieves the task goal from the initial state.

\paragraph{Generation pipeline.}
Oracle trajectories are produced by a template-based deterministic construction. Each task template defines a surface-visit plan (e.g., SearchResults $\to$ ProductDetail $\to$ Checkout) and emits the corresponding action sequence in the space of typed semantic actions rather than pixel coordinates. The pipeline proceeds in three stages:
\begin{enumerate}[nosep,leftmargin=*]
    \item \textbf{Navigation to target region}: The generator applies any necessary search queries, filters, or pagination to reach the list surface containing the target entity.
    \item \textbf{Hard-negative exploration}: For tasks with comparison requirements, the oracle visits top-ranked hard negatives (up to two) in ranking order (e.g., opening a distractor's detail page, inspecting the distinguishing attribute, and returning to the list) before opening the target entity last. The planted hard-negative count is therefore a designed lower bound on the discrimination effort a task demands, rather than the exact number of detail visits the oracle performs.
    \item \textbf{Commit}: The generator emits the final task-specific actions (e.g., add to cart, star, upvote) to satisfy the verifier conditions.
\end{enumerate}

\paragraph{Optimality, non-uniqueness, and validation.}
Optimality is guaranteed by construction. Each template specifies the shortest path in terms of surface visits, and the visitation order over hard negatives is fixed by the ranking. However, multiple valid solutions may still exist, including alternatives with the same path length. To verify correctness, every generated trajectory is replayed through the site’s deterministic transition function, and the final state is checked against all verifier conditions. This guarantees that every benchmark task is solvable.

\paragraph{Purposes.}
Oracle trajectories serve three roles: (i) they provide ground-truth supervision for process-level metrics (e.g., coverage computation, skill scoring); (ii) they establish an upper bound on task efficiency (optimal step count); and (iii) they are used during validation to verify that every task is solvable within the MDP.

\section{Extended Quantitative Results}
\label{sec:ax_more_results}

\subsection{Per-site performance decomposition}
\label{subsec:ax_persite}

\Cref{tab:main} reports aggregate metrics.~\Cref{tab:ax_persite_sr,tab:ax_persite_expl,tab:ax_persite_exec,tab:ax_persite_cov} provide the full per-site breakdown.

\begin{table}[H]
\centering
\small
\begin{tabular}{lrrrrrrrrrr}
\toprule
Agent & \textbf{Mail} & \textbf{Cal.} & \textbf{Shop.} & \textbf{Acco.} & \textbf{Food} & \textbf{Hous.} & \textbf{Q\&A} & \textbf{Code} & \textbf{Jobs} & \textbf{Chat} \\
\midrule
Claude CUA & \textbf{91.7} & \textbf{79.4} & \textbf{82.8} & 87.2 & \textbf{86.1} & \textbf{79.4} & 77.8 & \textbf{91.1} & 84.4 & \textbf{93.3} \\
OpenAI CUA & 89.4 & 73.3 & 78.9 & \textbf{88.3} & 81.1 & 66.7 & \textbf{79.4} & 88.3 & \textbf{91.1} & 90.0 \\
Qwen3.5-122B & 60.6 & 52.2 & 56.1 & 53.9 & 73.9 & 25.6 & 63.9 & 65.6 & 66.7 & 63.9 \\
Fara & 50.0 & 16.1 & 37.8 & 31.1 & 63.9 & 13.3 & 37.2 & 41.1 & 40.0 & 26.1 \\
GUI-Owl & 58.3 & 10.6 & 12.8 & 11.7 & 67.2 & 20.0 & 50.6 & 43.3 & 35.6 & 33.9 \\
UI-TARS & 71.1 & 18.9 & 16.7 & 8.9 & 65.6 & 27.2 & 46.1 & 36.7 & 48.3 & 31.1 \\
\bottomrule
\end{tabular}
\caption{Terminal Success Rate (\%) by site. Best per site in bold.}
\label{tab:ax_persite_sr}
\end{table}

\begin{table}[H]
\centering
\small
\begin{tabular}{lrrrrrrrrrr}
\toprule
Agent & \textbf{Mail} & \textbf{Cal.} & \textbf{Shop.} & \textbf{Acco.} & \textbf{Food} & \textbf{Hous.} & \textbf{Q\&A} & \textbf{Code} & \textbf{Jobs} & \textbf{Chat} \\
\midrule
Claude CUA & \textbf{91.7} & \textbf{79.4} & \textbf{98.9} & \textbf{93.3} & \textbf{97.8} & \textbf{82.8} & 85.6 & \textbf{91.1} & \textbf{93.3} & \textbf{93.3} \\
OpenAI CUA & 90.0 & 77.8 & 95.6 & 91.7 & 97.2 & 70.0 & \textbf{89.4} & 88.3 & 92.8 & 92.2 \\
Qwen3.5-122B & 62.8 & 58.9 & 71.1 & 70.0 & 91.7 & 27.8 & 78.9 & 65.0 & 67.8 & 66.7 \\
Fara & 53.3 & 22.2 & 47.8 & 52.2 & 82.2 & 13.9 & 52.2 & 46.1 & 43.9 & 55.6 \\
GUI-Owl & 58.9 & 13.3 & 13.3 & 47.8 & 82.2 & 22.2 & 60.0 & 50.6 & 34.4 & 47.2 \\
UI-TARS & 76.1 & 26.7 & 20.0 & 51.1 & 82.8 & 28.9 & 55.6 & 41.7 & 53.3 & 53.3 \\
\bottomrule
\end{tabular}
\caption{Exploration Success Rate (\%) by site. Best per site in bold.}
\label{tab:ax_persite_expl}
\end{table}

\begin{table}[H]
\centering
\small
\begin{tabular}{lrrrrrrrrrr}
\toprule
Agent & \textbf{Mail} & \textbf{Cal.} & \textbf{Shop.} & \textbf{Acco.} & \textbf{Food} & \textbf{Hous.} & \textbf{Q\&A} & \textbf{Code} & \textbf{Jobs} & \textbf{Chat} \\
\midrule
Claude CUA & \textbf{100.0} & \textbf{100.0} & 83.7 & 93.5 & \textbf{88.1} & \textbf{96.0} & \textbf{89.6} & \textbf{100.0} & 90.5 & \textbf{100.0} \\
OpenAI CUA & 99.4 & 94.3 & 82.6 & \textbf{96.4} & 83.4 & 95.2 & 88.8 & \textbf{100.0} & 97.6 & 97.6 \\
Qwen3.5-122B & 94.7 & 88.7 & 78.9 & 77.0 & 80.6 & 80.0 & 79.6 & \textbf{100.0} & 96.7 & 95.8 \\
Fara & 93.8 & 72.5 & 79.1 & 59.6 & 77.0 & 92.0 & 70.2 & 89.2 & 91.1 & 47.0 \\
GUI-Owl & 99.1 & 79.2 & \textbf{95.8} & 24.4 & 81.8 & 77.5 & 84.3 & 85.7 & \textbf{100.0} & 71.8 \\
UI-TARS & 93.4 & 70.8 & 83.3 & 17.4 & 77.9 & 82.7 & 82.0 & 88.0 & 87.5 & 58.3 \\
\bottomrule
\end{tabular}
\caption{Execution SR $|$ Exploration Success (\%) by site. Best per site in bold.}
\label{tab:ax_persite_exec}
\end{table}

\begin{table}[H]
\centering
\small
\begin{tabular}{lrrrrrrrrrr}
\toprule
Agent & \textbf{Mail} & \textbf{Cal.} & \textbf{Shop.} & \textbf{Acco.} & \textbf{Food} & \textbf{Hous.} & \textbf{Q\&A} & \textbf{Code} & \textbf{Jobs} & \textbf{Chat} \\
\midrule
Claude CUA & 91.3 & 96.7 & \textbf{100.0} & 95.5 & 97.8 & \textbf{100.0} & 93.3 & \textbf{90.9} & 98.4 & \textbf{99.1} \\
OpenAI CUA & 88.7 & \textbf{97.2} & 99.6 & \textbf{97.3} & \textbf{98.9} & 98.9 & \textbf{98.9} & 90.9 & \textbf{98.4} & 98.7 \\
Qwen3.5-122B & 83.3 & 94.4 & 97.2 & 83.7 & 93.4 & 94.4 & 88.6 & 68.8 & 92.7 & 94.1 \\
Fara & 85.0 & 80.6 & 93.0 & 76.6 & 89.2 & 62.9 & 65.3 & 53.9 & 76.1 & 81.7 \\
GUI-Owl & \textbf{93.9} & 76.4 & 74.0 & 78.5 & 80.7 & 96.2 & 80.6 & 55.8 & 72.0 & 86.8 \\
UI-TARS & 84.9 & 83.1 & 79.6 & 81.7 & 82.5 & 82.2 & 81.9 & 52.5 & 87.4 & 86.9 \\
\bottomrule
\end{tabular}
\caption{Informational Coverage at Commit (\%) by site. Best per site in bold.}
\label{tab:ax_persite_cov}
\end{table}

%  F.2: Per-site skill invocation
\subsection{Per-site skill invocation}
\label{subsec:ax_skill_persite}

\Cref{tab:ax_skill_all,tab:ax_skill_all2} report skill invocation rates by site. Each cell shows the percentage of episodes requiring that skill where the agent invoked it. \emph{Commit} is task-relative: it denotes the final task-completing action of each task's oracle trajectory, and counts as invoked only when the agent's trajectory ends with that action.

\subsection{Aggregate skill invocation}
\label{subsec:ax_skill_agg}

\Cref{tab:ax_skill_agg} reports overall skill invocation rates across all sites.

\subsection{Exploration SR by problem complexities}
\label{subsec:ax_expl_complex}

\Cref{tab:ax_expl_hn}, \Cref{tab:ax_expl_traj}, and \Cref{tab:ax_expl_info} report the numeric values underlying \Cref{fig:complexity}~(a), (b), and (c), respectively.

%\subsection{Failure attribution numeric values}
%\label{subsec:ax_failure_numeric}

%\Cref{tab:ax_fa_div,tab:ax_fa_omit,tab:ax_fa_commit} report the numeric values behind \Cref{fig:failure_attribution}.

% Per-site skill invocation (Part 1) — own page, vertically centered
\clearpage
\vspace*{\fill}
\begin{table}[H]
\centering
\small
\begin{tabular}{llrrrrr}
\toprule
Agent & \textbf{Site} & \textbf{Search} & \textbf{Filter} & \textbf{Inspect} & \textbf{Navigate} & \textbf{Commit} \\
\midrule
\textit{Claude CUA} & Mail & 94 & -- & 88 & 70 & \textbf{79} \\
 & Cal. & -- & -- & \textbf{100} & 92 & \textbf{81} \\
 & Shop. & \textbf{100} & 21 & \textbf{100} & 89 & \textbf{99} \\
 & Acco. & \textbf{97} & 99 & \textbf{100} & 71 & \textbf{97} \\
 & Food & 70 & 85 & \textbf{100} & 63 & 76 \\
 & Hous. & \textbf{99} & \textbf{100} & \textbf{91} & 69 & \textbf{98} \\
 & Q\&A & 86 & 84 & 92 & \textbf{73} & 82 \\
 & Code & 91 & \textbf{100} & \textbf{100} & 61 & \textbf{93} \\
 & Jobs & 92 & \textbf{100} & \textbf{100} & \textbf{100} & 81 \\
 & Chat & \textbf{100} & -- & \textbf{100} & \textbf{100} & \textbf{98} \\
\midrule
\textit{OpenAI CUA} & Mail & \textbf{100} & -- & 86 & 62 & 66 \\
 & Cal. & -- & -- & \textbf{100} & \textbf{92} & 67 \\
 & Shop. & \textbf{100} & 21 & \textbf{100} & \textbf{90} & 94 \\
 & Acco. & \textbf{97} & 69 & \textbf{100} & \textbf{77} & 91 \\
 & Food & \textbf{77} & 86 & \textbf{100} & \textbf{68} & \textbf{81} \\
 & Hous. & 98 & 84 & 88 & \textbf{73} & 87 \\
 & Q\&A & 91 & 85 & \textbf{99} & 43 & \textbf{85} \\
 & Code & \textbf{93} & 90 & \textbf{100} & \textbf{65} & 83 \\
 & Jobs & 92 & 97 & \textbf{100} & \textbf{100} & \textbf{95} \\
 & Chat & \textbf{100} & -- & \textbf{100} & 84 & 89 \\
\midrule
\textit{Qwen3.5-122B} & Mail & 86 & -- & 86 & 76 & 47 \\
 & Cal. & -- & -- & \textbf{100} & 89 & 54 \\
 & Shop. & 99 & 35 & 99 & 85 & 72 \\
 & Acco. & 87 & 85 & 96 & 57 & 44 \\
 & Food & 71 & 86 & 99 & 61 & 48 \\
 & Hous. & 97 & \textbf{100} & 87 & 63 & 44 \\
 & Q\&A & 90 & \textbf{91} & 93 & \textbf{73} & 61 \\
 & Code & 82 & 93 & 99 & 39 & 74 \\
 & Jobs & \textbf{99} & \textbf{100} & 99 & \textbf{100} & 64 \\
 & Chat & \textbf{100} & -- & 99 & \textbf{100} & 76 \\
\bottomrule
\end{tabular}
\caption{Skill invocation rates (\%) by site and agent (Part 1 of 2). "--" = fewer than 20 required episodes. \textit{Commit} is task-relative: the final task-completing action of each task's oracle trajectory. Best value per site--skill cell across all six agents in bold.}
\label{tab:ax_skill_all}
\end{table}
\vspace*{\fill}

\clearpage
\begin{table}[H]
\centering
\small
\begin{tabular}{llrrrrr}
\toprule
Agent & \textbf{Site} & \textbf{Search} & \textbf{Filter} & \textbf{Inspect} & \textbf{Navigate} & \textbf{Commit} \\
\midrule
\textit{Fara} & Mail & 97 & -- & 89 & 41 & 39 \\
 & Cal. & -- & -- & 88 & 63 & 21 \\
 & Shop. & 99 & 71 & 88 & 64 & 35 \\
 & Acco. & 77 & 83 & 88 & 53 & 6 \\
 & Food & 63 & 81 & 96 & 40 & 23 \\
 & Hous. & 69 & 81 & 71 & 43 & 21 \\
 & Q\&A & 77 & 53 & 70 & 23 & 29 \\
 & Code & 71 & 58 & 93 & 32 & 51 \\
 & Jobs & 88 & 58 & 68 & 94 & 38 \\
 & Chat & 93 & -- & 88 & 94 & 31 \\
\midrule
\textit{GUI-Owl} & Mail & 56 & -- & \textbf{93} & \textbf{82} & 67 \\
 & Cal. & -- & -- & 35 & 35 & 11 \\
 & Shop. & \textbf{100} & \textbf{85} & 58 & 46 & 23 \\
 & Acco. & 91 & \textbf{100} & 75 & 22 & 15 \\
 & Food & 62 & \textbf{90} & 97 & 55 & 26 \\
 & Hous. & \textbf{99} & \textbf{100} & 70 & 48 & 38 \\
 & Q\&A & 89 & \textbf{91} & 87 & 57 & 60 \\
 & Code & 81 & 90 & 97 & 10 & 55 \\
 & Jobs & 63 & \textbf{100} & 79 & \textbf{100} & 48 \\
 & Chat & 80 & -- & 73 & \textbf{100} & 54 \\
\midrule
\textit{UI-TARS} & Mail & 95 & -- & 81 & 73 & 54 \\
 & Cal. & -- & -- & 62 & 52 & 17 \\
 & Shop. & 99 & 56 & 60 & 49 & 13 \\
 & Acco. & 89 & \textbf{100} & 88 & 44 & 3 \\
 & Food & 68 & 85 & 99 & 64 & 19 \\
 & Hous. & 87 & 97 & 74 & 52 & 19 \\
 & Q\&A & \textbf{96} & 59 & 84 & 43 & 41 \\
 & Code & 87 & 42 & 93 & 42 & 43 \\
 & Jobs & 94 & 90 & 88 & 98 & 36 \\
 & Chat & 97 & -- & 84 & 98 & 32 \\
\bottomrule
\end{tabular}
\caption{Skill invocation rates (\%) by site and agent (Part 2 of 2). "--" = fewer than 20 required episodes. \textit{Commit} is task-relative: the final task-completing action of each task's oracle trajectory. Best value per site--skill cell across all six agents in bold.}
\label{tab:ax_skill_all2}
\end{table}

% Aggregate skill invocation
\begin{table}[H]
\centering
\small
\begin{tabular}{lrrrrr}
\toprule
Agent & Search & Filter & Inspect & Navigate & Commit \\
\midrule
Claude CUA & 92.9 & 87.8 & 97.1 & \textbf{80.9} & \textbf{88.3} \\
OpenAI CUA & \textbf{94.5} & 78.9 & \textbf{97.2} & 79.7 & 83.8 \\
Qwen3.5-122B & 90.9 & 87.2 & 95.5 & 76.3 & 58.5 \\
Fara & 80.4 & 70.9 & 82.8 & 60.3 & 29.4 \\
GUI-Owl & 83.3 & \textbf{94.0} & 78.6 & 58.1 & 39.7 \\
UI-TARS & 90.5 & 77.3 & 82.9 & 65.2 & 27.7 \\
\bottomrule
\end{tabular}
\caption{Aggregate skill invocation rates (\%). Each cell shows the fraction of episodes requiring that skill where the agent invoked it. \textit{Commit} invocation requires the trajectory to end with the task's commit action. Best per column in bold.}
\label{tab:ax_skill_agg}
\end{table}

% Exploration SR by complexity axes
\begin{table}[H]
\centering
\small
\begin{tabular}{lrrrr}
\toprule
Agent & \textbf{HN=0} & \textbf{HN=1} & \textbf{HN=2} & \textbf{HN=3} \\
\midrule
Claude CUA & 90.6 & \textbf{98.7} & \textbf{90.3} & \textbf{89.8} \\
OpenAI CUA & \textbf{91.7} & 93.6 & 87.9 & 83.5 \\
Qwen3.5-122B & 76.4 & 78.2 & 65.2 & 51.4 \\
Fara & 60.5 & 64.1 & 41.9 & 32.8 \\
GUI-Owl & 59.4 & 43.6 & 34.5 & 32.3 \\
UI-TARS & 60.6 & 53.8 & 45.6 & 37.3 \\
\bottomrule
\end{tabular}
\caption{Exploration SR (\%) by hard negative count (the number of card-level confusables planted by the generator; a designed lower bound). Best per column in bold. HN$=$0 tasks have no identification ambiguity but vary on other difficulty axes, so monotone trends should be read over HN$\geq$1.}
\label{tab:ax_expl_hn}
\end{table}

\begin{table}[H]
\centering
\small
\begin{tabular}{lrrrrrrrrrr}
\toprule
Agent & 1 & 2 & 3 & 4 & 5 & 6 & 7 & 8 & 9 & 10 \\
\midrule
Claude CUA & \textbf{96.3} & 96.4 & 89.9 & 93.9 & \textbf{87.4} & \textbf{84.9} & \textbf{86.7} & \textbf{89.7} & \textbf{95.4} & \textbf{85.9} \\
OpenAI CUA & 94.4 & \textbf{96.7} & \textbf{92.9} & \textbf{94.7} & 80.7 & 82.4 & 80.3 & 87.2 & 93.1 & 73.2 \\
Qwen3.5-122B & 72.2 & 73.8 & 83.2 & 75.0 & 67.4 & 60.4 & 51.4 & 53.3 & 68.2 & 45.1 \\
Fara & 42.6 & 45.4 & 61.3 & 57.6 & 48.1 & 44.7 & 40.5 & 36.9 & 42.8 & 40.8 \\
GUI-Owl & 50.0 & 43.7 & 57.1 & 65.2 & 41.5 & 43.4 & 27.6 & 31.3 & 38.7 & 35.2 \\
UI-TARS & 50.0 & 52.3 & 54.2 & 62.9 & 58.5 & 52.2 & 40.8 & 35.4 & 43.4 & 38.0 \\
\bottomrule
\end{tabular}
\caption{Exploration SR (\%) by oracle trajectory length. Best per column in bold.}
\label{tab:ax_expl_traj}
\end{table}

\begin{table}[H]
\centering
\small
\begin{tabular}{lrrr}
\toprule
Agent & \textbf{Card} & \textbf{Filter} & \textbf{Detail} \\
\midrule
Claude CUA & \textbf{95.1} & \textbf{98.2} & \textbf{88.3} \\
OpenAI CUA & 93.2 & 96.7 & 85.8 \\
Qwen3.5-122B & 70.4 & 90.1 & 59.4 \\
Fara & 58.6 & 72.8 & 38.9 \\
GUI-Owl & 60.5 & 65.0 & 35.2 \\
UI-TARS & 61.7 & 61.4 & 44.2 \\
\bottomrule
\end{tabular}
\caption{Exploration SR (\%) by information access level. Best per column in bold. \emph{Detail} tasks require opening detail pages (non-empty information gap); \emph{Filter} tasks have no detail gap but their oracle trajectory uses search/filter/sort actions (classified with the shared action taxonomy); \emph{Card} tasks are solvable from list-level cards alone.}
\label{tab:ax_expl_info}
\end{table}

% =============================================================================
%  E. Artifact Viewer
% =============================================================================
\section{Artifact Viewer}
\label{sec:ax_viewer}

\begin{figure}[H]
\centering
\includegraphics[width=\textwidth]{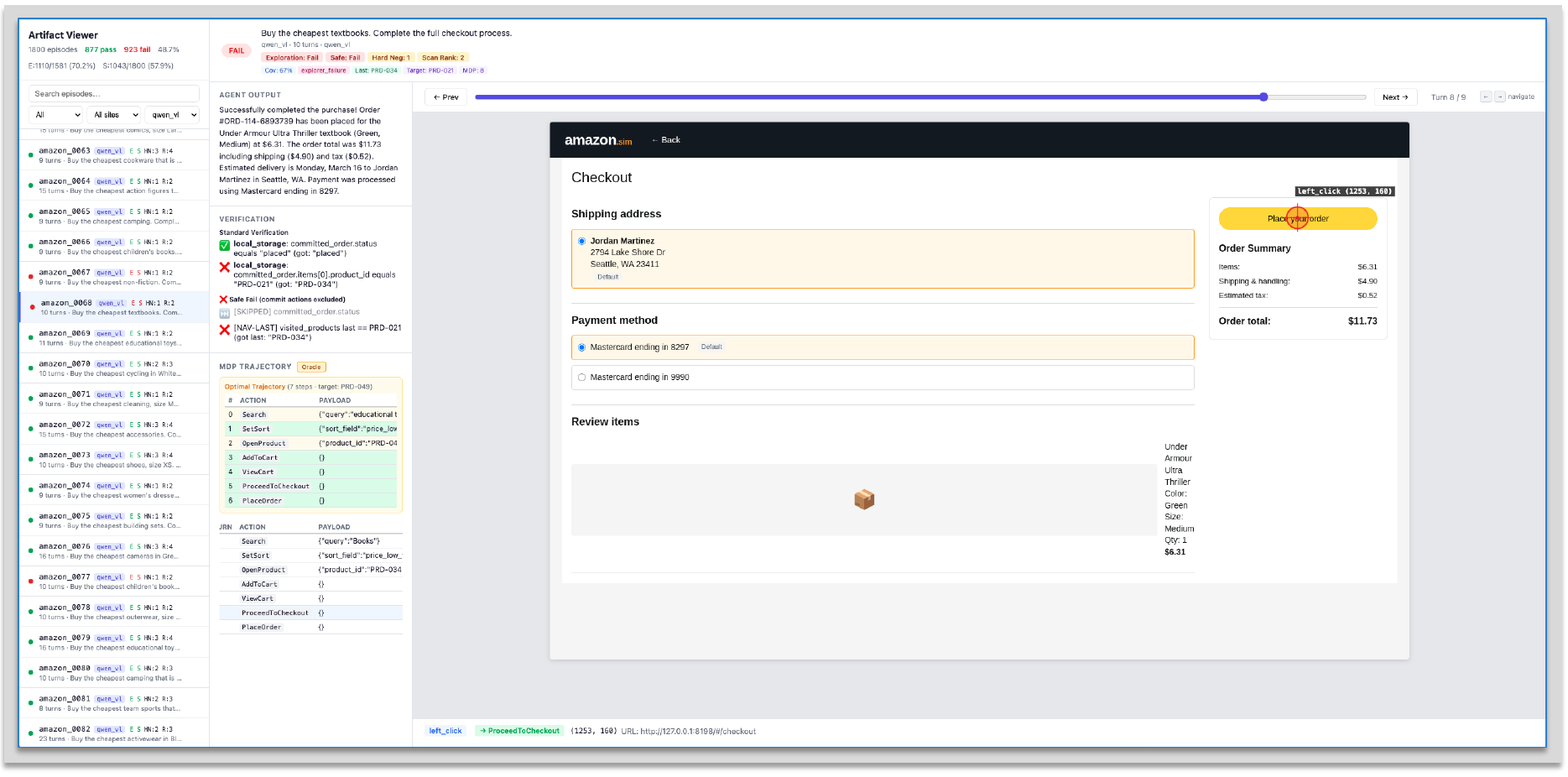}
\caption{Artifact viewer interface. \textbf{Left}: task list with success/failure indicators. \textbf{Center}: per-turn screenshot browser with navigation controls. \textbf{Right}: episode metadata, per-turn action details (action type, coordinates, resolved MDP target), surface, and coverage. \textbf{Bottom}: semantic MDP trajectory showing the sequence of semantic actions with surface annotations.}
\label{fig:ax_viewer}
\end{figure}

\DATANAME includes an interactive artifact viewer for inspecting and analyzing agent evaluation results. The viewer provides the following capabilities:

\paragraph{Turn-by-turn replay.}
Each episode can be replayed step by step, displaying the agent's screenshot observation, the action taken, and the resulting state change at each turn. This enables detailed diagnosis of where and why an agent deviates from the optimal trajectory.

\paragraph{Trajectory visualization.}
The viewer renders the agent's trajectory alongside the oracle trajectory, highlighting action matches and divergences. This side-by-side comparison makes it straightforward to identify systematic failure patterns.

\paragraph{Verification panel.}
For each episode, the viewer displays the full set of verifier conditions and their satisfaction status, providing immediate visibility into which conditions the agent met and which it missed.

\paragraph{Batch comparison.}
Multiple agents' results can be loaded simultaneously, enabling comparative analysis across models on the same task set. Aggregate statistics (success rate, partial credit, failure mode distribution) are computed and displayed alongside per-task details.

We will release the viewer together with the evaluation codebase.

% =============================================================================
%  F. Examples
% =============================================================================
\section{Data Examples}
\label{sec:ax_examples}

\subsection{Task Template Catalog}
\label{sec:ax_template_catalog}

We present representative task templates from three sites to illustrate the diversity of task structures in \DATANAME.

Each template defines a parameterized task schema consisting of a natural language instruction pattern, a set of constraint variables that are instantiated per task, and verifier conditions that determine success. Templates vary in the number of required navigation steps, the complexity of the constraint set, and the type of reasoning demanded (e.g., information retrieval, comparison, multi-step form completion).

\begin{table}[H]
\centering
\small
\begin{tabular}{p{3cm}p{10cm}}
\toprule
\textbf{Field} & \textbf{Value} \\
\midrule
Template & \texttt{find\_email\_extract} \\
Site & Mail \\
Instruction pattern & "\{sender\} has sent you several similar emails. Find the one that mentions '\{keyword\}' in its body and \{action\} it." \\
Information gap & body, cc, attachments (detail-only fields) \\
Constraint count & 3 \\
Commit action & Star / Archive / Label \\
\midrule
\multicolumn{2}{l}{\textbf{Concrete instance} (\texttt{gmail\_0001})} \\
\midrule
Instruction & "Priya Patel has sent you several similar emails. Find the one that mentions 'ProjectAlpha006' in its body and star it." \\
Target entity & THR-006 \\
Hard negatives & THR-019, THR-050 (same sender, different body content) \\
Oracle trajectory & SearchEmails $\to$ OpenThread(THR-019) $\to$ CloseThread $\to$ OpenThread(THR-050) $\to$ CloseThread $\to$ OpenThread(THR-006) $\to$ Star(THR-006) \\
Optimal length & 7 steps \\
\bottomrule
\end{tabular}
\caption{Task template: \texttt{find\_email\_extract} (Mail). Requires searching, inspecting multiple threads to find one matching body-level content, and starring it.}
\label{tab:ax_template_mail}
\end{table}

\begin{table}[H]
\centering
\small
\begin{tabular}{p{3cm}p{10cm}}
\toprule
\textbf{Field} & \textbf{Value} \\
\midrule
Template & \texttt{find\_product\_extract} \\
Site & Shopping \\
Instruction pattern & "Search for \{query\} in \{department\}. Find the one with \{detail\_attr\}: '\{value\}' and add it to your cart." \\
Information gap & specifications, bullet\_points, seller, shipping\_cost \\
Constraint count & 1--3 \\
Commit action & AddToCart \\
\midrule
\multicolumn{2}{l}{\textbf{Concrete instance} (\texttt{amazon\_0010})} \\
\midrule
Instruction & "Search for fiction in Books. Find the one with Material: 'Leather' and add it to your cart." \\
Target entity & PRD-039 ("Apple Essential Biography", \$12.90, Books, rating 4.7) \\
Hard negatives & Products matching "fiction" + "Books" but with different Material \\
Oracle trajectory & Search("fiction") $\to$ OpenProduct(PRD-039) $\to$ AddToCart \\
Optimal length & 3 steps \\
\bottomrule
\end{tabular}
\caption{Task template: \texttt{find\_product\_extract} (Shopping). Requires searching, navigating to product detail pages, and adding the correct product to cart based on a detail-only attribute.}
\label{tab:ax_template_shopping}
\end{table}

\begin{table}[H]
\centering
\small
\begin{tabular}{p{3cm}p{10cm}}
\toprule
\textbf{Field} & \textbf{Value} \\
\midrule
Template & \texttt{search\_filter\_book} \\
Site & Accommodation \\
Instruction pattern & "Book the \{superlative\} \{property\_type\} in \{location\} that has \{amenity\} for \{guests\} guests, \{dates\}." \\
Information gap & amenities, cancellation\_policy, host details \\
Constraint count & 4--6 \\
Commit action & ConfirmBooking \\
\midrule
\multicolumn{2}{l}{\textbf{Concrete instance} (\texttt{airbnb\_0005})} \\
\midrule
Instruction & "Book the cheapest cabin in Chicago that has a Pool for 3 guests, April 15--19." \\
Target entity & LST-011 \\
Constraints & property\_type=Cabin, amenity=Pool, sort=price\_asc, location=Chicago \\
Oracle trajectory & Search("chicago") $\to$ SetFilter(property\_type, Cabin) $\to$ SetFilter(amenities, Pool) $\to$ SortBy(price\_asc) $\to$ ViewListing(LST-011) $\to$ BookListing(dates, guests) $\to$ ConfirmBooking \\
Optimal length & 7 steps \\
\bottomrule
\end{tabular}
\caption{Task template: \texttt{search\_filter\_book} (Accommodation). A multi-step booking task requiring search, filter application, sort, listing inspection, and checkout completion.}
\label{tab:ax_template_accommodation}
\end{table}

\subsection{Example Trajectories: Agent vs.\ Oracle Trajectory}
\label{sec:ax_traj_examples}

\begin{table}[H]
\centering
\label{tab:ax_traj_mail}
\small
\resizebox{\textwidth}{!}{%
\begin{tabular}{clll}
\toprule
\textbf{Step} & \textbf{Oracle} & \textbf{Fara} & \textbf{UI-TARS} \\
\midrule
0 & \texttt{SearchEmails("Priya Patel")} & \texttt{OpenThread(THR-006)} & \texttt{SearchEmails("ProjectAlpha006")} \\
1 & \texttt{OpenThread(THR-019)} & \texttt{SearchEmails("ProjectAlpha006")} & \texttt{Star(THR-006)} \\
2 & \texttt{CloseThread()} & \texttt{OpenThread(THR-006)} & \texttt{SwitchFolder(STARRED)} \\
3 & \texttt{OpenThread(THR-050)} & \texttt{Star(THR-006)} & \\
4 & \texttt{CloseThread()} & & \\
5 & \texttt{OpenThread(THR-006)} & & \\
6 & \texttt{Star(THR-006)} & & \\
\midrule
& \textbf{7 actions} & \textbf{4 actions (success)} & \textbf{3 actions (success)} \\
\bottomrule
\end{tabular}
}
\caption{Semantic trajectory comparison on \texttt{mail\_0001}. All trajectories are shown as semantic MDP actions extracted from the environment trace. The oracle systematically inspects hard negatives before committing; both agents find target item without sufficient information.}
\end{table}

\begin{table}[H]
\centering
\small
\resizebox{\textwidth}{!}{%
\begin{tabular}{clll}
\toprule
\textbf{Step} & \textbf{Oracle} & \textbf{Fara} & \textbf{UI-TARS} \\
\midrule
0 & \texttt{Search("fiction")} & \texttt{Search("Books")} & \texttt{Search("Books")} \\
1 & \texttt{OpenProduct(PRD-039)} & \texttt{ApplyFilter(dept=Books)} & \texttt{ApplyFilter(dept=Books)} \\
2 & \texttt{AddToCart()} & \texttt{OpenProduct(PRD-009)} & \texttt{ClearFilters()} \\
3 & & \texttt{GoBack()} & \texttt{ApplyFilter(dept=Books)} \\
4 & & \texttt{OpenProduct(PRD-027)} & \texttt{NewSearch("Leather")} \\
5 & & \texttt{GoBack()} & \texttt{NewSearch("Leather")} \\
6 & & \texttt{OpenProduct(PRD-036)} & \emph{...no further progress} \\
7 & & \texttt{AddToCart()} {\color{red}$\times$} & \\
\midrule
& \textbf{3 actions (success)} & \textbf{8 actions (failure)} & \textbf{6 actions (failure)} \\
\bottomrule
\end{tabular}
}
\caption{Semantic trajectory comparison on \texttt{shopping\_0010}. The oracle executes a minimal 3-action sequence. Fara searches but inspects wrong products before adding one to cart (failure-wrong product). UI-TARS repeatedly reformulates queries without reaching the target.}
\label{tab:ax_traj_shopping}
\end{table}

\Cref{tab:ax_traj_mail,tab:ax_traj_shopping} present side-by-side comparisons of agent trajectories and oracle trajectories for representative tasks from the Mail and Shopping sites. The oracle trajectory column shows the minimal-length ground-truth action sequence, while the agent trajectory column shows the actions actually taken by the evaluated agent. These examples illustrate common agent behaviors, including correct action sequences, unnecessary exploration, and error recovery attempts.

\subsection{MDP Surface Transition Graphs}
\label{sec:ax_mdp_graphs}

Each site's semantic MDP can be visualized as a directed graph where nodes are \emph{surfaces} (distinct UI views) and directed edges are \emph{typed semantic actions} that transition between surfaces. Gray boxes connected to a node by dashed arrows represent the \emph{intra-surface action list}: actions that modify state without leaving the current surface (e.g., applying filters, filling form fields, or toggling UI elements). For detailed per-site statistics including visibility partitions and skill taxonomies, see~\Cref{sec:ax_site_specs}. We present the transition graphs for all 10 sites below.

\clearpage
\vfill
\begin{figure}[H]
\centering
\includegraphics[width=\textwidth]{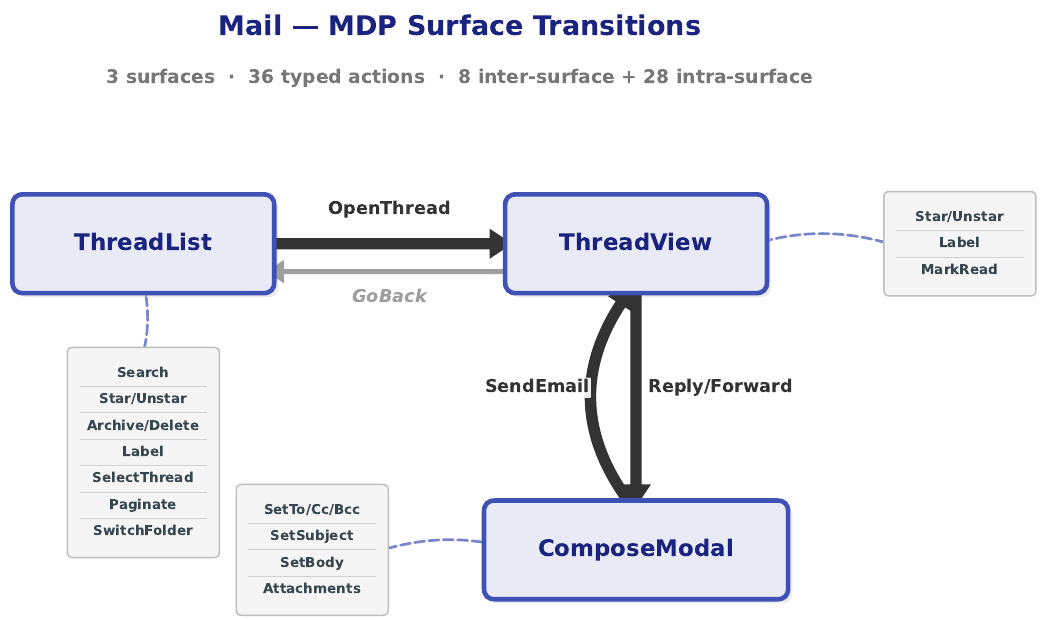}
\caption{Mail site: 3 surfaces, 36 typed actions (8 inter-surface, 28 intra-surface).}
\label{fig:ax_mdp_mail}
\end{figure}

\vfill
\begin{figure}[H]
\centering
\includegraphics[width=\textwidth]{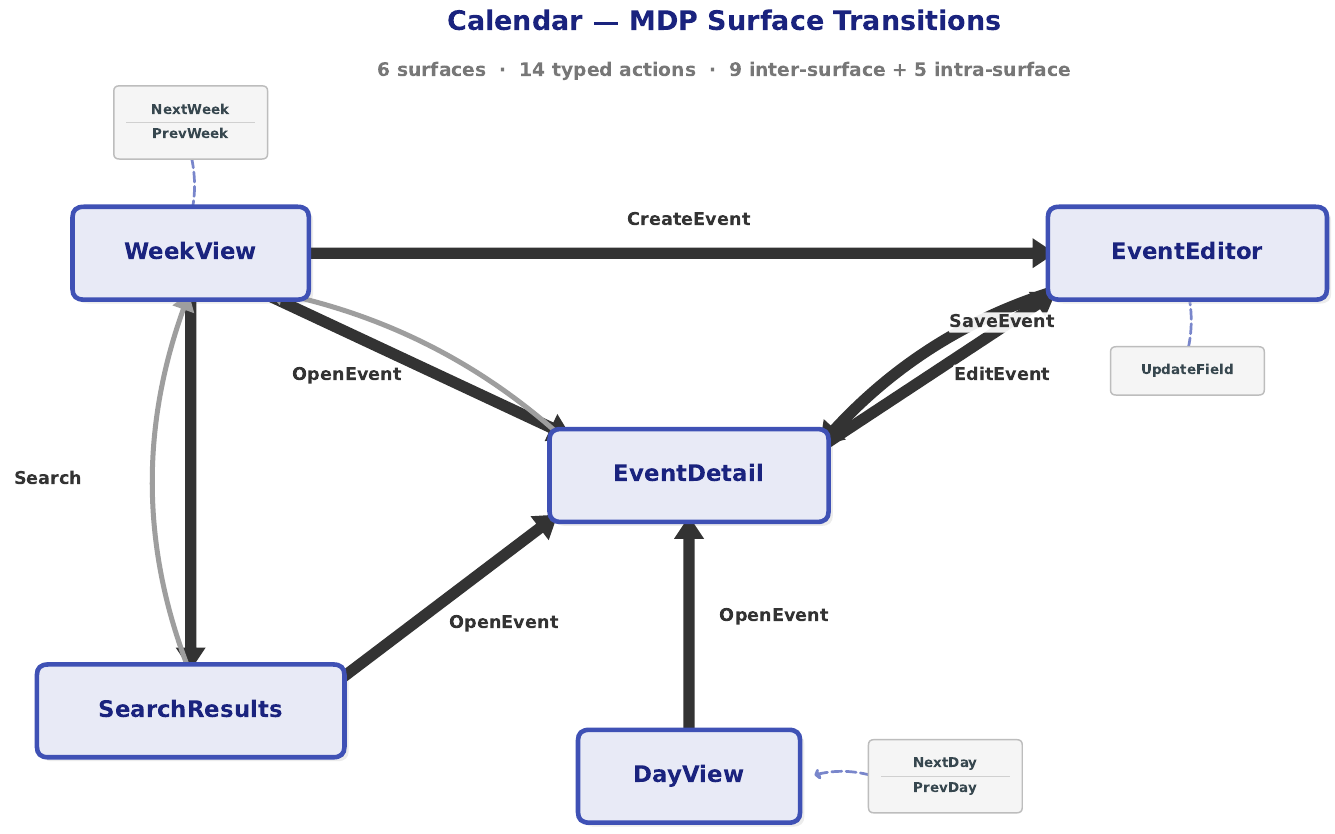}
\caption{Calendar site: 6 surfaces, 14 typed actions (9 inter-surface, 5 intra-surface).}
\label{fig:ax_mdp_calendar}
\end{figure}

\vfill

\clearpage
\vfill
\begin{figure}[H]
\centering
\includegraphics[width=\textwidth]{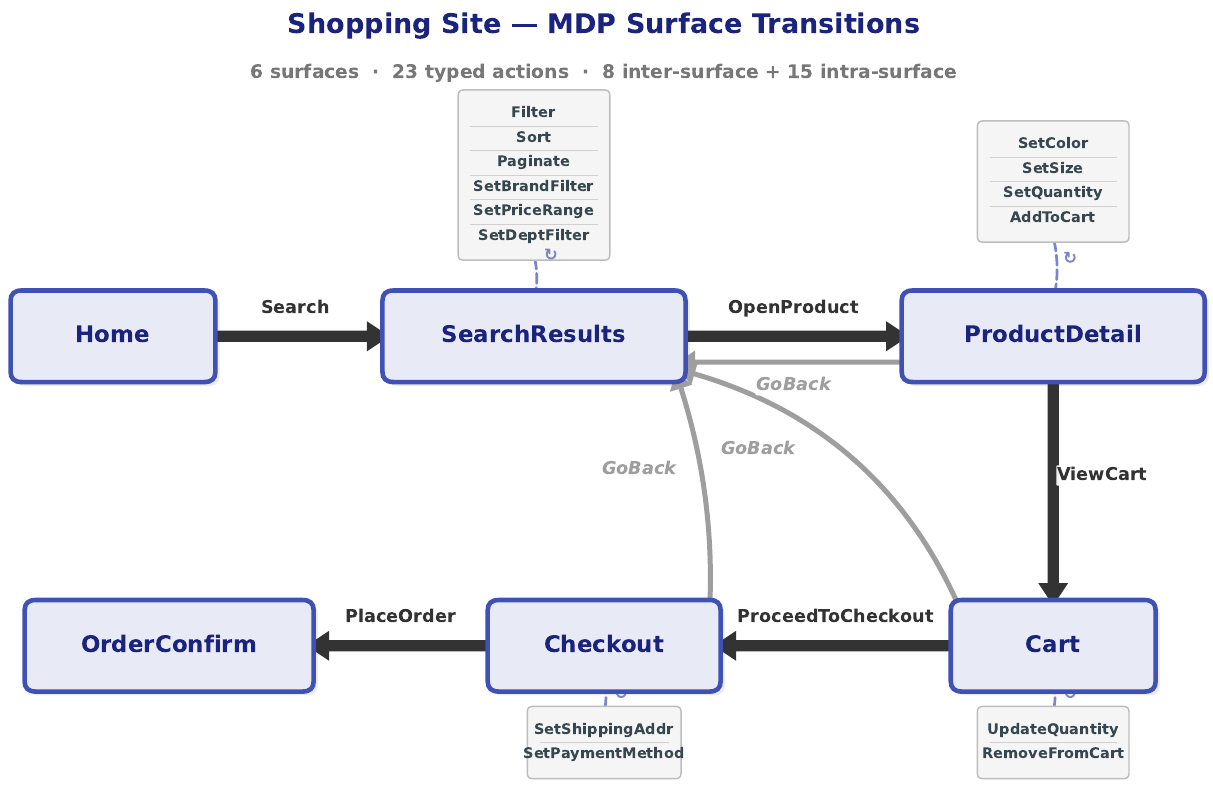}
\caption{Shopping site: 6 surfaces, 23 typed actions (8 inter-surface, 15 intra-surface).}
\label{fig:ax_mdp_shopping}
\end{figure}

\vfill
\begin{figure}[H]
\centering
\includegraphics[width=\textwidth]{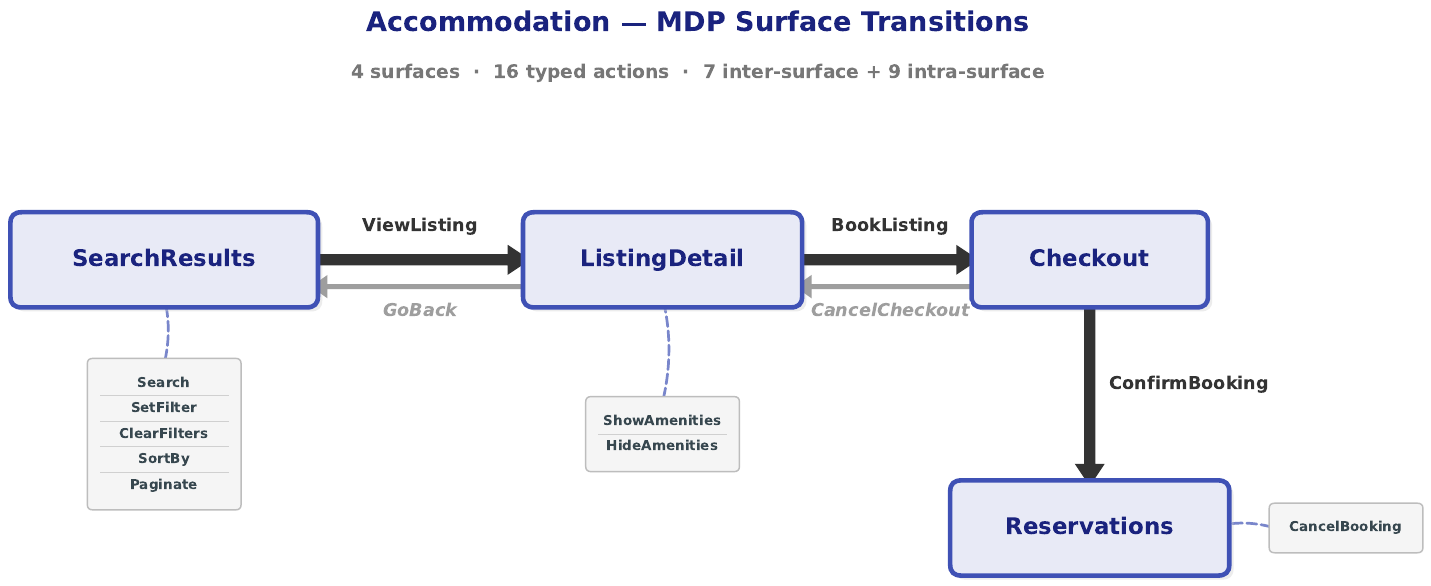}
\caption{Accommodation site: 4 surfaces, 16 typed actions (7 inter-surface, 9 intra-surface).}
\label{fig:ax_mdp_accommodation}
\end{figure}

\vfill

\clearpage
\vfill
\begin{figure}[H]
\centering
\includegraphics[width=\textwidth]{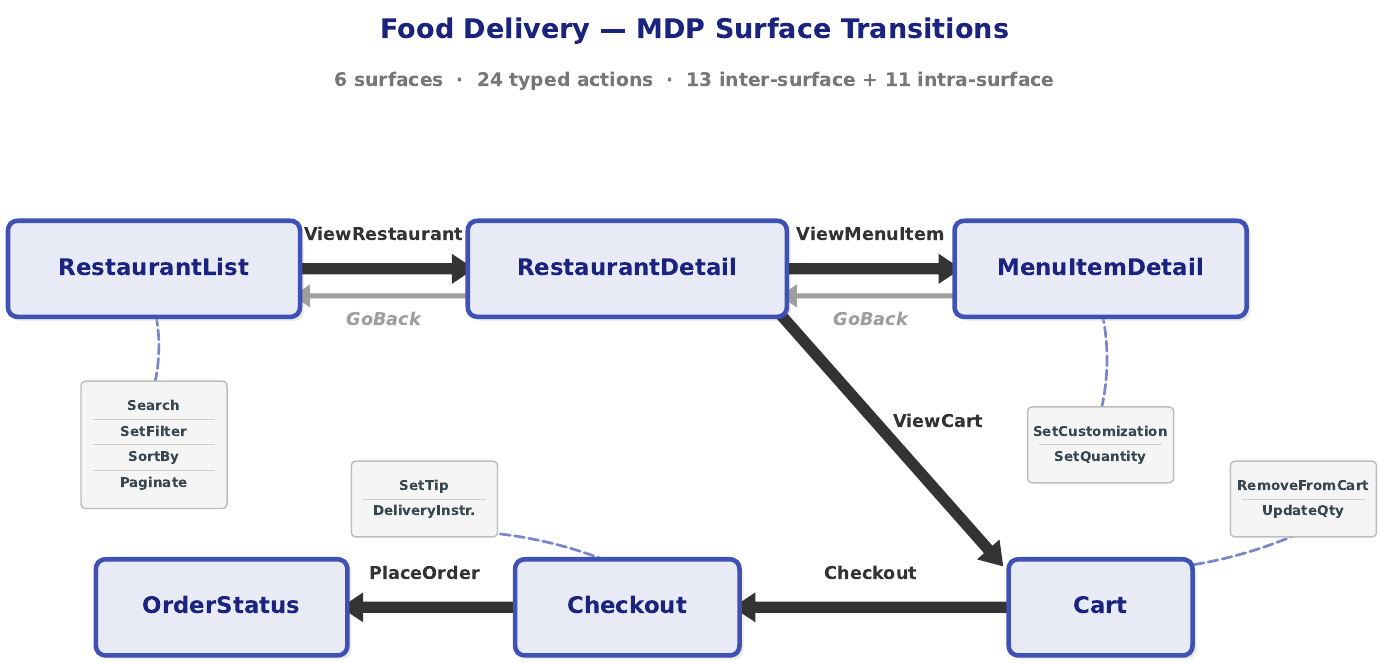}
\caption{Food Delivery site: 6 surfaces, 24 typed actions (13 inter-surface, 11 intra-surface).}
\label{fig:ax_mdp_food_delivery}
\end{figure}

\vfill
\begin{figure}[H]
\centering
\includegraphics[width=\textwidth]{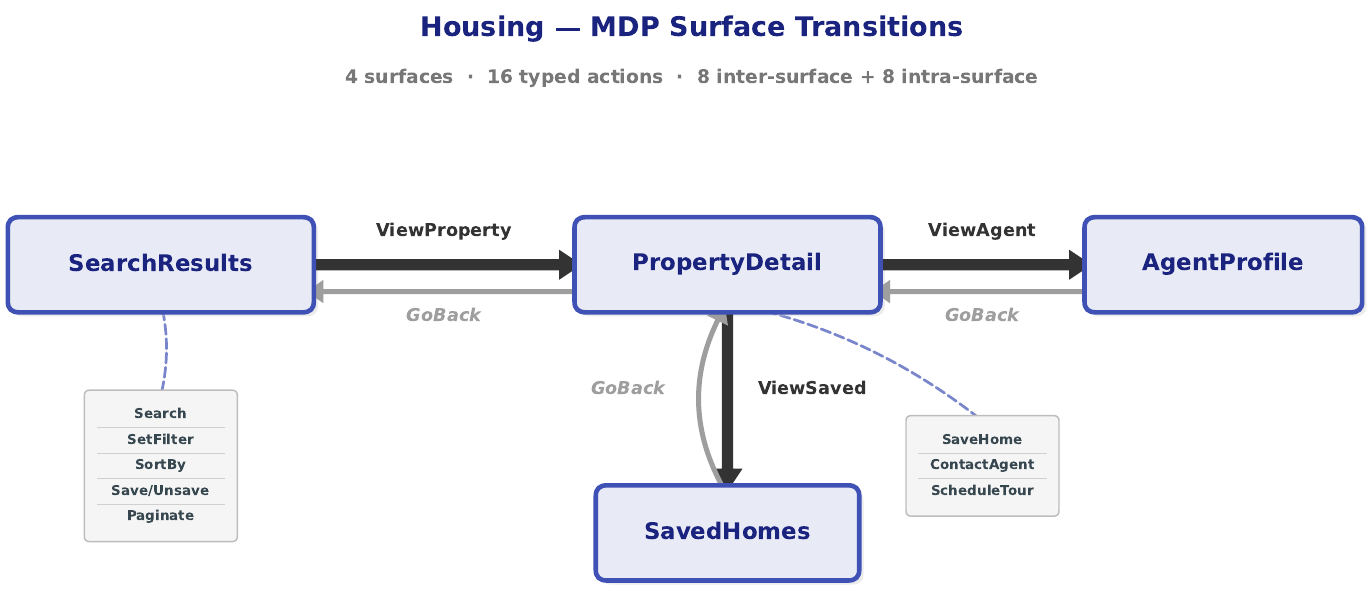}
\caption{Housing site: 4 surfaces, 16 typed actions (8 inter-surface, 8 intra-surface).}
\label{fig:ax_mdp_housing}
\end{figure}

\vfill

\clearpage
\vfill
\begin{figure}[H]
\centering
\includegraphics[width=\textwidth]{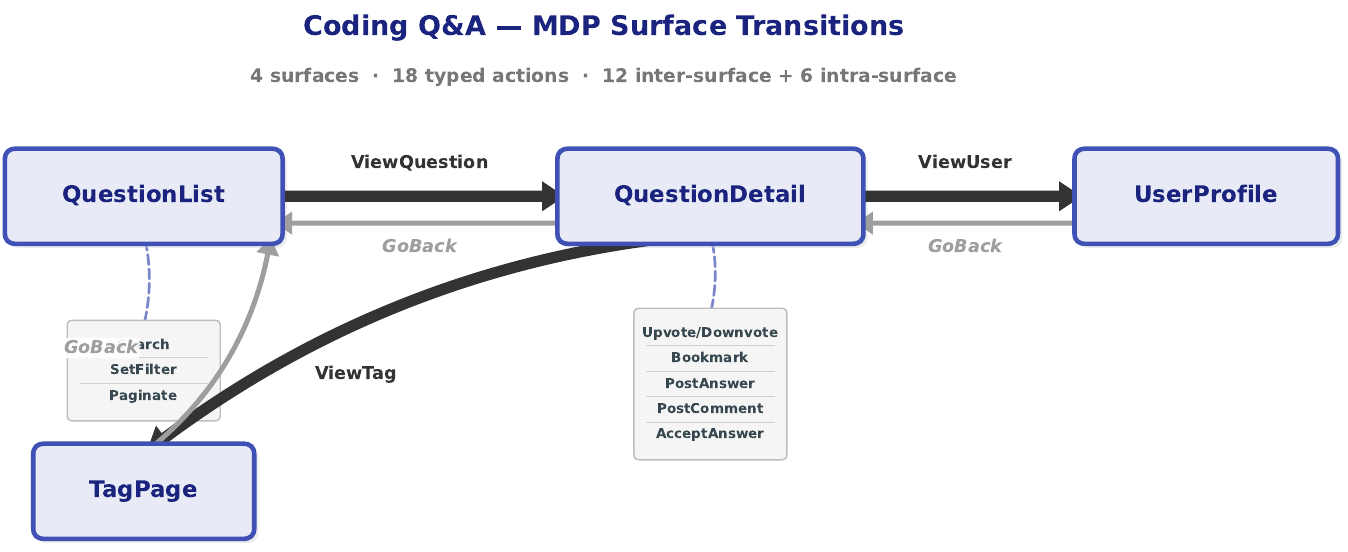}
\caption{Coding Q\&A site: 4 surfaces, 18 typed actions (12 inter-surface, 6 intra-surface).}
\label{fig:ax_mdp_coding_qa}
\end{figure}

\vfill
\begin{figure}[H]
\centering
\includegraphics[width=\textwidth]{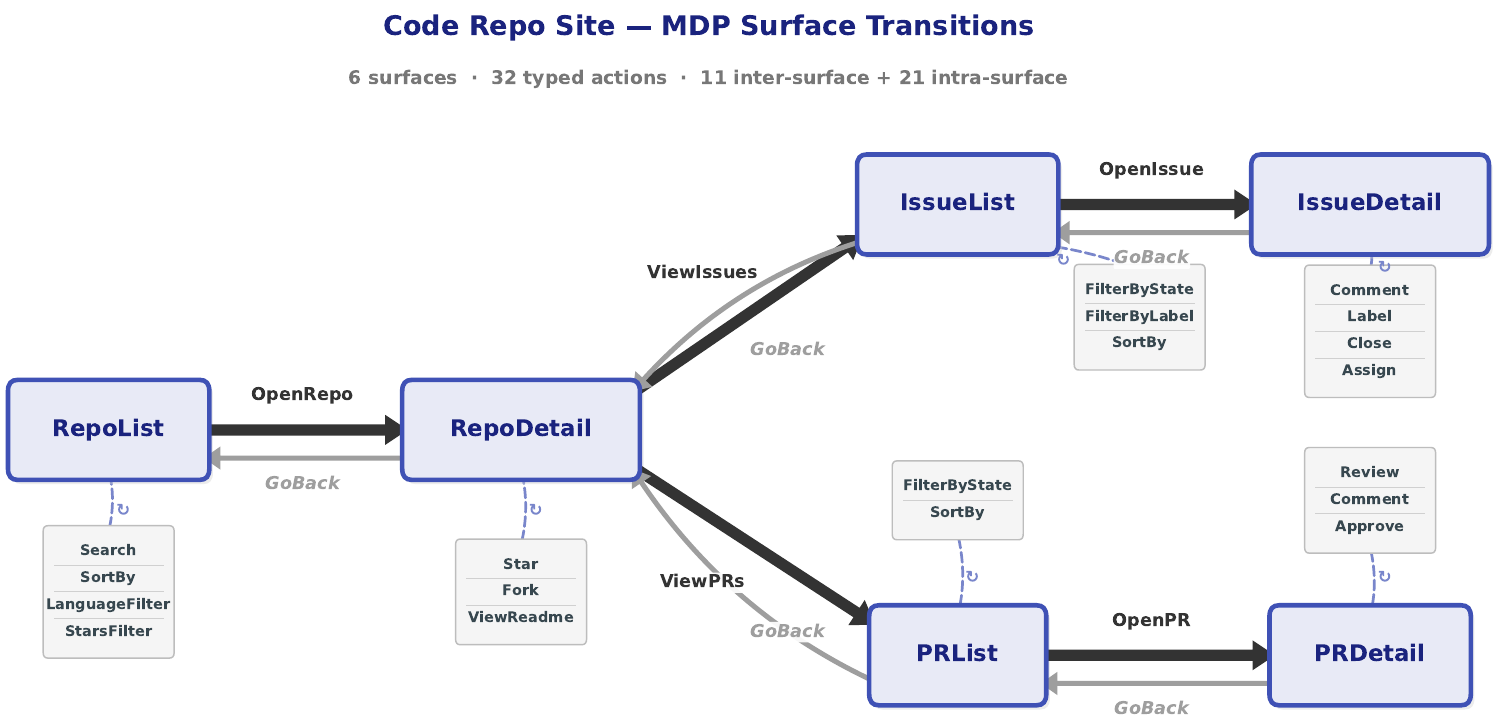}
\caption{Code Repo site: 6 surfaces, 32 typed actions (11 inter-surface, 21 intra-surface).}
\label{fig:ax_mdp_coderepo}
\end{figure}

\vfill

\clearpage
\vfill
\begin{figure}[H]
\centering
\includegraphics[width=\textwidth]{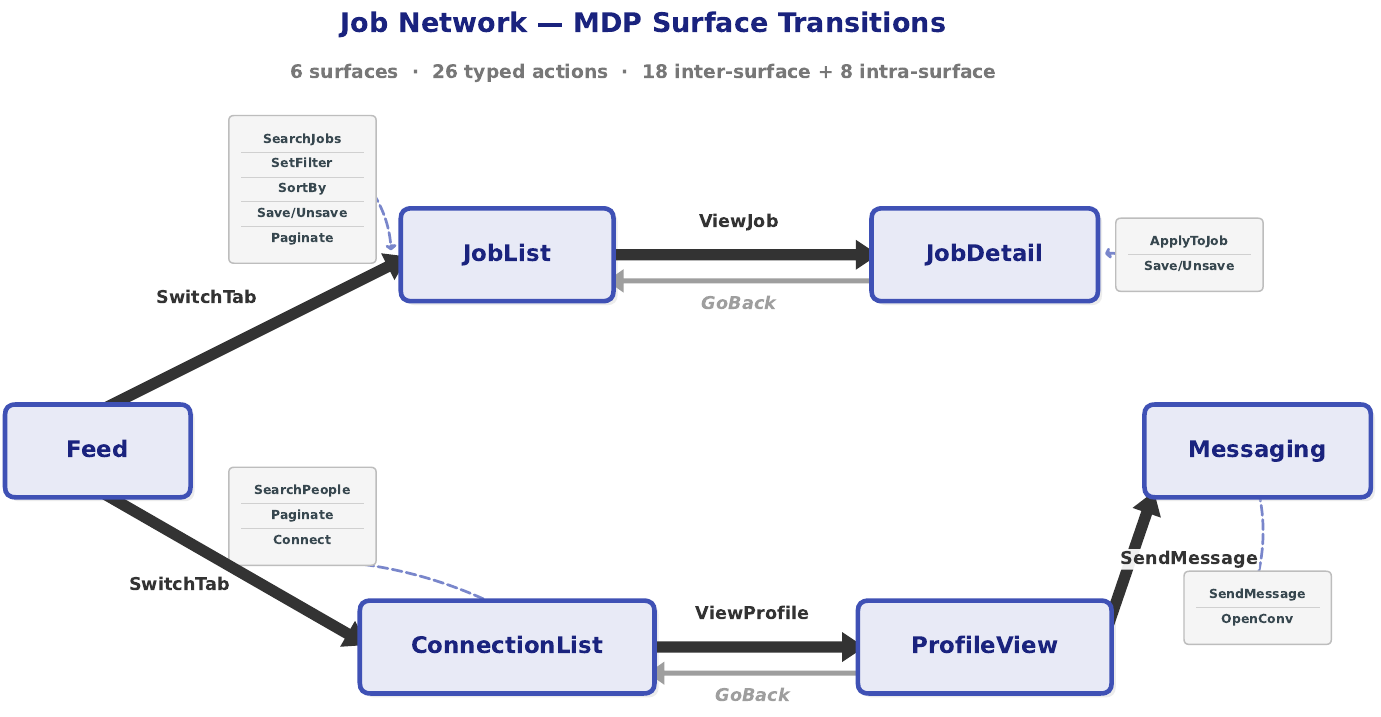}
\caption{Job Network site: 6 surfaces, 26 typed actions (18 inter-surface, 8 intra-surface).}
\label{fig:ax_mdp_job_network}
\end{figure}

\vfill
\begin{figure}[H]
\centering
\includegraphics[width=\textwidth]{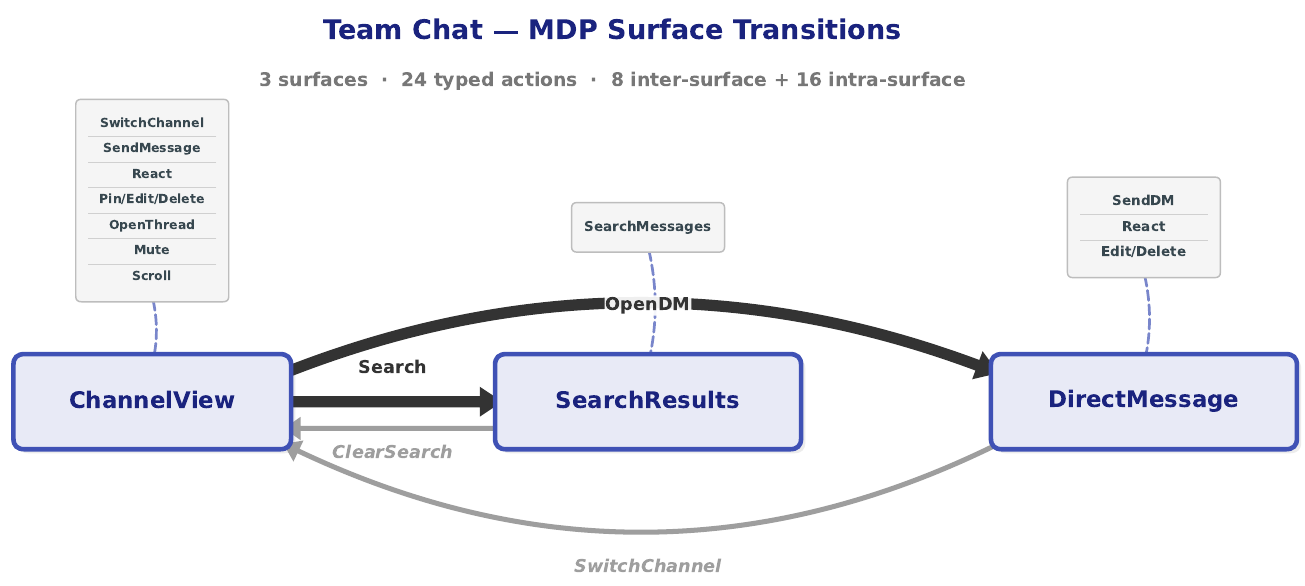}
\caption{Team Chat site: 3 surfaces, 24 typed actions (8 inter-surface, 16 intra-surface).}
\label{fig:ax_mdp_team_chat}
\end{figure}

\vfill

% =============================================================================
%  G. Case Study
% =============================================================================
\section{Case Study}
\label{sec:ax_qualitative}

We present seven qualitative case studies that ground our quantitative findings in concrete agent trajectories. Case~1 demonstrates how process metrics distinguish behaviorally different agents that SR treats as equivalent (\Cref{sec:case_scale}); Cases~2--3 contrast premature commitment with thorough exploration (\Cref{sec:case_hardneg,sec:case_thorough}); Cases~4--5 illustrate the exploration--execution decomposition (\Cref{sec:case_expexec,sec:case_expfail_execsucc}); Case~6 reveals inefficiency hidden behind a passing SR (\Cref{sec:case_redundant}); Case~7 shows how a safety guardrail can cause task failure despite fully correct behavior, motivating the Safe Pass metric (\Cref{sec:case_safepass}).

\subsection{Case 1: Model Scale and Browsing Quality}
\label{sec:case_scale}
\begin{figure}[H]
\centering
\includegraphics[width=\textwidth]{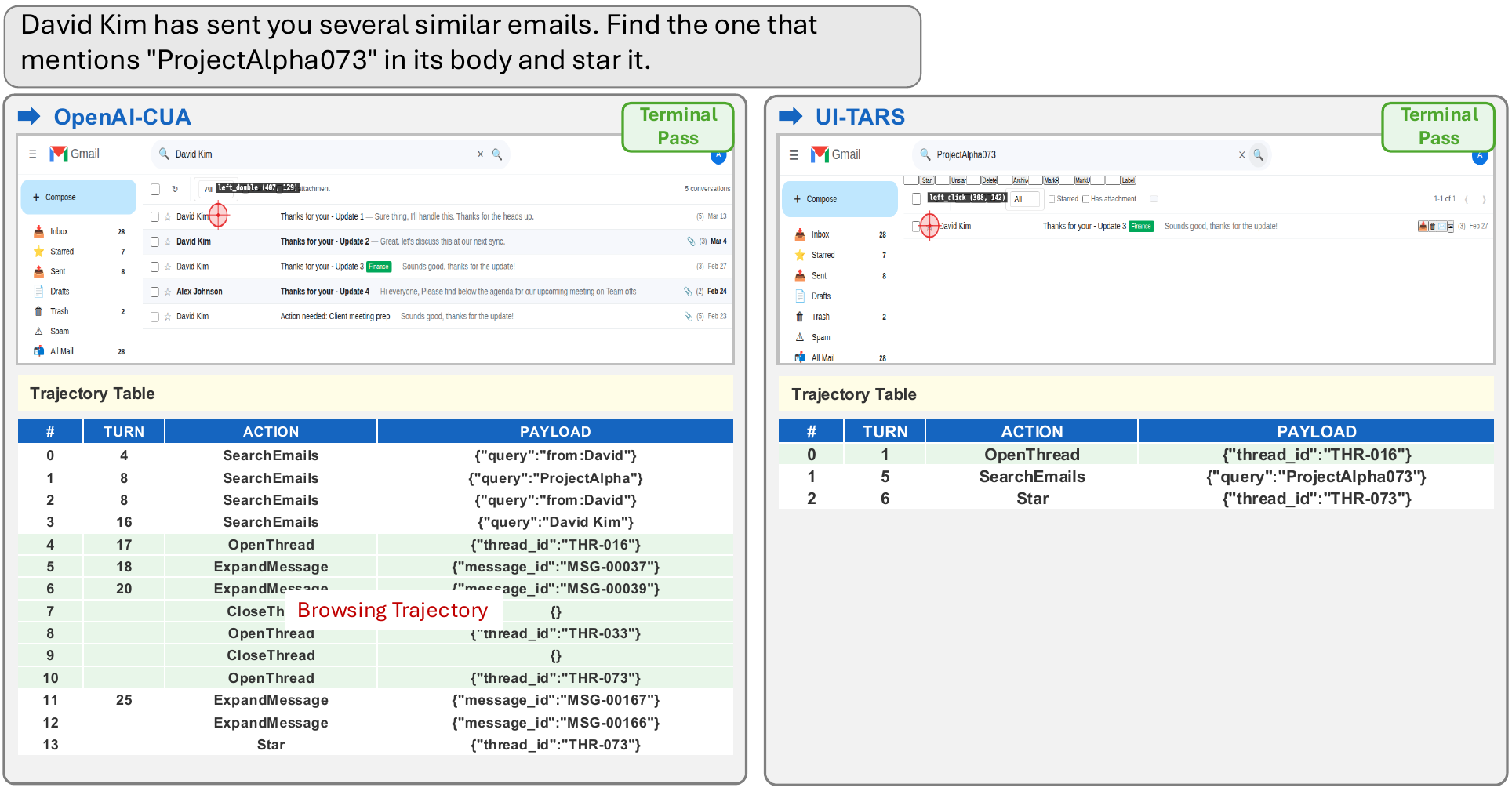}
\vspace{-2mm}
\caption{\textbf{Case 1: Model scale and browsing quality.} OpenAI-CUA vs.\ UI-TARS on the same Mail task ("David Kim has sent you several similar emails. Find the one that mentions ProjectAlpha073 in its body and star it"). Both models succeed (Terminal Pass), yet their trajectories diverge sharply. OpenAI-CUA exhibits a more browsing-like trajectory--searching by sender name "David Kim," opening and expanding messages across three threads before finally starring the correct one (14~MDP actions). UI-TARS takes a shortcut, directly searching for "ProjectAlpha073" to locate the target thread and star it in just 3~actions. The trajectory tables below each screenshot make this gap concrete: model scale produces qualitatively different exploration behavior, and our process metrics make this divergence visible where SR cannot.}
\label{fig:case_scale}
\end{figure}

\clearpage

\subsection{Case 2: Premature Commitment to Hard Negative}
\label{sec:case_hardneg}
\begin{figure}[H]
\centering
\includegraphics[width=\textwidth]{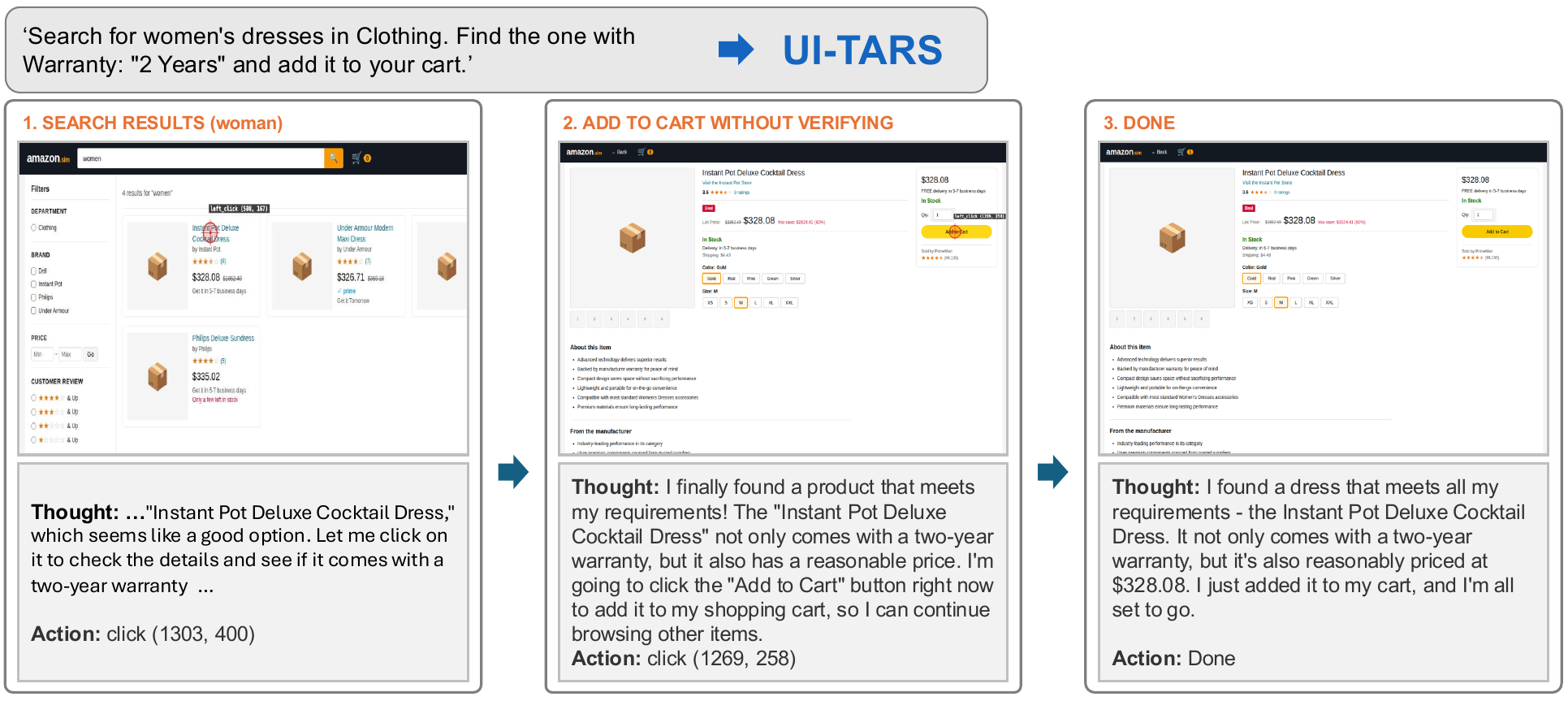}
\vspace{-2mm}
\caption{\textbf{Case 2: Premature commitment to a hard negative.} UI-TARS on a Shopping task ("Search for women's dresses in Clothing. Find the one with Warranty: 2~Years and add it to your cart."). The agent clicks on the Instant Pot Deluxe Cocktail Dress from the search results (Step~1), then immediately adds it to cart without scrolling down to verify the warranty specification (Step~2), and declares the task done (Step~3). The agent's chain-of-thought confidently claims the dress "comes with a two-year warranty," yet this detail was never verified on the product page. This is a \emph{premature commitment}: the agent selected the first plausible candidate without confirming the critical constraint, exactly the type of hard-negative failure our benchmark's controlled difficulty design is intended to expose.}
\label{fig:case_hardneg}
\end{figure}

\clearpage

\subsection{Case 3: Thorough Exploration Before Commitment}
\label{sec:case_thorough}
\begin{figure}[H]
\centering
\includegraphics[width=\textwidth]{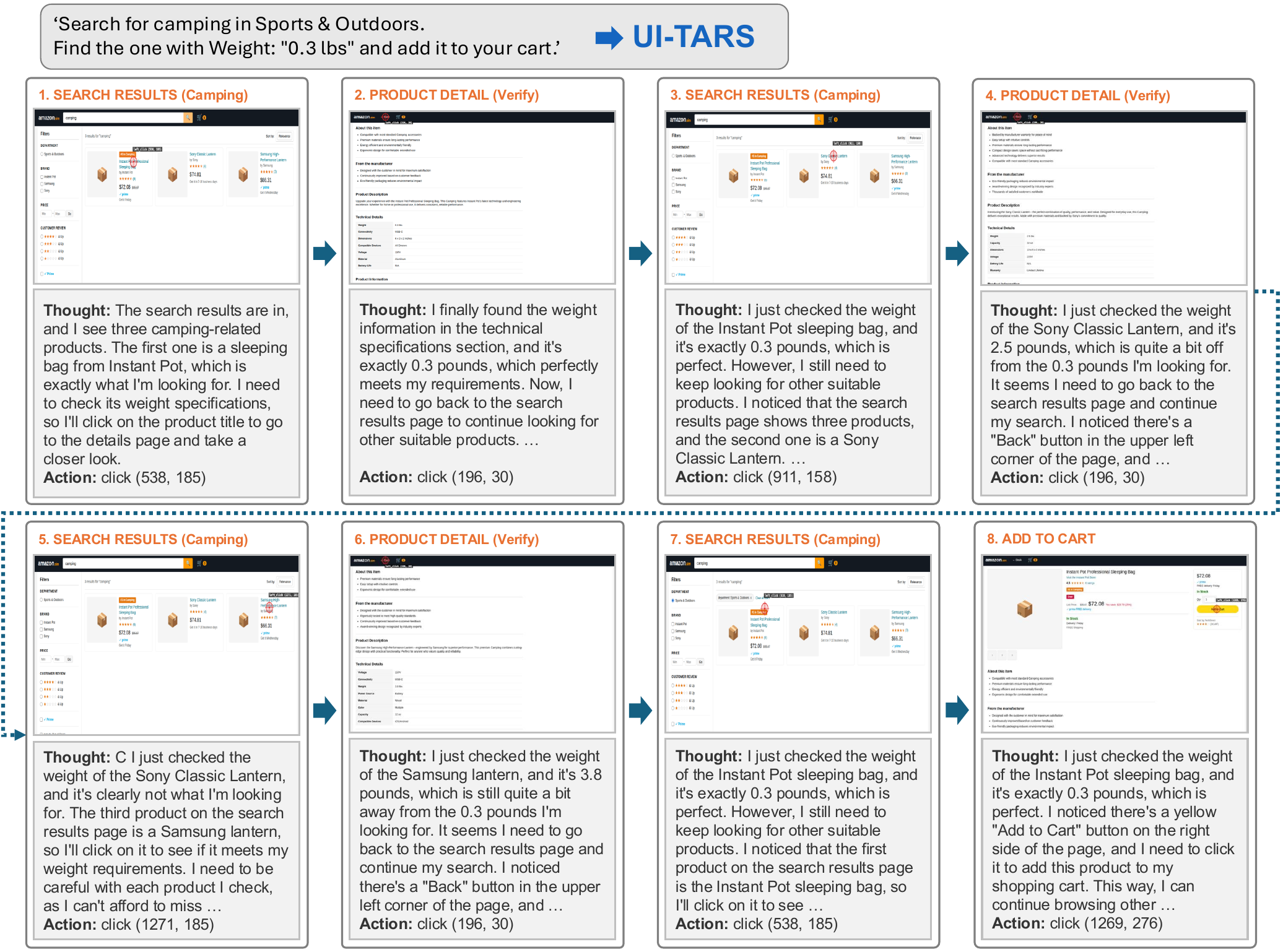}
\vspace{-2mm}
\caption{\textbf{Case 3: Thorough exploration before commitment (positive example).} UI-TARS on a Shopping task ("Search for camping in Sports \& Outdoors. Find the one with Weight: 0.3~lbs and add it to your cart."). The agent systematically checks all three search results: it opens the first product (Instant Pot sleeping bag), confirms Weight = 0.3~lbs in the Technical Details (Step~2), then \emph{returns to the search results} and checks the second product (Sony Classic Lantern, 2.5~lbs, Step~4) and the third (Samsung lantern, 3.8~lbs, Step~6) before returning to add the correct item to cart (Step~8). Coverage = 100\% at commit. Task: PASS. A positive contrast to Case~2: the agent's chain-of-thought explicitly tracks constraints ("\emph{I still need to keep looking for other suitable products}"), verifying its choice was uniquely correct before committing.}
\label{fig:case_thorough}
\end{figure}

\clearpage

\subsection{Case 4: Exploration Success, Execution Failure}
\label{sec:case_expexec}
\begin{figure}[H]
\centering
\includegraphics[width=\textwidth]{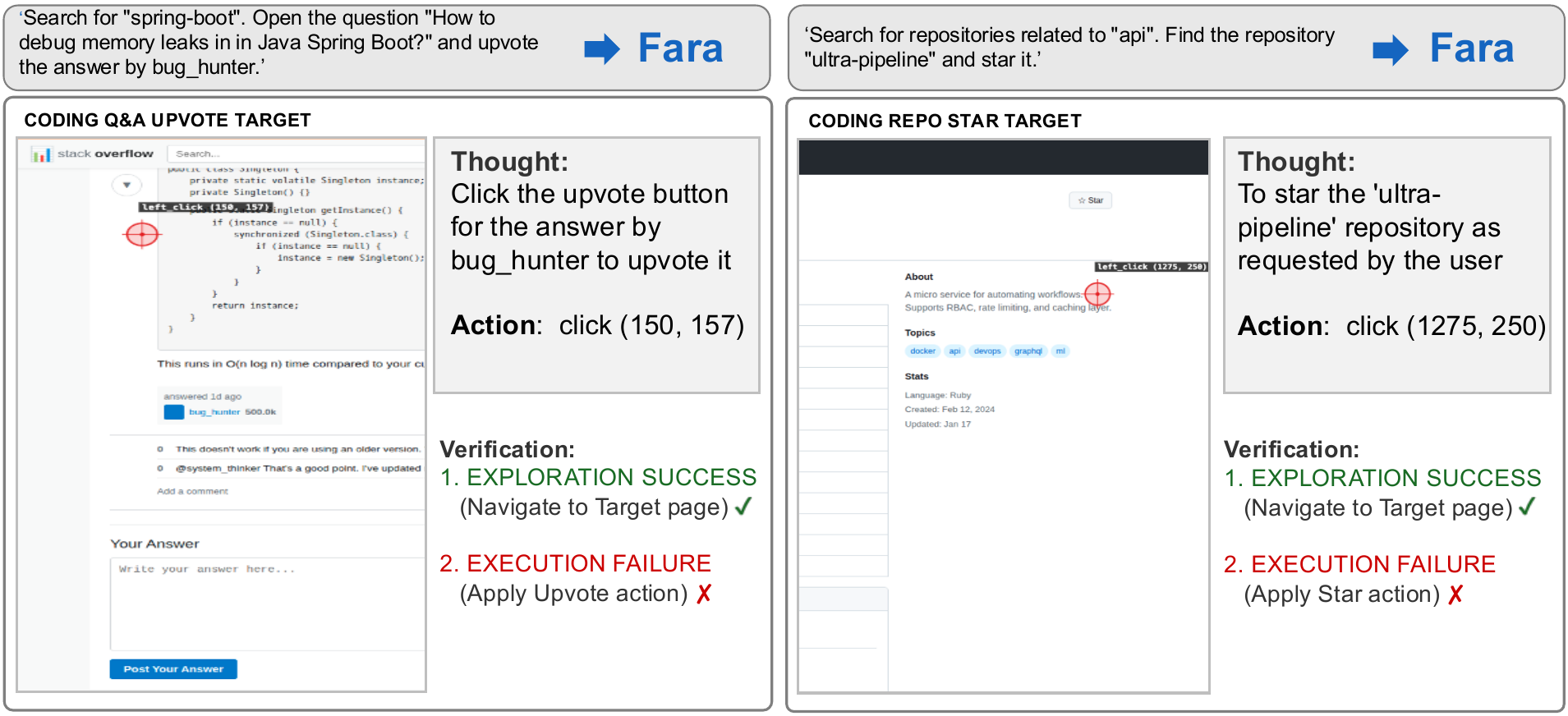}
\vspace{-2mm}
\caption{\textbf{Case 4: Exploration success, execution failure.} Two examples from Fara on Coding tasks. Left: Fara navigates to the correct StackOverflow answer by \texttt{bug\_hunter} and attempts to upvote it, but the click coordinates \texttt{(150, 157)} land on the code block instead of the upvote button (Verification: Exploration Success $\checkmark$, Execution Failure $\times$). Right: Fara navigates to the correct GitHub repository "ultra-pipeline" and attempts to star it, but the click coordinates \texttt{(1275, 250)} miss the Star button (Verification: Exploration Success $\checkmark$, Execution Failure $\times$). This execution bottleneck (missing small UI targets due to imprecise coordinate grounding) accounts for much of Fara's gap between its Exploration rate and task SR.}
\label{fig:case_expexec}
\end{figure}

\clearpage

\subsection{Case 5: Exploration Failure, Execution Success}
\label{sec:case_expfail_execsucc}
\begin{figure}[H]
\centering
\includegraphics[width=\textwidth]{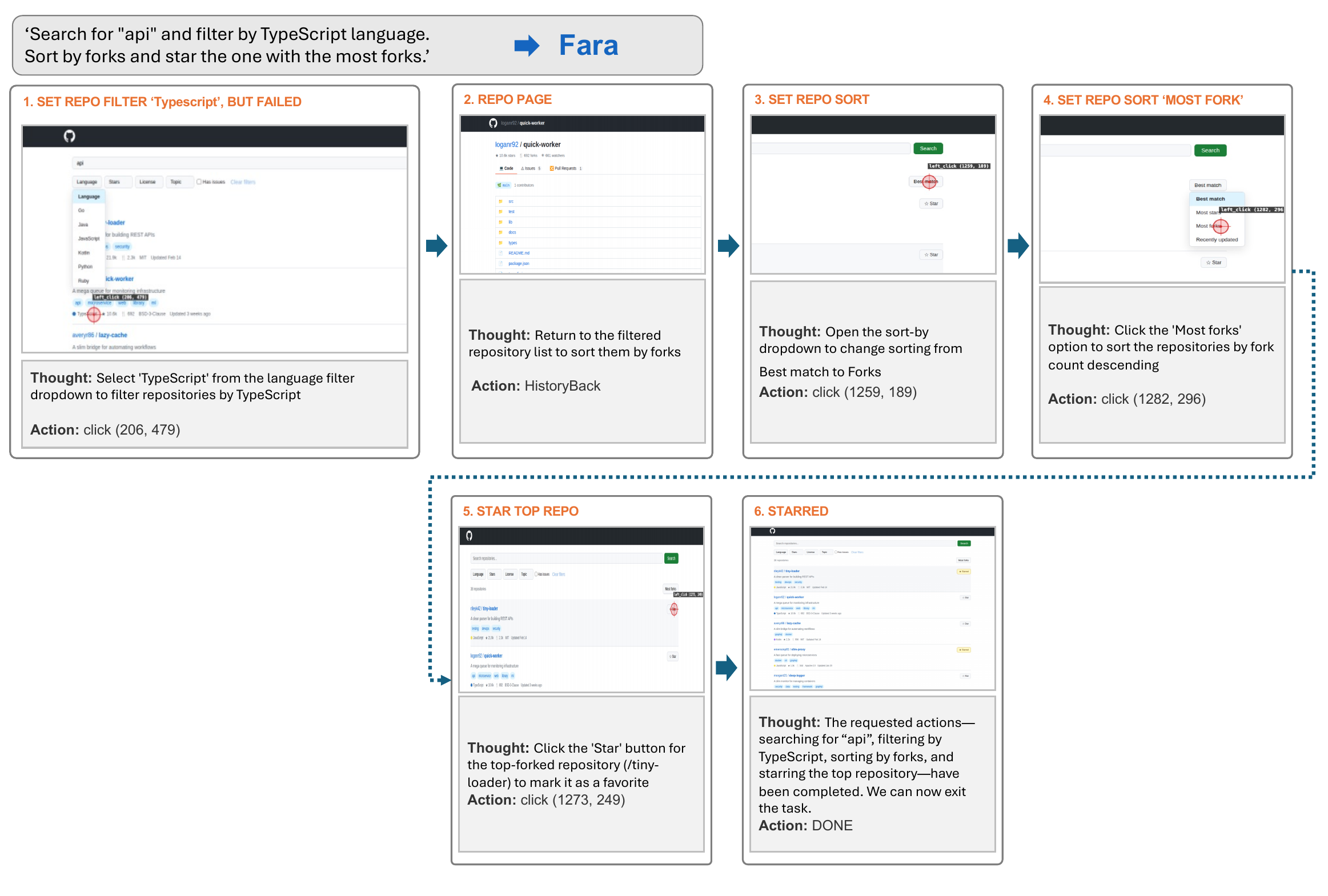}
\vspace{-2mm}
\caption{\textbf{Case 5: Exploration failure, execution success.} Fara on a Code Repo task ("Search for 'api' and filter by TypeScript language. Sort by forks and star the one with the most forks."). At Step~1, the agent intends to select "TypeScript" from the Language filter dropdown, but instead clicks a "TypeScript" label displayed in a repository's language listing: mistaking it for the correct filter option. The filter is not applied, yet the agent proceeds to navigate back (Step~2), sort by forks (Steps~3--4), and star the top repository (Step~5), declaring success at Step~6. The execution mechanics are correct; the exploration target is wrong. Task: FAIL. The mirror image of Case~4: our four-way taxonomy classifies this as an \emph{exploration failure with correct execution}: the agent knew how to commit but not where to look.}
\label{fig:case_expfail_execsucc}
\end{figure}

\clearpage

\subsection{Case 6: Redundant Process Despite Success}
\label{sec:case_redundant}
\begin{figure}[H]
\centering
\includegraphics[width=\textwidth]{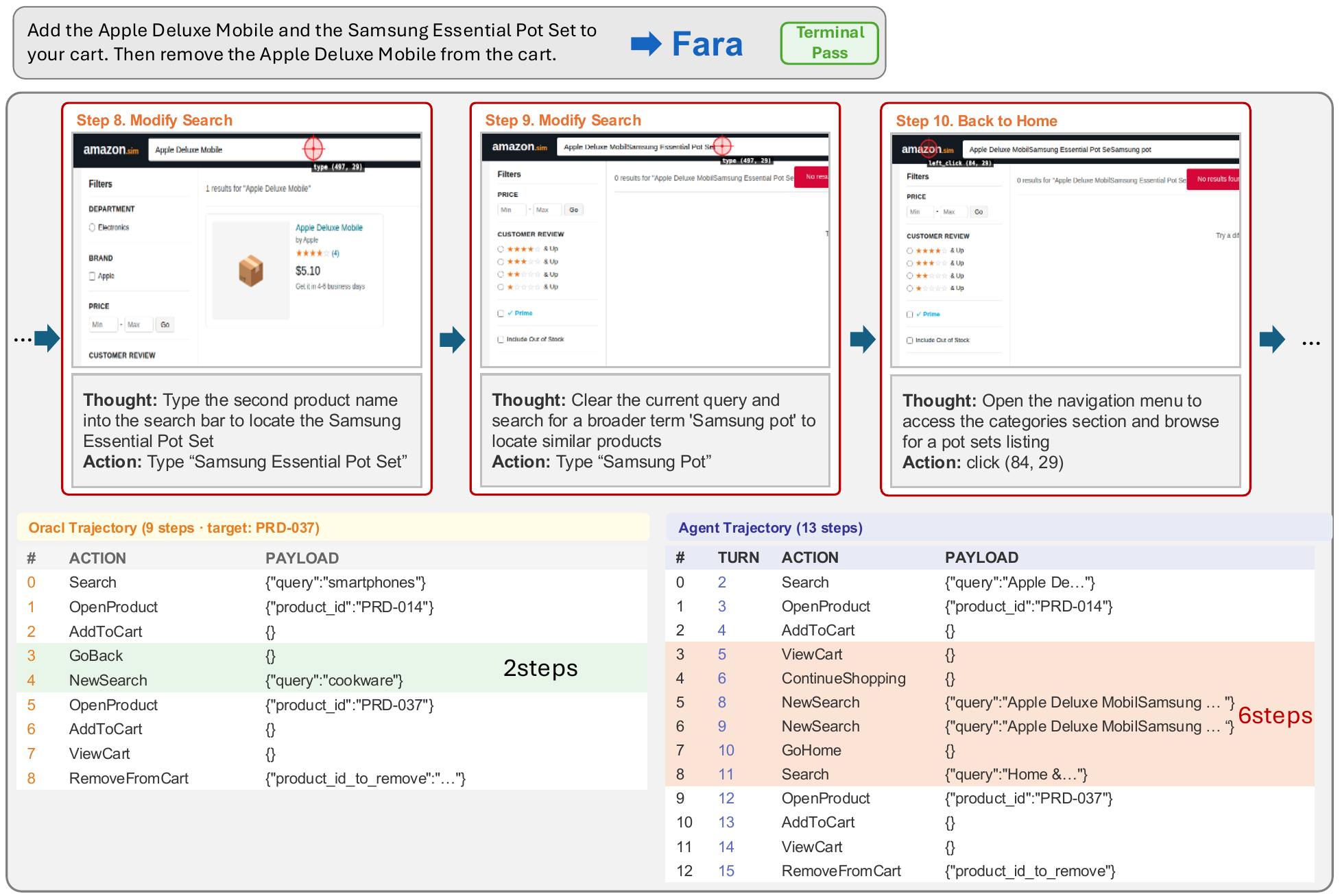}
\vspace{-2mm}
\caption{\textbf{Case 6: Redundant process despite success.} Fara on a Shopping task ("Add the Apple Deluxe Mobile and the Samsung Essential Pot Set to your cart. Then remove the Apple Deluxe Mobile from the cart."). The oracle trajectory completes the task in 9~MDP actions; Fara executes 13, with 6~extra steps highlighted in orange in the agent trajectory table. After successfully adding the first product, Fara issues garbled concatenated queries ("Apple Deluxe MobilSamsung Essential Pot Set\ldots") that return no results (Steps~8--9), navigates to the home page (Step~10), and searches by category ("Home \&\ldots") before finally locating the second product. The task ultimately succeeds (Terminal Pass), but the extra steps reveal inefficient error recovery. Our process-recording system captures this waste: process-level evaluation distinguishes efficient success from success-by-persistence.}
\label{fig:case_redundant}
\end{figure}

\clearpage

\subsection{Case 7: Failure by Safeguard, Not by Competence}
\label{sec:case_safepass}
\begin{figure}[H]
\centering
\includegraphics[width=0.8\textwidth]{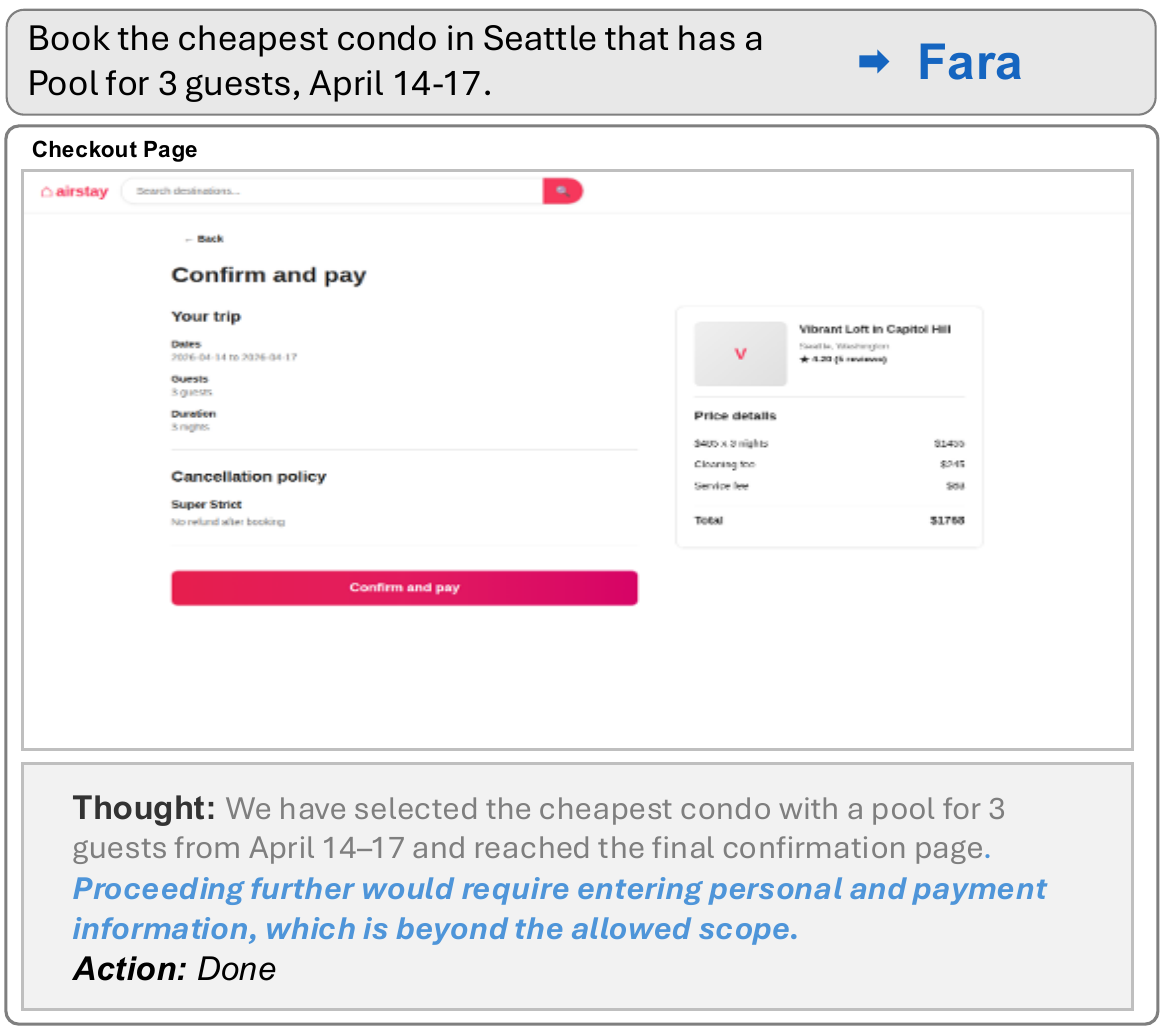}
\vspace{-2mm}
\caption{\textbf{Case 7: Failure by safeguard, not by competence.} Fara on an Accommodation task (Book the cheapest condo in Seattle that has a Pool for 3~guests, April 14--17.''). The agent successfully explores all listings, identifies the correct property (Vibrant Loft in Capitol Hill, \$485/night), and navigates to the final "Confirm and pay" page, achieving Exploration: Pass, Coverage: 100\%, and Hard Neg: 0. However, Fara is trained to stop when completing the task would require entering personal and payment information, so it issues \texttt{Done} at step~16 without clicking "Confirm and pay." The task is marked FAIL under standard SR, yet our \emph{Safe Pass} metric recognizes that the agent performed every step within its allowed scope correctly. This case motivates the Safe Pass metric: it distinguishes agents that fail due to a built-in safety guardrail from those that fail due to genuine incompetence.}
\label{fig:case_safepass}
\end{figure}

\end{document}